\documentclass[twoside,11pt]{article}

\usepackage[preprint]{jmlr2e}
\usepackage{amsmath}
\usepackage{times}
\usepackage{color}
\usepackage{multirow}

\usepackage{rotating}
\usepackage{bbm}
\usepackage{latexsym}

\usepackage{float}
\usepackage{subfig}
\usepackage[dvipsnames]{xcolor}
\usepackage{booktabs} 
\usepackage{mathtools}
\usepackage{hyperref}
\usepackage{xspace}
\usepackage{mdframed}

\newcommand{\nameNeuralCoherencePart}{Component\xspace}

\newmdtheoremenv[ 
  backgroundcolor=gray!10,
  linecolor=gray!50,
  leftmargin=0pt,
  rightmargin=0pt,
  innertopmargin=0.5em,
  innerbottommargin=0.5em
]{definitionbox}{Definition}

\usepackage{lastpage}
\jmlrheading{23}{2025}{1-\pageref{LastPage}}{1/21; Revised 5/25}{9/25}{21-0000}{Simon Guiroy}

\ShortHeadings{Neural Coherence}{Guiroy, Richter, Chandar, Pal}
\firstpageno{1}

\begin{document}

\title{Neural Coherence :\\Find higher performance to out-of-distribution tasks from few samples}

\author{\name Simon Guiroy \email simon.guiroy@umontreal.ca \\
       \addr Mila - Quebec AI Institute\\
       University of Montreal\\
       Montreal, Canada
       \AND
       \name Mats Richter \email mats@hcompany.ai \\
       \addr Mila - Quebec AI Institute\\
       Polytechnique Montreal\\
       Montreal, Canada
       \AND
       \name Sarath Chandar \email sarath.chandar@mila.quebec \\
       \addr Mila - Quebec AI Institute\\
       Polytechnique Montreal\\
       Montreal, Canada
       \AND
       \name Christopher Pal \email christopher.pal@polymtl.ca \\
       \addr Mila - Quebec AI Institute\\
       Polytechnique Montreal\\
       Montreal, Canada}
\editor{My editor}
\maketitle

\begin{abstract}
To create state-of-the-art models for many downstream tasks, it has become common practice to fine-tune a pre-trained large vision model. However, it remains an open question of how to best determine which of the many possible model checkpoints resulting from a large training run to use as the starting point.
This becomes especially important when data for the target task of interest is scarce, unlabeled and out-of-distribution. In such scenarios, common methods relying on in-distribution validation data become unreliable or inapplicable. This work proposes a novel approach for model selection that operates reliably on just a few \emph{unlabeled} examples from the target task. Our approach is based on a novel concept: \textit{Neural Coherence}, which entails characterizing a model's activation statistics for source and target domains, allowing one to define model selection methods with high data-efficiency.
We provide experiments where models are pre-trained on ImageNet1K and examine target domains consisting of  Food-101, PlantNet-300K and iNaturalist. We also evaluate it in many meta-learning settings. Our approach significantly improves generalization across these different target domains compared to established baselines.
We further demonstrate the versatility of Neural Coherence as a powerful principle by showing its effectiveness in training data selection.
\\
\end{abstract}


\clearpage
\section{Introduction}

\begin{figure}[ht]
    \centering
    \subfloat[Evolution of the source and target activation trajectories (omitting $\hat{m}_4$ for 3D visualization). Here their divergence is  highest at layer 4, along $\hat{m}_2$. ]{\includegraphics[width=0.36\textwidth]{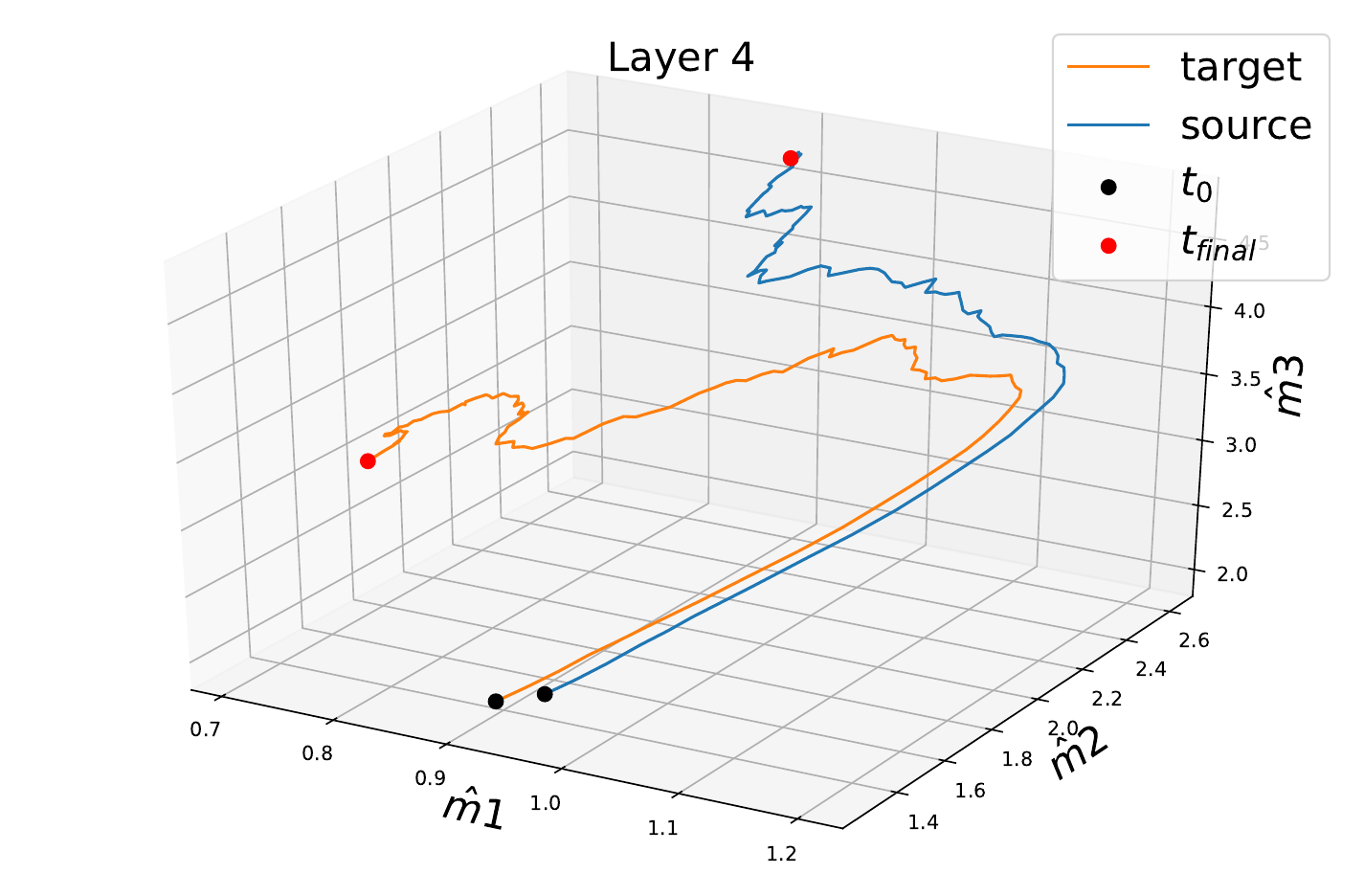}\label{fig:exp:divergence:a}.}
    \hskip 0.1in
    \subfloat[Target and source activations trajectories at layer 4 and along $\hat{m}_2$, their divergence is detected at time  $\hat{t}$, after which their plots become negatively correlated.]{\includegraphics[width=0.28\textwidth]{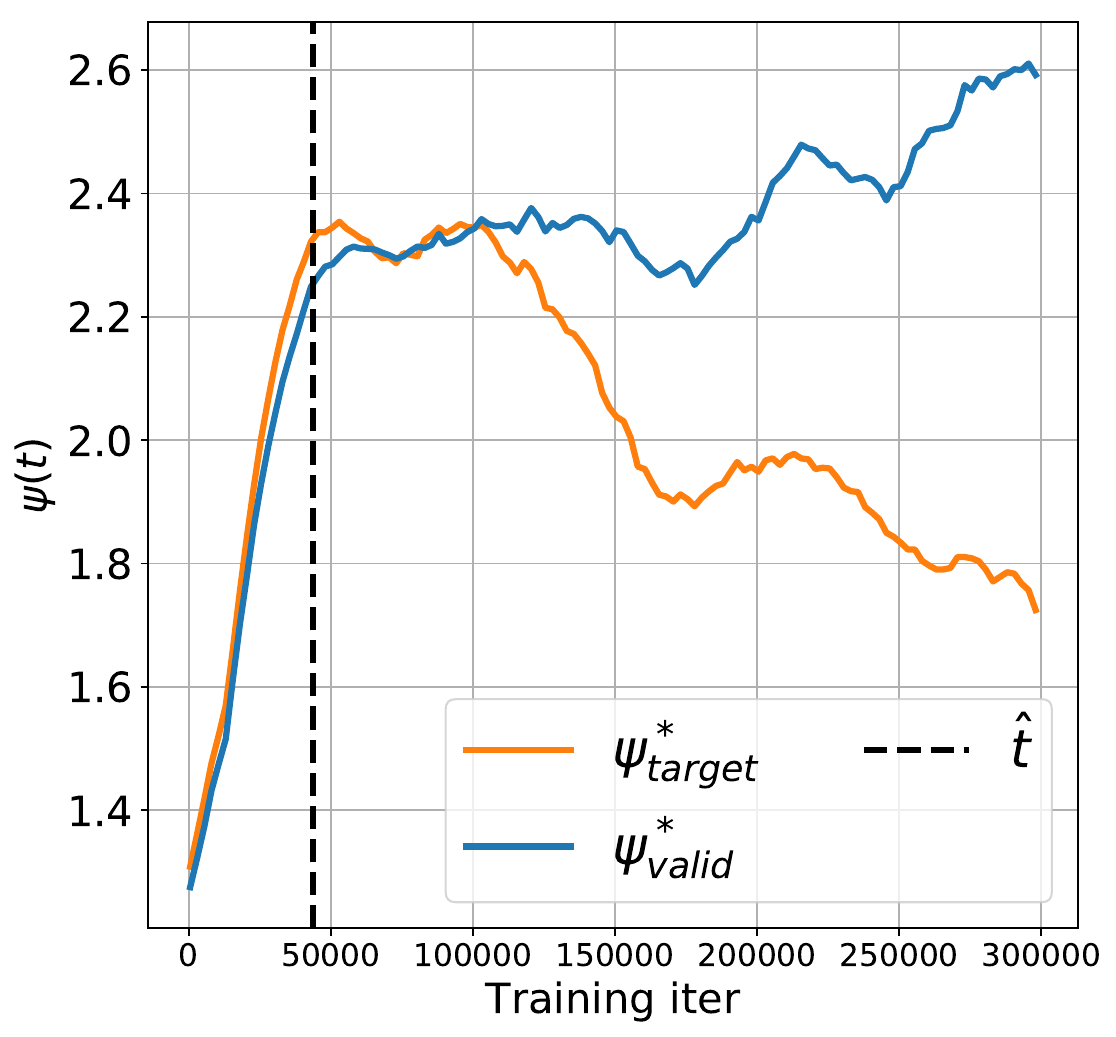}\label{fig:exp:divergence:b}}
    \hskip 0.1in
    \subfloat[Neural Coherence improves target generalization by early-stopping at time $\hat{t}$, much closer to the true optimum $t^*$ compared to $t^*_{valid}$ of the validation accuracy.]{\includegraphics[width=0.285
    \textwidth]{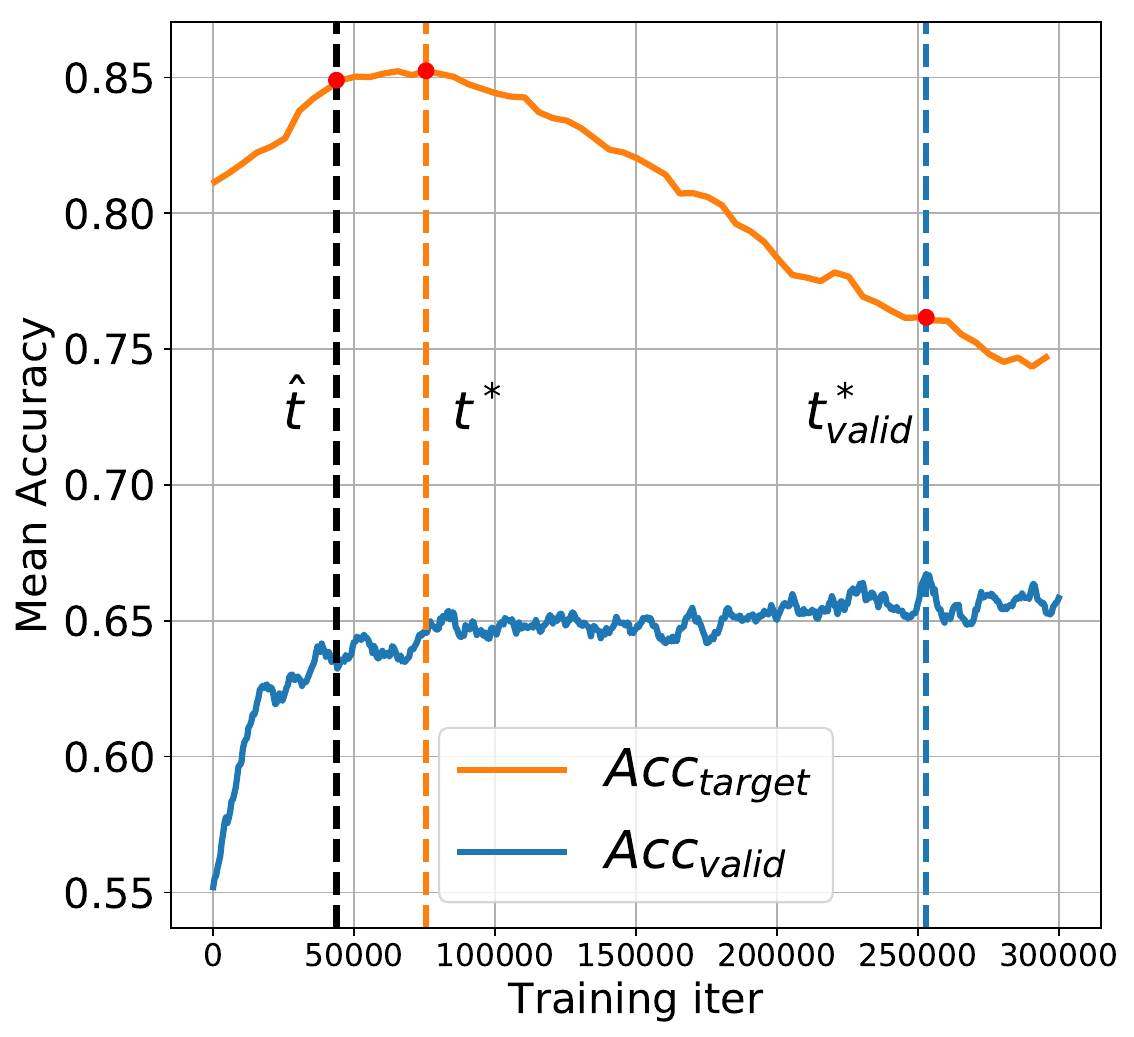}\label{fig:exp:divergence:finding_t}}
    \caption{Inferring when to stop in few-shot transfer learning from the unlabelled efxamples (5) of a single target task, using Neural Coherence. Setting shown: MAML, CNN, Quickdraw to Omniglot, 5-way 1-shot. We first identify the critical layer $l^*$ and critical moment $\hat{m}^*$, where neural activation dynamics of the target inputs diverge the most from those of the source domain. The strongest divergence is detected at layer 4 along $\hat{m}_2$. We then early-stop at $\hat{t}$, i.e. when the Pearson correlation of $\psi(\mathbf{z}_{target}; t)$ and $\psi(\mathbf{z}_{valid}; t)$, both evaluated at $l^*$ and $\hat{m}^*$, flips from being positive between $t_0$ and $\hat{t}$, to being negative after $\hat{t}$. Here, activation-based early-stopping at $\hat{t}$ achieves much greater target generalization than validation-based early-stopping at $t^*_{valid}$.}
    \label{fig:analysis:activation-based_early-stopping}
\end{figure}

In recent years, the practice of using large models pre-trained on huge datasets has become the de facto standard for attaining high performance on various downstream tasks \citep{efficientnet2, convnext, convnext2}. This paradigm offers a compelling advantage: the ability to utilize large, powerful models when downstream target data is scarce, which is a true constraint for many real-world problems \citep{imagenet_medical}. Furthermore, this allows significant computational resources to be expended once to create a foundation model that is subsequently used by many others for various tasks. Consequently, a key challenge lies in effectively transferring and adapting the representations of pre-trained models to the specific downstream problem at hand, so that downstream performance is maximized. Since downstream tasks are frequently out-of-distribution (OOD) with respect to the original training data, maximum downstream performance may not coincide with the maximum validation set performance of the original dataset during pre-training \citep{2020arXiv200701434G}. Creating a large pre-trained model typically requires significant investments of resources, but storing checkpoints of model parameters throughout training is common practice. Because fine-tuning from different checkpoints can lead to substantial differences in performance, in this paper, we focus on this particular type of \textbf{model selection}. 
For instance, and perhaps surprisingly, it is quite common to perform foundation model selection through the heuristic of simply taking the model obtained after a fixed number of epochs known to have worked well in the past \citep{imagenet21k, efficientnet, efficientnet2, coatnet, vit}. Alternatively, another common strategy is to rely on an in-distribution validation (Source-Val) set to stop the pre-training early \citep{greyscale_imagenet, chinchilla}.
\\
\\
In this work, we propose a method for this type of checkpoint selection. This method is based on a novel principle we refer to as \textbf{Neural Coherence}, which takes the layer activation statistics over the course of the training into account.
Leveraging activation statistics in this way allows us to do effective checkpoint selection, even when the data from the target domain is unlabeled and extremely scarce.
More concretely, we show that our method can provide \textbf{good checkpoint selection results with as little as 5 unlabeled samples from the target domain.} See Fig. \ref{fig:analysis:activation-based_early-stopping} for a qualitative demonstration.
Finally, we demonstrate that Neural Coherence is not necessarily limited to selecting checkpoints during a single training run, but can also be used for other model selection tasks, such as selection of pre-training data.
Neural Coherence is a generalization of Activation Based Early-Stopping (ABE), a specific model selection method, itself limited to early-stopping.
In this work, we make the following contributions:
\begin{itemize}
\itemsep0em
    \item We define the general framework of the Neural Coherence principle, to identify trained models with higher performance on an out-of-distribution target task, given a few target samples.
    \item We present an implementation of the principle for the problem of model checkpoint selection.
    \item We empirically validate the effectiveness and generality of our approach in various OOD settings: few-shot learning, zero-shot generalization, transfer learning; training regimes: meta-learning, supervised pre-training (and self-supervised); neural architectures, and many datasets.
    \item We demonstrate the versatility of Neural Coherence by applying it to the problem of selecting good pre-training data, given a few examples from the out-of-distribution target task.
\end{itemize}
The remainder of this paper is organized as follows: 
In Section \ref{sec:neural_coherence}, we introduce a high level formulation of the Neural Coherence principle.
In Section \ref{sec:related_works} we review the related literature.
Next, in Section \ref{sec:practical_implementation}, we lay out a practical implementation of Neural Coherence, and in Section \ref{sec:practical_implementation:checkpoint_selection} we apply it to checkpoint selection.
We then present our experiments on Neural Coherence. We first present our results on checkpoint selection in Section \ref{sec:exps:ckpt_selection}, which are first applied to small models in the context of meta-learning for few-shot classification. We then present our results on larger-scale setups, with vision foundation models applied in zero-shot learning and transfer learning, along with a study of statistical efficiency of our proposed method.
Having demonstrated the results on checkpoint selection, we then introduce in Section \ref{sec:exps:data_selection} a formulation for applying Neural Coherence to the selection of training data, given an OOD task a a few examples, with our experimental results of this application.
In Section \ref{sec:exps:foundation}, we provide the chain of experimental work and analyses that lead to Neural Coherence in its current form.
Section \ref{sec:conclusion} concludes this work and Section \ref{sec:app:exp_detail}
contains additional experimental details.

\section{The Neural Coherence Principle} \label{sec:neural_coherence}
We can outline Neural Coherence as follows: 
\textit{While incrementing a hyperparameter improves the performance of trained models on the source task, given a few examples from the target task, observe their neural \textbf{activations across all layers} to \textbf{characterize their distribution} and follow its \textbf{trajectory}, while \textbf{contrasting it with the source} trajectory. Keep incrementing the hyperparameter as long as the two trajectories remain \textbf{coherent}, but stop if they start going in opposite directions.}

In this section, we will introduce the formulation of the Neural Coherence principle, which we break down into five constituent components. Later in Sec. \ref{sec:practical_implementation}, we will present an implementation of this principle, applied to the problem of checkpoint selection. Let us now begin by introducing the problem setting. A deep neural network or model $f$, parametrized by a parameter vector $\theta$, given an input sample drawn from a distribution, i.e. $\mathbf{x} \sim p(\mathbf{x})$, produces a prediction output such that $\hat{\mathbf{y}} = f(\mathbf{x}; \theta)$.
A deep neural network $f$ is composed of many layers $f_1, f_2, ..., f_L$ each producing their own outputs.
The neural network is trained by optimizing some loss function $\ell \big( f(\mathbf{x}; \theta), \mathbf{y} \big)$, 
over a dataset $\mathcal{D} = \{ (\mathbf{x}_i, \mathbf{y}_i) \}_{i=1}^N \sim p(\mathbf{x}, \mathbf{y})$ of pairs of input examples and their corresponding ground-truth outputs or labels, so as to minimize the empirical risk $\hat{\mathcal{L}} = \frac{1}{N} \sum_{i=1}^N \ell \big( f(\mathbf{x}_i; \theta), \mathbf{y}_i \big)
$, where the goal is to achieve minimal true expected risk $\mathcal{L} = \mathbb{E}_{(\mathbf{x}, \mathbf{y}) \sim p(\mathbf{x}, \mathbf{y})} \Big[ \ell \big( f(\mathbf{x}; \theta), \mathbf{y} \big) \Big]$.
The optimizer $\mathcal{O}$ is a function of the model architecture, its parameter vector, and the set $\Omega$ of all other hyperparameters, and results in the optimized parameter vector $\theta^*$ of the network, such that $\theta^* = \mathcal{O} ( f, \theta, \Omega )$. 
For instance, with Stochastic Gradient Descent, $\mathcal{O}_{\text{SGD}} : \;\; \theta_{t+1} = \theta_t - \alpha \nabla_{\theta} \hat{\mathcal{L}}(f_{\theta_t}, \mathcal{D}_t) \;\; \text{where} \;\; \mathcal{D}_t = \{ (\mathbf{x}_i, \mathbf{y}_i) \}_{i=1}^{\beta} \overset{\underset{\mathrm{i.i.d.}}{}}{\sim} \mathcal{D}$.
The hyperparameters can thus include many tunable training options such as the amount of training time or iterations $t$, the learning rate $\alpha$, mini-batch size $\beta$, and even the training data itself $\mathcal{D}$, or augmentations of it (e.g. $\Omega = \{ t, \alpha, \beta, \mathcal{D} \}$),or any other hyperparameters of the training loss function, like regularization coefficients.
Note that even if we can select or tune the training data, the actual downstream task $\mathcal{T}$ is defined by a target data distribution, along with the loss function to be minimized over it, $\mathcal{T} = \{ p(\mathbf{x}, \mathbf{y}), \ell \}$. 
In most OOD contexts, the model is trained on a source task by minimizing $\mathcal{L}_S$, but to be deployed on a target (downstream) task, which usually has very few examples available, and even fewer labeled examples (otherwise it could be learned directly).
In such scenarios, the goal is thus to get the best trained model $f_{\theta^*}$ for the target task, hence to optimize $\Omega$ in order to minimize $\mathcal{L}_T$.
In this section, we present our Neural Coherence principle for inferring good hyperparameters given a few examples from the OOD target task. 
We will now break down the presentation of Neural Coherence into its five constituent components. 

\paragraph{\nameNeuralCoherencePart 1: Neural activations throughout the network.}
The first component of Neural  consists in analyzing the neural activations across the network depth, rather than solely logits or final representations (embeddings).
Given an input sample, the neural network produces activations at the outputs of its various layers. In this context, we refer to the neural activations as the concatenation of all those activations. We choose to characterize activations across the full depth of the network architecture, rather than focusing solely on final outputs or penultimate representations

\begin{definitionbox}[Neural activations]
Given a neural network $f(\mathbf{x}; \theta)$, with parameter vector $\theta$ and $L$ layers $f_1$ to $f_L$, and given an input sample drawn from a data distribution $\mathbf{x} \sim p(\mathbf{x})$, the neural activations $\mathbf{z}$ are defined as :
\begin{equation}
    \mathbf{z} \doteq [ f_1(\mathbf{x}), f_2(\mathbf{x}), \cdots, f_L(\mathbf{x}) ]  \;\; \text{given}\;\; f(\mathbf{x}; \theta) \;\; \text{and}\;\; \mathbf{x} \sim p(\mathbf{x})
\label{eq:activations}
\end{equation}
\end{definitionbox}

\begin{figure}[ht]
    \centering
    \includegraphics[width=0.5\linewidth]{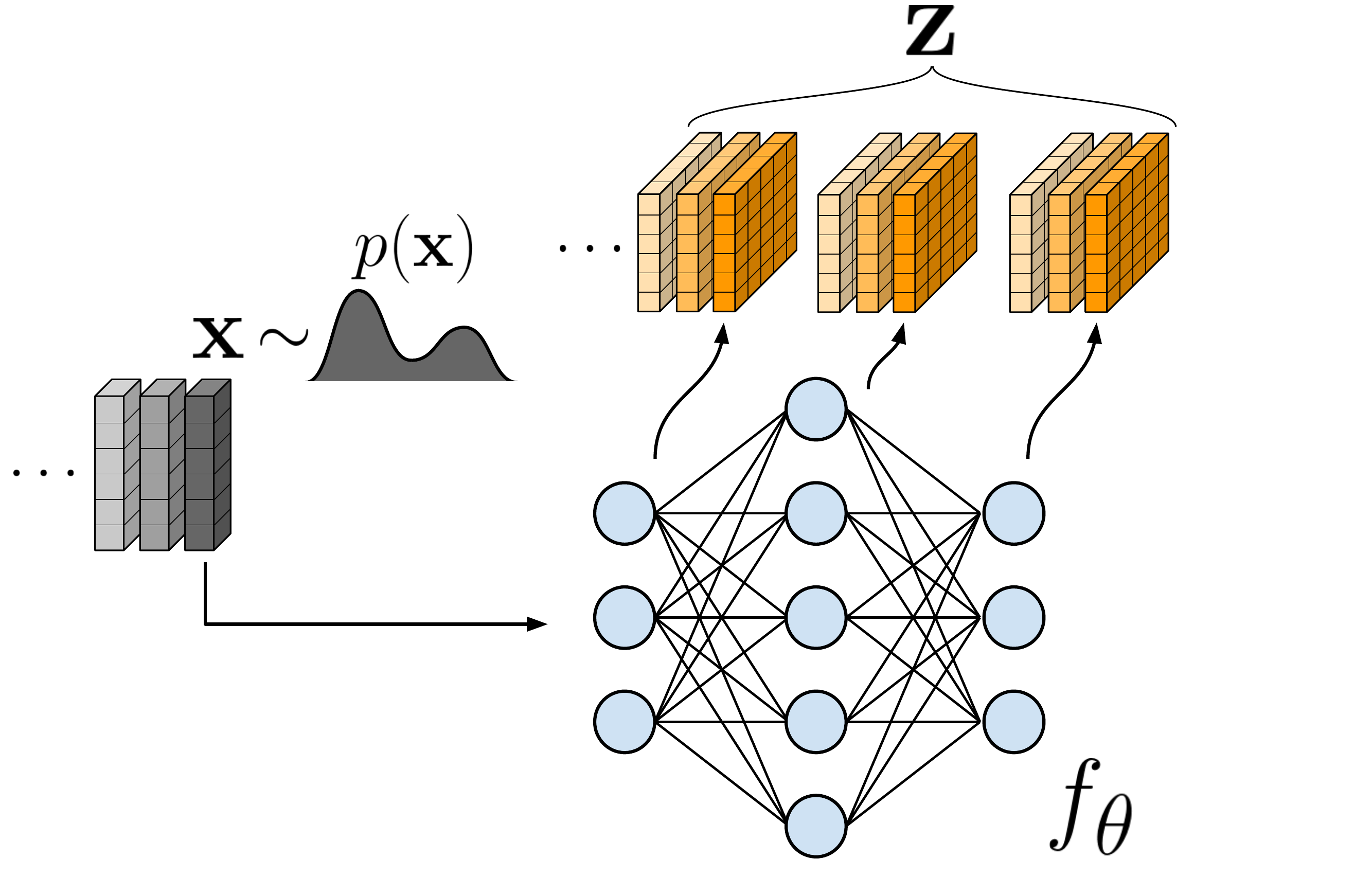}
    \caption{(component 1) For a neural network $f$ parametrized by $\theta$ and a set of inputs $\mathbf{x}$ drawn from a given data distribution $p(\mathbf{x})$, we analyze the distribution of their activations $\mathbf{z}$ across the  network.
    }
    \label{fig:neural_coherence:concept-1}
\end{figure}

Intermediate activations can offer improved statistical efficiency for identifying how the network behaves as a function of an input distribution. This is because the feedforward connectivity inherent to neural networks—whether fully or partially—creates a hierarchical dependency structure, whereby final outputs are influenced by the entire cascade of intermediate representations. Consequently, the model’s loss is not only a function of the final output but also implicitly shaped by patterns in intermediate activations.

However, accurately estimating this loss can be statistically challenging, particularly when data is limited. High variance and poor statistical efficiency can impair the reliability of loss-based methods—challenges that are exacerbated in scenarios with very few, or entirely unlabeled, target examples. In such cases, the final prediction layer (or task-specific head) often becomes a major source of noise, especially when it cannot be fine-tuned. In contrast, the loss may in fact be driven by a latent factor—namely, a shift in the distribution of intermediate activations—that is more stable and statistically efficient to estimate \citep{DBLP:journals/corr/YosinskiCBL14}.
Moreover, due to the increasing functional complexity of representations across layers \citep{raghu2017expressivepowerdeepneural}, disruptions in generalization may become less identifiable at the network’s output. Intermediate layers may therefore offer a clearer window into the causes of generalization breakdown. Importantly, despite differences in architecture, most neural networks (including recurrent ones when unfolded) exhibit a finite, ordered sequence of activations that can be analyzed layer by layer. Finally, one of our assumptions is that, under a strong distributional shift between source and target data distributions, which would induce a more severe generalization gap, these latent disruptions are likely to be more pronounced—making them easier to detect and quantify through patterns in intermediate activations. And inversely under the same assumption, low distributional shift would make it harder to detect the breakdown in neural coherence from those patterns, resulting in the selection of the same $\Omega^*$ as for the source domain (validation set), but the target optimum is likely to be close to the source optimum. This hypothesis is backed by and derived from our empirical observations (Sec.\ref{sec:exps:ckpt_selection:meta-learning},  \ref{sec:exps:foundation}). 

\paragraph{\nameNeuralCoherencePart 2: Characterizing the distribution of neural activations}
Now that we observe the activations across the network, we need a way to characterize their distribution, in order to identify the model in function space \citep{benjamin2018measuring}, as a function of its input distribution.
As the input samples are drawn from some data distribution $p(\mathbf{x})$, given a fixed model $f$, there is a corresponding neural activation distribution $p(\mathbf{z})$ :
\begin{equation}
    p(\mathbf{z}) = p\big([f_1(\mathbf{x}), f_2(\mathbf{x}), \cdots, f_L(\mathbf{x})]\big)  \;\; \text{given}\;\; f(\mathbf{x}; \theta) \;\; \text{and}\;\; \mathbf{x} \sim p(\mathbf{x})
\label{eq:activations_distribution}
\end{equation}
We can sample the input distribution, to obtain the empirical distribution of neural activations, and map it to a point $\psi(\mathbf{z})$ in a vector space:

\begin{definitionbox}[Characterization of the neural activations distribution]
Let \( p(\mathbf{z}) \) be the distribution of neural activations, as defined in Eq. \ref{eq:activations_distribution}, induced by a network \( f(\mathbf{x}; \theta) \) and input distribution \( p(\mathbf{x}) \). Given $N$ samples \( \{ \mathbf{z}_i \}_{i=1}^N \overset{\underset{\mathrm{i.i.d.}}{}}{\sim} p(\mathbf{z}) \), where $\mathbf{z} \in \mathbb{R}^D$, we define :

\begin{equation}
    \psi(\mathbf{z}) : \mathbb{R}^{N \times D} \mapsto \mathbb{R}^M
    \quad \text{with} \quad M \ll D
\end{equation}

as a low-dimensional, statistically efficient mapping that characterizes the neural activations distribution. Ideally, \(\psi\) would be bijective, i.e. uniquely determines $p(\mathbf{z})$, but in practice it is only required to retain sufficient information to distinguish relevant differences between activation distributions.
\end{definitionbox}

\begin{figure}[ht]
    \centering
    \includegraphics[width=0.75\linewidth]{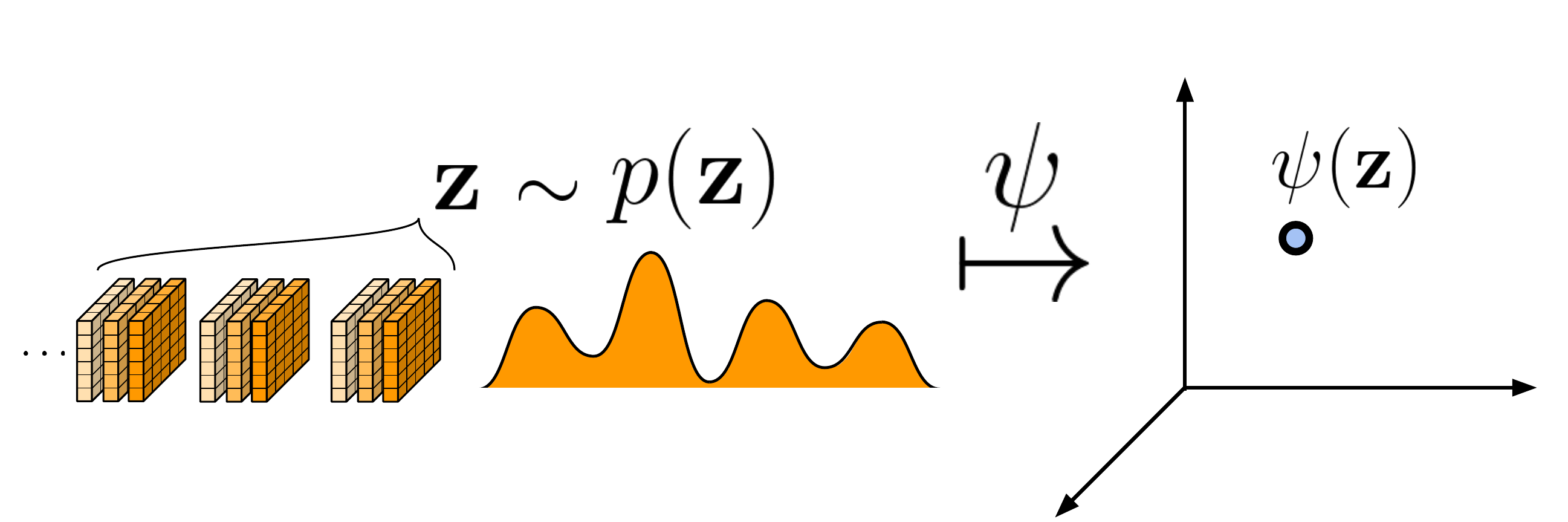}
    \caption{(component 2) We characterize the empirical distribution of activations $p(\mathbf{z})$ with a transformation $\psi$ that maps it to a low-dimensional point  $\psi(\mathbf{z})$ in a vector space.
    }
    \label{fig:neural_coherence:concept-2}
\end{figure}

Since an object in $\mathbb{R}^{N \times D}$ is potentially of very high dimensionality, tracking and analyzing it could be impractical. 
We thus need to map it to a more compact representation. 
In the ideal case, this mapping would be a bijection, such that any distribution $p(\mathbf{z})$ corresponds to exactly one characterization point $\psi(\mathbf{z})$, and vice versa. 
Although this might be difficult to achieve, especially with the constraint of low dimensionality, and even more so because we typically have very few examples from the target data. 
This mapping should thus be expressive enough to capture most of the useful information about $p(\mathbf{z})$, while limiting its complexity so as to be compact and statistically efficient to estimate accurately. Moreover, many methods in deep learning, namely regularization methods, rely on computing and adjusting batch statistics on intermediate activations throughout the network \citep{ioffe2015batch,ba2016layer,wu2018group,ulyanov2016instance}.

\paragraph{\nameNeuralCoherencePart 3: Neural activation trajectory}
Given a characterization of activation distributions, the third component of Neural Coherence analyzes how these distributions evolve along~$\Omega$, what we call the \emph{neural activation trajectory}, rather than evaluating any single activation state in isolation.

Considering an entire trajectory leverages substantially more information than single-point evaluations, prone to misestimation due to low signal-to-noise ratios.
A core motivation for analyzing trajectories is the classical \emph{underfitting--overfitting} behavior of model performance. Our working assumption is that the target loss $L_T$, as a function of a tunable hyperparameter $\Omega$, is typically dominated by this regime structure: an initial underfitting phase, followed by an overfitting phase, with the optimum lying at the transition between the two. 
Empirically, our experiments suggest that $L_T$ often follows a characteristic unimodal or approximately convex trend across hyperparameter increments, with a clearly identifiable global minimum, an observation made by other works \citep{Li2025probing}.
Because model selection is chiefly concerned with locating this optimum---rather than estimating the exact value of $L_T$ at each point---analyzing the ordered model sequence is both sufficient and significantly more robust.
Importantly, neural activation distributions characterize the \emph{state} of the model with respect to its input domain.
Thus, the neural activation trajectory reflects the evolution of this state as the hyperparameter varies. 
When passing the optimum, the generalization trend will change its course, between the underfitting and overfitting regimes, and such changes should manifest as detectable shifts in activation dynamics. 
However, detecting such transitions requires a suitable \emph{reference} for interpreting changes in the activation trajectory---this role is fulfilled by the fourth component of Neural Coherence.
Identifying a metric that reliably estimates generalization is notoriously difficult \citep{jiang2019fantastic}. In contrast, \emph{identifying the best-performing model among a sequence} is substantially easier---especially when the sequence possesses meaningful structure rather than being an unordered set. 
Many hyperparameters of practical interest are numeric and therefore induce a natural order; if generalization along this sequence is not random but exhibits coherent trends, leveraging this ordered structure makes it significantly easier to locate the optimum.
\\
\\
For a given configuration $\Omega_i$ of the set of training hyperparameters $\Omega$, we obtain a trained model with its optimized parameter vector, $\theta^*_{\Omega_i} \doteq \mathcal{O} ( f, \theta, \Omega = \Omega_i)$. Given an input distribution $p(\mathbf{x})$, we get $p(\mathbf{z}_{\Omega_i}) = p([ f_1(\mathbf{x}; \theta^*_{\Omega_i}), f_2(\mathbf{x}; \theta^*_{\Omega_i}), \cdots, f_L(\mathbf{x}; \theta^*_{\Omega_i}) ])$, the neural activation distribution, and its corresponding characterization point $\psi(\mathbf{z}_{\Omega_i}) \doteq \psi \big( \mathbf{z} = \mathbf{z}_{\Omega_i} \big) $, which characterizes the state of trained model, as a function of the input distribution. By varying the set of hyperparameters, 
we obtain, for the same input distribution, another optimized parameter vector, trained model, neural activation distribution, and thus finally, we get another characterization point of the neural activation distribution. 
By sequentially incrementing a training hyperparameter, e.g. number of training iterations $t$, we have a sequence of hyperparameter values $\big( \Omega_i \big)_{i=0}^{\tau} = \big( \{ t= t_i, \alpha, \beta, \mathcal{D} \} \; | \; i = 0, 1, ..., \tau \; ; \; t_{i+1} > t_i \big)
    $.
From this sequence $\big( \Omega_i \big)_{i=0}^{\tau}$ we thus get a sequence of points characterizing the neural activations distribution, which we refer to as the neural activation trajectory.

\begin{definitionbox}[Neural activation trajectory]
Let \( (\Omega_i)_{i=0}^{\tau} \) be a sequence of $\tau +1$ monotonically increasing values for a training hyperparameter $\Omega$, and let \( p(\mathbf{x}) \) be a fixed input distribution. For each \( \Omega_i \), let $f(\mathbf{x}; \theta^*_{\Omega_i})$ denote the model trained using \( \Omega_i \), where $\theta^*_{\Omega_i} = \mathcal{O} ( f, \theta, \Omega_i )$. Let \( \mathbf{z}_{\Omega_i} \) denote the neural activations of this model, with respect to $p(\mathbf{x})$, and let \( \psi(\mathbf{z}_{\Omega_i}) \) be the characterization of $p(\mathbf{z}_{\Omega_i})$. We define the neural activation trajectory over \( \Omega \) as :

\begin{equation}
    \psi(\mathbf{z}; \Omega) \doteq \Big( \psi\big(\mathbf{z}_{\Omega_i}\big) \Big)_{i=0}^{\tau}
\end{equation}
This trajectory describes the evolution of the neural activations distribution as a function $\Omega$.

\end{definitionbox}

\begin{figure}[ht]
    \centering
    \includegraphics[width=0.75\linewidth]{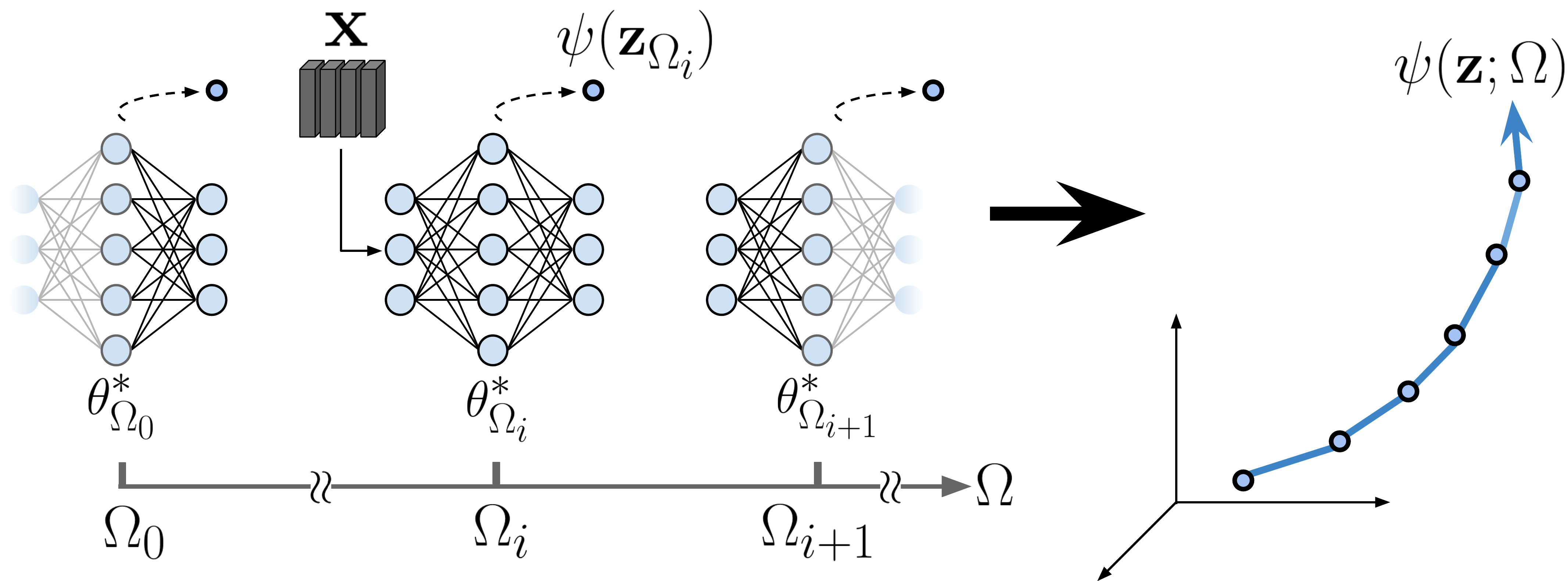}
    \caption{(component 3) By sweeping a training hyperparameter $\Omega$ (e.g. total number of training iterations, learning rate, batch size), we obtain a sequence of trained (or partially trained) models with parameter vector $\theta_{\Omega_i}^*$, for each value of $\Omega_i$. For a same given set of input $\mathbf{x}$, each such model yields a multivariate point $\psi(\mathbf{z}_{\Omega_i})$ characterizing its activations distribution. The sequence of those points constitute the ``neural activation trajectory'' $\psi(\mathbf{z}, \Omega)$ along the dimension $\Omega$.
    }
    \label{fig:neural_coherence:concept-3}
\end{figure}

\paragraph{\nameNeuralCoherencePart 4: Contrasting source and target activation trajectories}
Having defined the notion of neural activation trajectory, here we introduce the fourth component of Neural Coherence, namely the necessity of performing a contrastive analysis between the target with the source trajectories, rather than examining either one in isolation. This contrastive approach is essential: focusing solely on the source activation trajectory would make the analysis agnostic to the target task, potentially leading to suboptimal models \citep{2020arXiv200701434G}. On the other hand, relying exclusively on the target is both impractical and insufficient, as target data is typically scarce, often unlabeled, and does not provide enough signal to reliably characterize generalization dynamics. Moreover, the mechanisms driving generalization on the target domain are often unknown, making it difficult to interpret target-only behavior.

Our goal is to identify, when comparing the target and source activation trajectories, the point where learning of the source task (source generalization increases) begins to degrade performance on the target task, that is, the onset of overfitting from the perspective of the target distribution. 
Importantly, the evolution of the target activation trajectory is not arbitrary; it is a response to the model’s progressive optimization on the source task. This inherent dependency suggests that insights about generalization breakdown on the target can only be meaningfully interpreted in the context of the source training trajectory. For this reason, we adopt a contrastive analysis between source and target activation trajectories, in order to identify when the latent representations learned from the source begin to misalign with the target task. This comparison serves as a critical signal for model selection, particularly under distribution shift and limited supervision. Given a model $f$, for any input distribution $p_A(\mathbf{x})$, we can analyze its neural activations: $\mathbf{z}_{\mathbf{x}=\mathbf{x}_A} \doteq [ f_1(\mathbf{x}_A), f_2(\mathbf{x}_A), \cdots, f_L(\mathbf{x}_A) ]  \;\; \text{given}\;\; \mathbf{x}_A \sim p_A(\mathbf{x})
    $. 
Thus, for the sequence of models obtained by varying the hyperparameters, given  source and target input examples, we can analyze and compare the source and target neural activation trajectories $p_{\text{source}}(\mathbf{x})$ and $p_{\text{target}}(\mathbf{x})$.

\begin{definitionbox}[Source and target activation trajectories]
Given the definition of a neural activation trajectory, and given the source and target input data distributions $p_{\text{source}}(\mathbf{x})$ and $p_{\text{target}}(\mathbf{x})$, we define the source and target activation trajectories respectively as :
\begin{equation}
    \psi(\mathbf{z}_S; \Omega) \doteq \psi \big( \mathbf{z}_{\mathbf{x} = \mathbf{x}_S} \, ; \, \Omega \big) \;\; \text{where} \;\; \mathbf{x}_S \sim p_{\text{source}}(\mathbf{x})
\label{eq:trajectory_source}
\end{equation}
\begin{equation}
    \psi(\mathbf{z}_T; \Omega) \doteq \psi \big( \mathbf{z}_{\mathbf{x} = \mathbf{x}_T} \, ; \, \Omega \big) \;\; \text{where} \;\; \mathbf{x}_T \sim p_{\text{target}}(\mathbf{x})
\label{eq:trajectory_target}
\end{equation}
\end{definitionbox}

\begin{figure}[ht]
    \centering
    \includegraphics[width=0.75\linewidth]{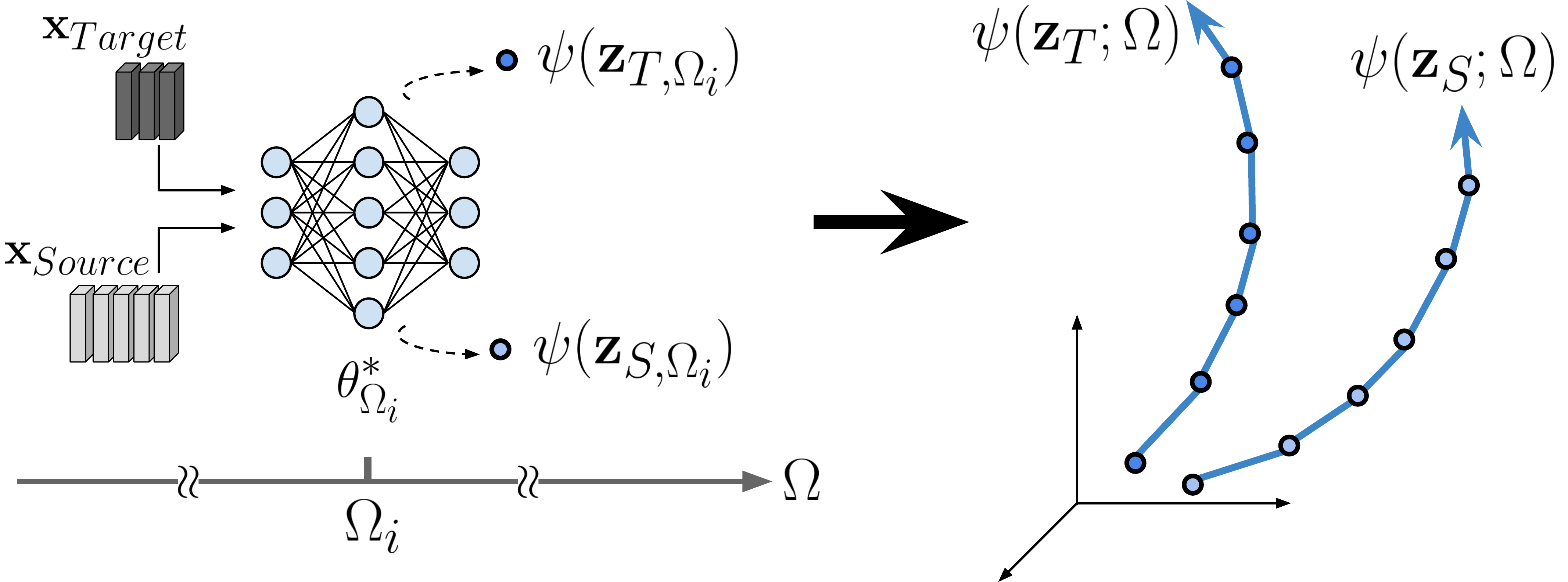}
    \caption{(component 4) For a model with (training) source data $\mathbf{x}_{source}$ and to be deployed on a downstream or out-of-distribution task with target data $\mathbf{x}_{target}$, and given access to a few unlabeled examples from $\mathbf{x}_{target}$, we perform a contrastive analysis of the target activation trajectory $\psi(\mathbf{z}_T; \Omega)$, by comparing it with the source activation trajectory $\psi(\mathbf{z}_S; \Omega)$.
    }
    \label{fig:neural_coherence:concept-4}
\end{figure}

\paragraph{\nameNeuralCoherencePart 5: Coherence of trajectory directions}
Finally, having established that we need to look contrastingly at the neural activation trajectories of the target and source, here we specify how this contrastive analysis should be done. Concretely, we analyze how similar are the directions of the target and source activation trajectories. 

For the sequence of models obtained by varying the hyperparameters, given the source and target input distributions, we compare the source and target neural activation trajectories.
Within the sequence of hyperparameter (set) values, we seek to identify the value $\Omega^*$ which yields the model with the best performance for the target task, among this sequence of models.
The empirical loss with respect to the source task can be estimated, as there is ample available data from the source task.
Given a hyperparameter to tune, the sequence of trained models is precisely obtained by incrementing (or decrementing) the hyperparameter so as to observe a monotonic decrease in the trend of empirical loss over the source task.
For now, to simplify the presentation, we will assume that, for a given sequence of models, the optimum model for the target task will not appear before the optimum for the source task.
We will later explain how Neural Coherence generalizes outside of this context.
Thus, in the present context, as long as the source loss decreases, the target loss is assumed to decrease as well, until reaching its optimum.
After that point, assuming the general trend of the loss is convex (as mentioned earlier), the target loss would start to increase, while the source loss keeps decreasing.
We can thus note that, before the target optimum, i.e. $\Omega < \Omega^*$, the source and target loss functions both decrease with respect to $\Omega$, and are therefore coherent, or positively correlated, i.e. $\text{Corr}(\mathcal{L}_{S}, \mathcal{L}_{T}) > 0$. 
We refer to this interval $\Omega < \Omega^*$ as the ``coherent phase''. After the target optimum, the target loss increases while the source loss decreases. Within that interval $\Omega > \Omega^*$, the two loss functions are therefore divergent, or negatively correlated, i.e. $\text{Corr}(\mathcal{L}_{S}, \mathcal{L}_{T}) < 0$.
We refer to the interval $\Omega > \Omega^*$ as the ``divergent phase''. Based on this insight, we define, for two trajectories, a \textit{coherence} metric, which measures how similar their global directions are, in a given interval over $\Omega$ :

\begin{definitionbox}[Neural Coherence]
Given two activation trajectories $\psi(\mathbf{z}_S; \Omega)$ and $\psi(\mathbf{z}_T; \Omega)$ over a hyperparameter interval $[\Omega_i, \Omega_j]$, we define their Neural Coherence as:
\begin{equation}
\mathrm{NC}(\mathbf{z}_S, \mathbf{z}_T; \Omega_i, \Omega_j) \overset{\underset{\mathrm{def}}{}}{=} d \Big( \psi(\mathbf{z}_S; \Omega) \,,\, \psi(\mathbf{z}_T; \Omega) \Big) 
\label{eq:neural_coherence}
\end{equation}
where $\Omega \in [\Omega_i, \Omega_j]$ and $0 \leq i < j \leq \tau$, and where $d : \mathbb{R}^{2 \times \tau \times M} \mapsto [-1, 1]$ is a directional similarity measure between two trajectories (of length $|j - i|$), such that $\; \mathrm{NC}>0 \;$ indicates ``coherence'' between the trajectories, whereas $\; \mathrm{NC}<0 \;$ indicates ``divergence''.
\end{definitionbox}

Having defined the notion of Neural Coherence in Eq.\ref{eq:neural_coherence}, we use it to infer $\Omega^*$, the value for the hyperparameter $\Omega$ yielding the trained model achieving the lowest target loss $\mathcal{L}^*_T$.
The core intuition being that, before $\Omega^*$, the source and target losses curves $\mathcal{L}_S(\Omega)$ and $\mathcal{L}_T(\Omega)$ remain coherent, but diverge from each other after $\Omega^*$, and this should be reflected by their activation trajectories $\psi(\mathbf{z}_S; \Omega)$ and $\psi(\mathbf{z}_T; \Omega)$.
We thus infer $\Omega^*$ as the point between $\Omega_0$ and $\Omega_{\tau}$ that maximizes the difference between the coherent phase on the left, and the divergent phase on the right (See Eq.\ref{eq:neural_coherence:decision}). Note that if no divergence occurs, then $\Omega^* = \Omega_{\tau}$. With this inferred optimal hyperparameter(s), we get the inferred parameter vector $\theta^*_{\Omega^*} \doteq \mathcal{O} ( f, \theta, \Omega = \Omega^*)$ and its model $f^* = f(\mathbf{x}; \theta = \theta^*_{\Omega^*})$ is selected and  deployed on the target task.
\begin{equation}
    \Omega^* = \underset{\Omega_i \in [\Omega_0, \Omega_{\tau}]}{\operatorname*{arg\,max}} \Big[ \underset{\text{coherent phase}}{\underbrace{\mathrm{NC}\big(\mathbf{z}_S, \mathbf{z}_T; \Omega_0, \Omega_i \big)}} - \underset{\text{divergent phase}}{\underbrace{\mathrm{NC}\big(\mathbf{z}_S, \mathbf{z}_T; \Omega_i, \Omega_{\tau}\big)}} \Big]
    \label{eq:neural_coherence:decision}
\end{equation}
\begin{figure}[ht]
    \centering
    \includegraphics[width=0.75\linewidth]{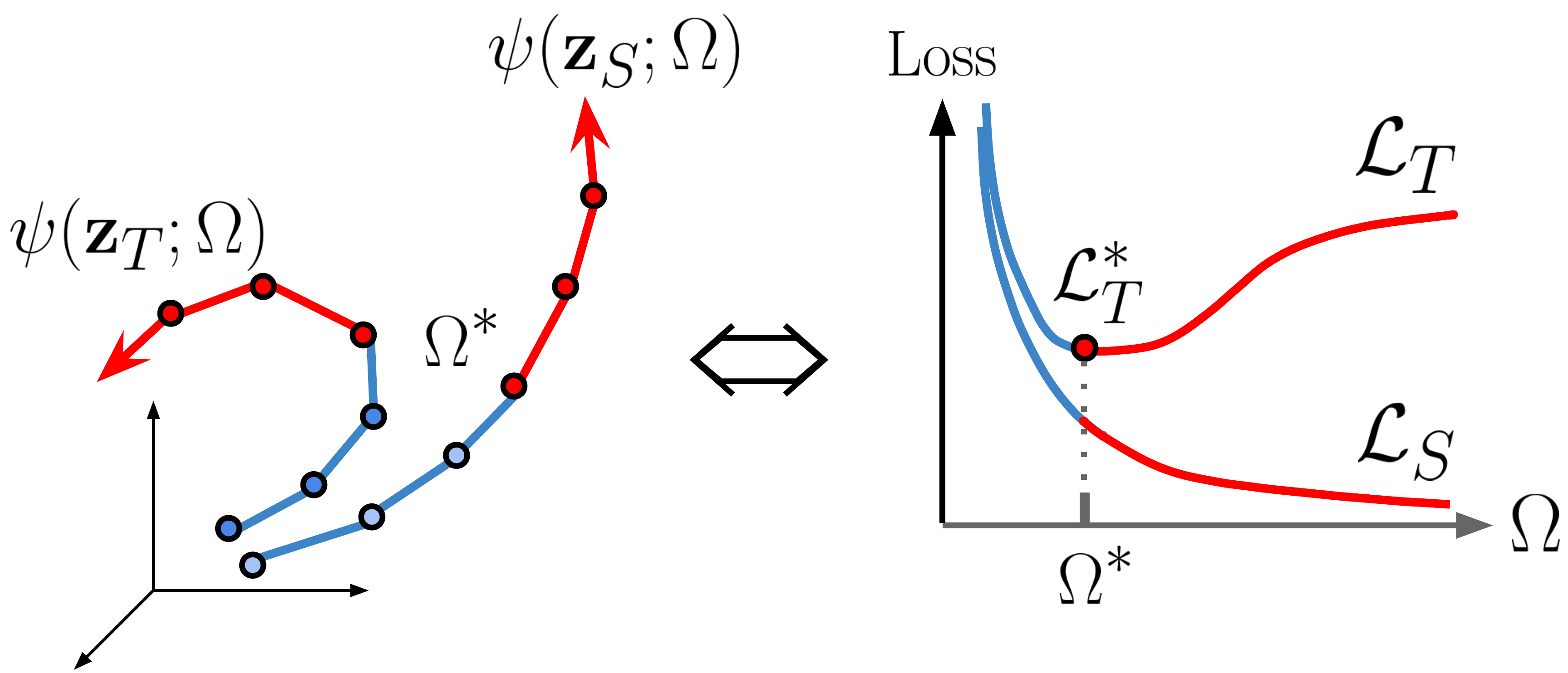}
    \caption{(component 5) The value for the hyperparameter $\Omega$ yielding optimal target loss is inferred and selected as the value $\Omega^*$, before which the target and source activation trajectories remain coherent, but after which they become divergent, i.e. they go in different directions according to a metric $d$. The model trained with $\Omega^*$ is inferred as giving the optimal target loss $\mathcal{L}^*_T$ for the target task. The intuition: before $\Omega^*$, the source and target losses curves $\mathcal{L}_S(\Omega)$ and $\mathcal{L}_T(\Omega)$ remain coherent, but diverge after $\Omega^*$, and this should be reflected by their activation trajectories $\psi(\mathbf{z}_S; \Omega)$ and $\psi(\mathbf{z}_T; \Omega)$.
    }
    \label{fig:neural_coherence:concept-5}
\end{figure}
We can generalize to scenarios where the order of a hyperparameter $\Omega$ isn't as natural as training time $t$, where it is unclear which value represents $\Omega_0$ and which represents $\Omega_{\tau}$. Here the source optimum $\Omega^*_{valid}$ still lies somewhere on the interval of $\Omega$, but the target optimum may lie on either side of it.
To do so, we compute Neural Coherence on the interval going from $\Omega_0$ to $\Omega^*_{valid}$, and then, on the interval going from $\Omega_{\tau}$ to $\Omega^*_{valid}$. 
Which interval between $[\Omega_0, \Omega^*_{valid}]$ and $[\Omega_{\tau}, \Omega^*_{valid}]$ yields the highest coherence is interpreted as leading to the target loss decreasing while the source loss decreases. Consequently, the other interval, with lower coherence and thus higher divergence, is interpreted as containing the target optimum $\Omega^*$. This optimum is then identified as the separation point between the coherent and divergent phase of that interval, just as we've defined in Eq.\ref{eq:neural_coherence:decision}.
\\
\\
Neural Coherence analyzes how neural activation distributions evolve with respect to $\Omega$, and our fifth component focuses specifically on the \emph{direction} of these trajectories. We measure directional similarity between source and target trajectories, rather than their point-wise distance, and experimentally we find this to be much more predictive of target generalization. In particular, target performance typically peaks right when directional similarity collapses—marking a transition from a coherent to a divergent phase. This mirrors the behavior of source and target loss curves: the absolute loss gap between source and target can shrink or grow without indicating the target optimum; instead, the optimum occurs when their “directions” diverge, i.e., the source loss continues to decrease while the target loss starts to increase.

\section{Related Work} \label{sec:related_works}
The problem of model selection for out-of-distribution domains, including applications such as checkpoint selection (early-stopping), has been examined in works such as those of \cite{Goodfellow-et-al-2016, 2020arXiv200701434G, fisch2023towards}. Many subfields of machine learning, such as meta-learning, typically aim at developing optimizers or models capable of easy and efficient adaptation to new tasks \citep{DBLP:journals/corr/FinnAL17, prototypical, DBLP:journals/corr/VinyalsBLKW16}.
There are three widely used approaches when estimating the performance of a given pre-training setup on a target dataset.
The first is to assume that in-distribution validation performance (Source-Val) is well correlated with the target domain, but this is not guaranteed.
The second option is to validate the performance on the target domain (Target-Val), which depends on the presence of sufficient target data and labels.
A third approach is to use heuristics, which may allow for better model selection in scenarios where the aforementioned methods would provide suboptimal results.

A common strategy for assessing the efficacy of models is to rely on an in-distribution validation set (Source-Val) to stop the pre-training early or select a checkpoint for fine-tuning \citep{greyscale_imagenet, chinchilla,zhang2021optimization,bonet2021channel,yao2022metalearning}. These methods have the major limitation that they are not informed by knowledge about the target domain/task, thus potentially suffering from a generalization gap. Indeed, \cite{2020arXiv200701434G} also argue that model selection methods for Out-of-Distribution and domain generalization must be informed, in some capacity, by the actual downstream domain, if they are to be effective.
Different works have proposed heuristics for estimating model generalization to out-of-distribution domains and performing model selection. They often revolve around the idea of measuring a specific metric on the model embeddings in representation space. Some are independent of the target domain, but such methods, just like source validation, can be suboptimal due to the potential distributional shift between the source and target domains.
Other approaches consider the target domain \citep{nce, Guiroy2020}.  We found that most such heuristics tend to use some of the five core elements that constitute Neural Coherence, such as analyzing model representations, inspecting activations across the entire network, or considering a broader family of statistical distances. However, to the best of our knowledge, none have used all those elements in conjunction, particularly not the use of trajectories of model states nor the concept of trajectory coherence, since scalar distance functions are predominant \citep{parc, sun2016deep, dwivedi2020duality, dwivedi2019representation, kornblith2019similarity, raghu2017svcca}.

Other works have studied the relationship between OOD generalization and latent dynamics in neural networks. \cite{DBLP:journals/corr/YosinskiCBL14} show that the ability of neural networks for knowledge transfer can be related to important intermediate hidden layers. In a similar vein, \cite{spectral_analysis} and \cite{featurespace_saturation} have demonstrated that shifts in the data distribution can cause measurable inefficiencies in hidden layers. Other works study OOD generalization in the context of meta-learning or few-shot learning. \citep{2019arXiv190909157R} observed that when a neural network is fine-tuned on a new task, while the linear classification head changes drastically, the backbone remains approximately invariant. In  \cite{DBLP:journals/corr/abs-2002-06753}, the authors observed that generalization might be related to a notion of representation clustering. On the other hand, \cite{DBLP:journals/corr/abs-1909-02729} and \cite{2019arXiv190201889F} observed that, in OOD settings, the clustering of final representations might not be indicative of target generalization.

Other works have analyzed theoretical aspects of gradient-based Meta-Learning. In particular, \cite{DBLP:journals/corr/abs-1902-08438} provided a theoretical upper bound for the regret of MAML in an online Meta-Learning setting, where an agent faces a sequence of tasks.
Moreover, \cite{denevi2019learning} studied meta-learning from the perspective of biased regularization, where the model adapts to new tasks by starting from a biased parameter vector, which we refer to as the meta-training solution.
When a model learns simple tasks such as linear regression and binary classification with stochastic gradient descent, they theoretically prove the advantage of starting from the meta-training solution, under the assumption that similar tasks lead to similar weight parameterization. 
Considering online convex optimization, where the model learns from a stream of tasks, \cite{DBLP:journals/corr/abs-1902-10644} stated that the optimal solution for each task lies in a small subset of the parameter space. Then, under this assumption, they proposed an algorithm using Reptile, a first-order Meta-Learning method, such that the task-averaged regret scales with the diameter of the aforementioned subset \citep{DBLP:journals/corr/abs-1803-02999}.
More closely related to our approach, \cite{DBLP:journals/corr/abs-1907-07287} empirically studied the objective landscapes of gradient-based Meta-Learning, with a focus on few-shot classification. They notably observed that average generalization to new tasks appears correlated with the average inner product between their gradient vectors. In other words, as gradients appear more similar in the inner product, the model will, on average, better generalize to new tasks, after following a step of gradient descent. Similar findings on the relation between local gradient statistics and generalization have also been previously discussed by \cite{mahsereci2017early}. 

Related to this work is the method of Activation-Based Early-Stopping (ABE), which Neural Coherence generalizes. Experimentally, this work extends ABE by tackling a wider range of experimental settings, transfer learning and out-of-distribution regimes, larger scales for models and datasets. Conceptually, however, Neural Coherence is a generalization of ABE. It defines a family of possible implementations, with ABE being one particular instance. Formalizing Neural Coherence from the five components allows the distinction between this general framework and its possible implementations. Moreover, we discuss and show how Neural Coherence can be applied for selecting other training hyperparameters, such as training data, another fundamental hyperparameter in deep learning. We leave other hyperparameters for future work, providing Neural Coherence as a framework for tackling such quantities as learning rate, bach size, and other regularizations.

As model selection methods, recent work exploits internal activations or unlabeled target data to improve OOD robustness and model selection. Neuron Activation Coverage (NAC) defines density-based coverage scores over neuron states and uses them both for post-hoc OOD detection and source-only model evaluation~\cite{liu2023neuron}, while BLOOD and Discriminability-Driven Channel Selection (DDCS) instead leverage, respectively, between-layer transformation smoothness and channel-wise discriminability across layers for OOD detection~\cite{jelenić2023out0of0distribution,10657753}. ~\cite{harun2024variables} analyze large pretrained vision models and identify architectural and data-related variables that shape representation compression and transfer OOD performance. Complementary to these representation-centric approaches, Average Thresholded Confidence (ATC) and Detection Adaptation Score (DAS) use unlabeled target data to define confidence- and geometry-based criteria for unsupervised model or checkpoint selection under shift~\cite{garg2022leveraging,yu2024towards}. In contrast, our Neural Coherence framework explicitly tracks multi-layer activation trajectories over training and measures directional alignment between source and target dynamics to drive checkpoint and data selection.

In neuroscience, the term ``neuronal coherence'' has been used to refer to a measure of synchronicity of electrical activity between neurons across different parts of the brain \citep{nc2, nc3, nc4, FRIES2005474, WOMELSDORF2006182}. In machine learning, the term has been used in the context of natural language processing, with a very different meaning, in reference to ``text coherence'' \citep{nc1, nc5, nc6, nc7}.

\section{Implementing Neural Coherence} \label{sec:practical_implementation}

Now that we have laid out the general principle and framework for Neural Coherence, we here propose an implementation of it. First, we need to characterize the neural activations distribution $p(\mathbf{z})$.
Because many distributions have been shown to be fully determined by their sequence of moments \citep{2019arXiv191200160Y}, we use the moment sequence to characterize the neural activation distributions. Recall that, for an arbitrary random variable $Z$ with a probability distribution $p(z)$, its n-th order moment is defined as $m_n(z) = \mathbb{E}_{p(z)}[z^n]$. For a multivariate random vector $\mathbf{z}$, its sequence of moments starts with the mean vector $\mathbf{m}_1(\mathbf{z}) = \mathbb{E}_{p(\mathbf{z})}[\mathbf{z}]$, followed by the autocorrelation matrix $\mathbf{m}_2(\mathbf{z}) = \mathbb{E}_{p(\mathbf{z})}[\mathbf{z}\mathbf{z}^{\mathrm{T}}]$, etc. We can write this as: 

\begin{equation} 
\mathbf{m}_1(\mathbf{z}) \;,\; \mathbf{m}_2(\mathbf{z}) \;, ...  \; = \;  \mathbb{E}_{p(\mathbf{z})}[\mathbf{z}] \;,\; \mathbb{E}_{p(\mathbf{z})}[\mathbf{z}\mathbf{z}^{\mathrm{T}}] \;,\; ... \;
= \;
\mathbb{E}
\begin{bmatrix}
z_1 \\
z_2 \\
\vdots \\
z_D \end{bmatrix} ,\;
\mathbb{E}
\begin{bmatrix}
z_1^2 & z_1z_2 & ... & z_1z_D \\
z_2z_1 & z_2^2 & ... & z_2z_D \\
\vdots  & \vdots & \ddots  & \vdots \\
z_Dz_1 & z_Dz_2 & ... & z_D^2 \\
\end{bmatrix} , ...
\nonumber
\end{equation}

For tractability, and because only a few target samples are to be available, and since the estimation error grows with moment order \citep{casella2002statistical,kendallstuart1969advanced,mood1974introduction,fisher1930moments}, we limit sequence estimation to the second order moment, avoiding the excessive standard error of higher order statistics. Moreover, computing batch-wise statistics over the activations, especially the first and second order moments, is common in deep learning, namely with normalization techniques such as Batch-Norm \citep{ioffe2015batch}.
To improve the signal-to-noise ratio, we further aggregate those statistics by computing their moments across the $D$ feature dimensions. This dimensionality reduction also renders our method computationally feasible. Feature-wise statistics over the activations are also commonly used in normalization techniques such as Layer-Norm \citep{ba2016layer}. When computing the feature-wise moments, to avoid introducing higher-order statistics, we only compute the 1st-order feature-wise moment over $\mathbf{m}_2(\mathbf{z})$. To preserve meaningful information, the diagonal and non-diagonal elements of $\mathbf{m}_2$ are aggregated separately. We obtain $\hat{m}_1$, $\hat{m}_2$, $\hat{m}_3$ and $\hat{m}_4$, the \textit{aggregated moments}, defined in Eq.\ref{eq:aggregated_moments}:

\begin{equation} \label{eq:aggregated_moments}
\begin{array}{l@{\hspace{3em}}l}
    \displaystyle \hat{m}_1(\mathbf{z}) = \frac{1}{D} \sum_{i}^D \mathbf{m}_1(\mathbf{z})_i
    & \displaystyle \hat{m}_2(\mathbf{z}) = \frac{1}{D} \sum_{i}^D \mathbf{m}_1(\mathbf{z})_i^2 \\[4ex]
    \displaystyle \hat{m}_3(\mathbf{z}) = \frac{1}{D} \sum_{i}^D \mathbf{m}_2(\mathbf{z})_{i,i}
    & \displaystyle \hat{m}_4(\mathbf{z}) = \frac{1}{D^2 - D} \sum_{i}^D \sum_{j \neq i}^D \mathbf{m}_2(\mathbf{z})_{i,j}
\end{array}
\end{equation}
where $\mathbf{m}_1(\mathbf{z})_i$ is the i-th feature of $\mathbf{m}_1(\mathbf{z})$, and $\mathbf{m}_2(\mathbf{z})_{i,j}$ is the element at the i-th row and j-th column of $\mathbf{m}_2(\mathbf{z})$. 
The function space spanned by the possible linear combinations of $m_1$ to $m_4$, while being simple to compute, is general enough to represent many functions of the activation vectors, such as their average $l_2$-norm, pair-wise inner products, pair-wise $l_2$ distances, sample-wise variance, average feature-wise variance, to name a few.
Our experiments show that target generalization $Acc_{target}(t)$ in few-shot learning is often proportional to a linear combination of the presented aggregated moments when computed on a few unlabeled examples of the target data $p_{Target}(\mathbf{x})$. 
This suggests that there may be multiple factors linked to out-of-distribution generalization of neural networks, and such factors are likely dependent on the experimental setting.
Finally, we characterize the activation distributions at each layer independently, without inter-layer dependencies. This massively reduces the dimensionality of the statistics to compute and the complexity of the patterns to analyze, which is crucial given the scarcity of available target data. We thus characterize the trajectory along $\Omega$ of neural activation distributions as the tensor $\psi(\mathbf{z};\Omega)$ in Eq.\ref{eq:collas:neural_activation_dynamics}:
\begin{equation} 
    \psi(\mathbf{z}; \Omega) =
    \begin{bmatrix}
\hat{m}_1(f_1(\mathbf{x}; \Omega)) & \hat{m}_2(f_1(\mathbf{x}; \Omega)) & \hat{m}_3(f_1(\mathbf{x}; \Omega)) & \hat{m}_4(f_1(\mathbf{x}; \Omega)) \\
\vdots & \vdots & \vdots & \vdots \\
\hat{m}_1(f_L(\mathbf{x}; \Omega)) & \hat{m}_2(f_L(\mathbf{x}; \Omega)) & \hat{m}_3(f_L(\mathbf{x}; \Omega)) & \hat{m}_4(f_L(\mathbf{x}; \Omega)) \\
\end{bmatrix}
\label{eq:collas:neural_activation_dynamics}
\end{equation}

The implementation of the Neural Coherence score is based the correlation of the source and target activation trajectories.
We define the correlation matrix $C$ between the source trajectory $\psi(\mathbf{z}_S; \Omega)$ and target trajectory $\psi(\mathbf{z}_T; \Omega)$, treating $\Omega$ as their  domain. The matrix $C$ is an $L \times K$ matrix whose element $c_{l, k}$ is given in Eq.\ref{eq:element_of_matrix_C} :
\begin{equation}
    c_{l, k} = \underset{\Omega \in [\Omega_0, \Omega_\tau]}{\operatorname*{Corr}} \Big( \psi(\mathbf{z}_S; \Omega)_{l,k} \,,\, \psi(\mathbf{z}_T; \Omega)_{l,k} \Big) = \frac{\mathrm{Cov} \Big( \psi(\mathbf{z}_S; \Omega)_{l,k} \,,\, \psi(\mathbf{z}_T; \Omega)_{l,k} \Big)}{\sqrt{\mathrm{Var}(\psi(\mathbf{z}_S; \Omega)_{l,k})} \; \sqrt{\mathrm{Var}(\psi(\mathbf{z}_T; \Omega)_{l,k})}}
\label{eq:element_of_matrix_C}
\end{equation}
The notion of coherence (or divergence) of two trajectories, over an interval $[\Omega_i, \Omega_j]$, with $0 \leq i < j \leq \tau$, is given by the product of the magnitude of their correlation, its sign, and the interval length (See Eq.\ref{eq:coherence_factor_d}). In essence, these quantities  account for the strength and significance of the observed coherence or divergence:
\begin{equation}
    d_{l,k,i, j} = \left( \frac{j - i}{\tau} \right) \cdot c_{l, k} \;\;\;\text{with}\;\;\; \Omega_i \leq \Omega \leq \Omega_j
    \label{eq:coherence_factor_d}
\end{equation}
As explained earlier, we seek a separation point $i$ between a strong coherent phase and a strong divergent phase. In other words, this point $i$ is the value that maximizes the difference between the Neural Coherence prior to itself, and the Neural Coherence past itself. For a given layer $l$ and moment $k$, this point would be 
$i^* = \mathrm{argmax}_i ( d_{l,k,0, i} - d_{l,k,i, \tau})$.
For the definitive function $d$ over the entire set of trajectory pairs, to be used in the decision function Eq.\ref{eq:neural_coherence:decision} to find $\Omega^*$, multiple options are possible. 
For small neural networks, we observed that this separation tends to be dominated by a single layer $f_l$ and moment $\hat{m}_k$. Hence, we simply pick the layer $l^*$ and moment $k^*$ with the strongest difference in Neural Coherence, and the optimal value $\Omega^*$ is where this difference is maximized at the layer $l^*$ and moment $k^*$, as defined in Eq.\ref{eq:omega_prime:unweighted} :
\begin{equation}
    \Omega^* = \Omega_{i^*} \Leftarrow  l^*, k^*, i^* = \underset{l,k, i}{\operatorname*{arg\,max}} ( d_{l,k,0, i} - d_{l,k,i, \tau} )
    \label{eq:omega_prime:unweighted}
\end{equation}

\paragraph{Neural Coherence Based Checkpoint Selection} \label{sec:practical_implementation:checkpoint_selection}
Here we present how we apply Neural Coherence for checkpoint selection through training time $t$, also called early-stopping, where $\Omega = t$. 
We first make the assumption that the optimal early-stopping time for the target task does not happen after the optimal stopping time for the source task, as our experiments have suggested, such that $t^* = \mathrm{argmin}_t \, \mathcal{L}_{T} \leq \mathrm{argmin}_t \, \mathcal{L}_{S}$.
For architectures with four or fewer layers, \cite{DBLP:conf/collas/GuiroyPMC22} have shown that simply focusing  on the layer with the strongest divergence can be sufficient for determining the early stopping time, and $t^*$ is obtained from Eq.\ref{eq:omega_prime:unweighted}.
While focusing on the strongest divergence has been demonstrated to work in small-scale neural networks with few hidden layers, we find it insufficient when working with very deep models comprising hundreds of layers, such as ConvNeXt. We found early-stopping decisions could be unreliable. Moreover, in very deep models, multiple layers might show significant divergence. Thus the question arises of how to weigh their contributions to make the early-stopping decision.
We propose a weighting function of the contributions of each layer, based on their individual divergence scores, where layers showing strong divergence have more influence over the final early-stopping decision $t^*$, while layers showing little divergence have little influence.
For each layer $l$, the stopping time $t^*_l$ is given by Eq.\ref{eq:time_per_layer} and the strength $\mathrm{NC}^{div}_l$ of their divergent phase is given by Eq.\ref{eq:NC_per_layer}, where $0 \leq \mathrm{NC}^{div}_l \leq 1$:
\begin{equation}
    t^*_l \Leftarrow \{ t^*_l, k^*_l \} = \underset{t, k}{\operatorname*{arg\,max}} \; \Big[ \mathrm{NC}(\mathbf{z}_S, \mathbf{z}_T; t_0, t)_{l,k} - \mathrm{NC}(\mathbf{z}_S, \mathbf{z}_T; t, t_{\tau})_{l,k} \Big]
\label{eq:time_per_layer}
\end{equation}
\begin{minipage}{0.7\linewidth}
\begin{equation}
    \mathrm{NC}^{div}_l = \underset{t, k}{\operatorname*{max}} \; \Big[ 
    - \mathrm{NC}(\mathbf{z}_S, \mathbf{z}_T; t, t_{\tau})_{l,k} \Big]
\label{eq:NC_per_layer}
\end{equation}
\end{minipage}
\hfill
\begin{minipage}{0.3\linewidth}
\begin{equation}
    \alpha_l = \frac{\mathrm{NC}^{div}_l}{\sum_{l=1}^L \mathrm{NC}^{div}_l}
    \label{eq:NC-weight_per_layer}
\end{equation}
\end{minipage}
We therefore weight the contribution of a given layer from the strength of its divergence. These weighting scores defined in Eq.\ref{eq:NC-weight_per_layer} are  normalized so that $\sum_l \alpha_l = 1$, and the resulting stopping time $t^*$ is given in Eq.\ref{eq:weighted_NC}.

\begin{equation}
    t^* = \sum_{l=1}^L \alpha_l t^*_l
    \label{eq:weighted_NC}
\end{equation}
We note that this final score is biased towards $t^*_{valid}$. which is normal and intended, since when layers show weak or no divergence, their resulting early-stopping time $t^*_l$ will be close to the last allowed iteration. See ablation study of Fig. \ref{fig:empirical_demos:weighting_neural_coherence} for a demonstration of the effect of normalizing the neural coherence scores. Later in Sec. \ref{sec:exps:data_selection} we will demonstrate how to implement Neural Coherence for the problem of training data selection.

\section{Experiments on Model Checkpoint Selection} \label{sec:exps:ckpt_selection}

This section, "Experiments on Model Checkpoint Selection," presents the empirical evaluation of Neural Coherence (NC) for model selection, primarily focusing on its application in determining the optimal model checkpoint during training. We begin by exploring its effectiveness in smaller-scale settings before demonstrating its capabilities with larger foundation models. This section covers:

\textbf{Section \ref{sec:exps:ckpt_selection:meta-learning} --} \textit{Meta-Learning for Few-Shot Classification:} 
This subsection provides initial motivation for Neural Coherence by detailing experiments in the few-shot learning paradigm, a relevant setting for studying out-of-distribution (OOD) generalization. It showcases how our method, using only a few unlabeled examples from a single target task, significantly outperforms validation-based early-stopping in improving target generalization across various meta-learning algorithms (MAML, Prototypical Network, Matching Network) and for several source-target dataset pairs.

\textbf{Section \ref{sec:exps:ckpt_selection:large_vision_models} --} 
\textit{Large vision models:} 
Here, we extend our experimental validation to state-of-the-art large vision models, such as ConvNeXt-L and Vision Transformer (ViT) architectures. We demonstrate the performance of Neural Coherence in two different OOD paradigms: zero-shot generalization and transfer learning under extreme sparsity of target data, consistently outperforming various baselines even with as few as 5 unlabeled target samples.

\textbf{Section \ref{sec:exps:ckpt_selection:statistical_efficiency} --} 
\textit{Statistical Efficiency of Neural Coherence:} 
Finally, this subsection highlights a key advantage of NC by demonstrating its statistical efficiency, showing that it consistently requires fewer target examples than other baselines to achieve a comparable level of downstream performance. Our results indicate that NC, even with a single unlabeled sample, can outperform methods that use up to 20 labeled samples.

\subsection{Meta-Learning for Few-Shot Classification}
\label{sec:exps:ckpt_selection:meta-learning}

In this section, we adopt the few-shot learning setup and meta-learning algorithms to experimentally evaluate and demonstrate our principle for inferring out-of-distribution (OOD) performance of deep learning models. Few-shot learning inherently tests a model's ability to generalize from limited data, a scenario closely tied to the challenges of OOD inference, where training and testing distributions differ significantly. Meta-learning, with its focus on learning-to-learn across tasks, provides a robust framework for evaluating adaptability and generalization. This combination allows us to effectively analyze how well our proposed principle captures and predicts model performance in scenarios requiring extrapolation beyond the training distribution.
Important practical progress has been made in meta-learning over the years \citep{Lake2015HumanlevelCL,Koch2015SiameseNN,DBLP:journals/corr/AndrychowiczDGH16,Rezende:2016:OGD:3045390.3045551,pmlr-v48-santoro16,DBLP:journals/corr/MishraRCA17,prototypical,DBLP:journals/corr/VinyalsBLKW16,DBLP:journals/corr/abs-1711-06025,2017arXiv171104043G,DBLP:journals/corr/FinnAL17,DBLP:journals/corr/abs-1810-03642,DBLP:journals/corr/abs-1904-03758,sachin,DBLP:journals/corr/WichrowskaMHCDF17,2018arXiv180400222M,2019arXiv190203356P,Maclaurin:2015:GHO:3045118.3045343,2018arXiv180604910F,Lian2020Towards}. Yet, understanding the underlying phenomena behind the transitioning of a neural network's generalization to novel tasks, from the underfitting to the overfitting regime, has remained an open research question.

We now present experimental results of the performance of our early-stopping proposed method based on Neural Coherence. For each experiment, tasks are 5-way 1-shot classification, and we only use the unlabelled input examples from the support set of a single target task. In other words, only 5 unlabelled images of a single target task are used to analyze the neural activation dynamics. At the beginning of an experiment, we randomly sample a task $\mathcal{T}_i$ from $p(\mathcal{T}_{target})$ and keep only its set of support input examples, which we use for early-stopping, and evaluate the resulting target accuracy (performance). We repeat this for 50 independently and identically distributed support sets from $p(\mathcal{T}_{target})$, and report the average performance. This is then repeated for 3 independent training runs. As the baseline for direct comparison, we use the common early-stopping approach based on performance on the validation set. For each experiment, the validation accuracy and target accuracy, for each checkpoint, are both averaged over 600 validation and target tasks respectively. Note that the examples used to measure the neural activations and perform early-stopping are \textit{not used} in the evaluation of generalization. To benchmark our method, we use the standard 4-layer convolutional neural network architecture described by \cite{DBLP:journals/corr/VinyalsBLKW16} trained on 3 different Meta-Learning algorithms: MAML \citep{DBLP:journals/corr/FinnAL17}, Prototypical Network \citep{prototypical}, and Matching Network \citep{DBLP:journals/corr/VinyalsBLKW16}. Our source and target datasets comprise Mini-Imagenet \citep{DBLP:journals/corr/VinyalsBLKW16}, Omniglot \citep{42e58b2b32904ebdbb4581c0bd34691d}, Quickdraw \citep{DBLP:journals/corr/HaE17}, CUB Birds \citep{fg6w-vh29-25}, VGG Flower \citep{vgg_flower}, Traffic Sign \citep{traffic_sign}, Aircraft \citep{DBLP:journals/corr/MajiRKBV13} and Imagenet \citep{russakovsky2015imagenet}. These datasets are also used in the Meta-Dataset \citep{Triantafillou2020Meta-Dataset}. We refer to Sec.~\ref{sec:app:exp_detail} for additional experimental details.

\begin{figure}
    \centering
    \subfloat[MAML]{
        \includegraphics[width=0.33\linewidth]{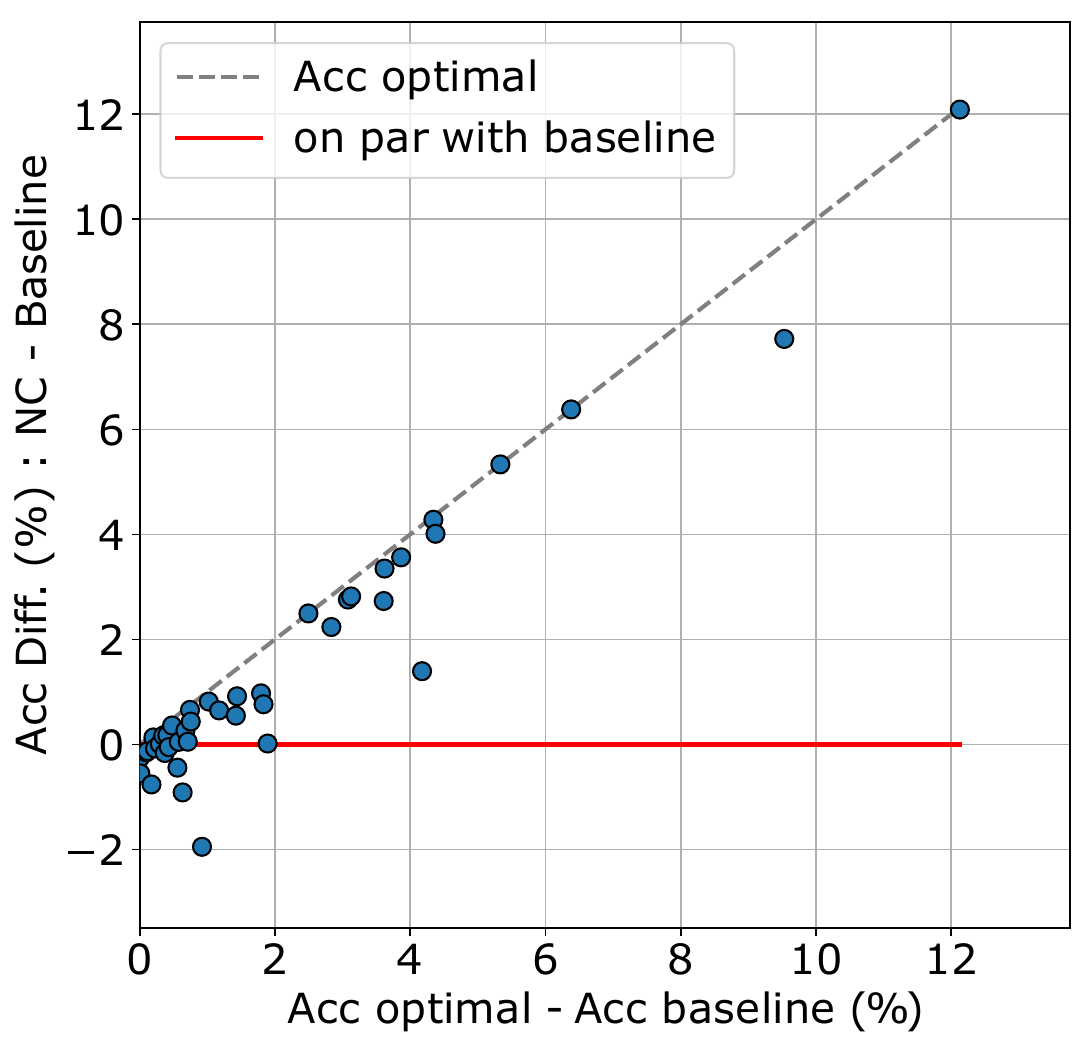}
        \label{fig:scatter:our_method-baseline:maml}}
    \subfloat[Prototypical Network]{%
\includegraphics[width=0.33\linewidth]{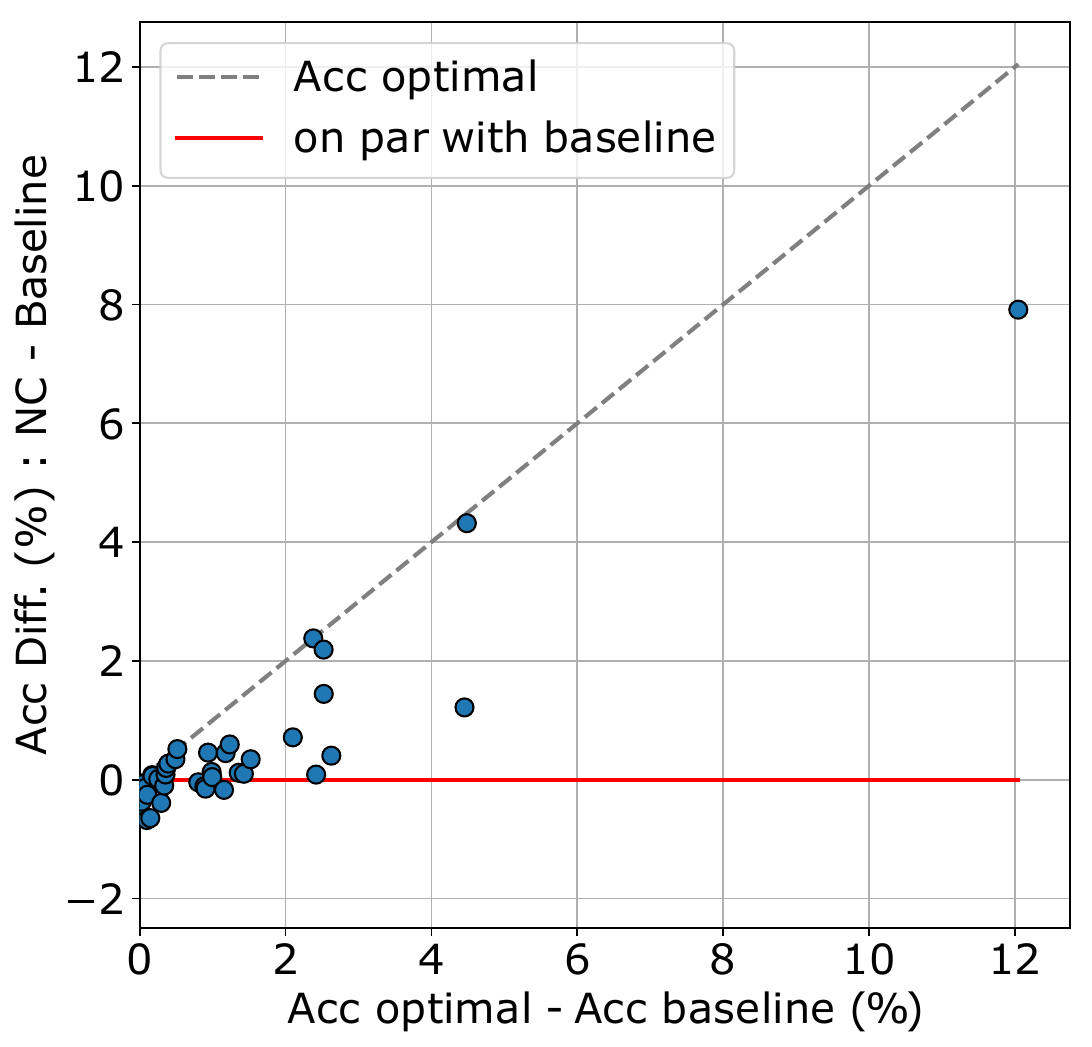}
        \label{fig:scatter:our_method-baseline:prototypical}}
    \subfloat[Matching Network]{%
        \includegraphics[width=0.33\linewidth]{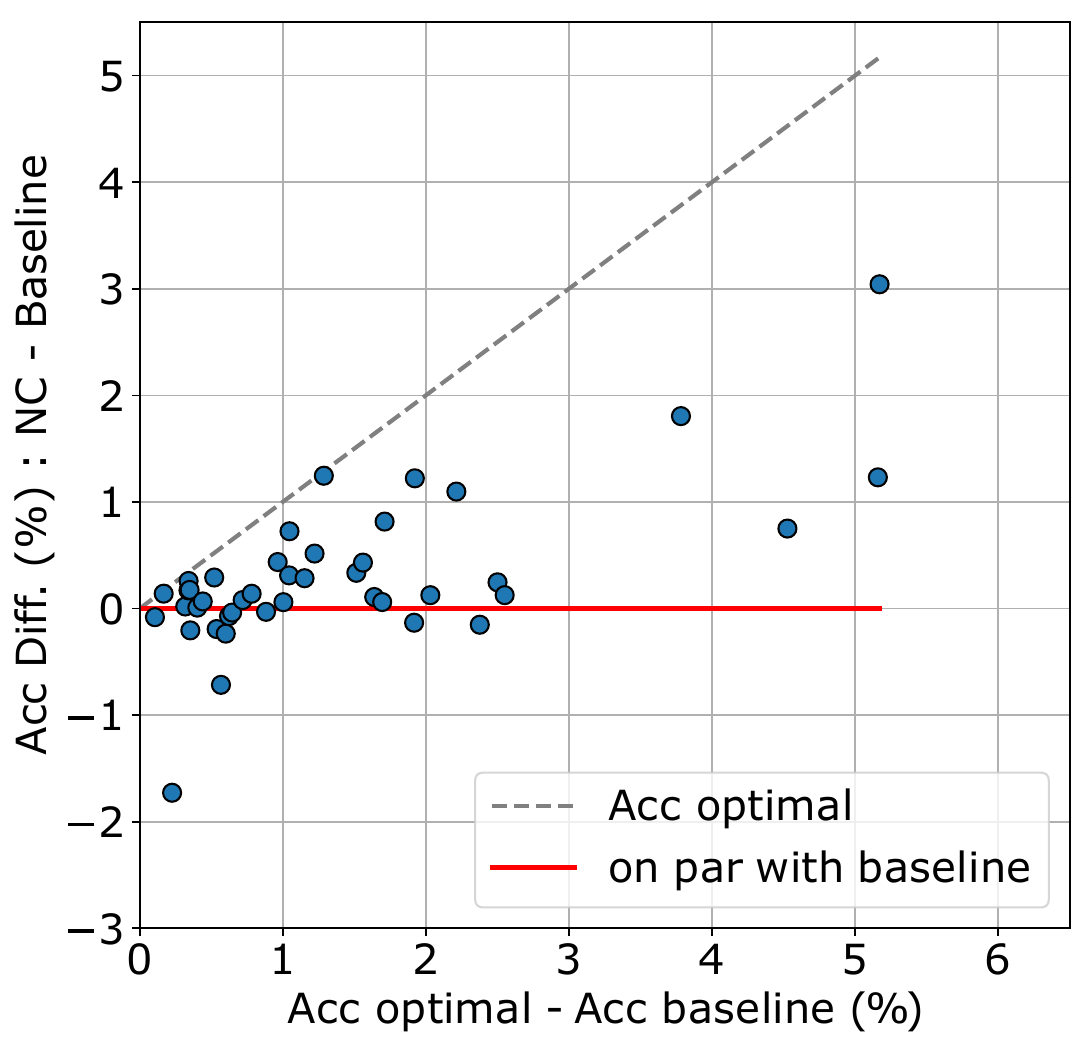}
        \label{fig:scatter:our_method-baseline:matching}}
    \caption{Gain in target accuracy from our early-stopping method (Neural Coherence), compared to validation-based early-stopping (baseline). The performance gain (y-axis) is shown against the baseline generalization gap (x-axis). For each experiment, we plot the accuracy difference between our method and the baseline. The red horizontal bar indicates \textit{on par} with the baseline (no difference). The grey dashed line indicates the max achievable performance (completely filling the baseline's generalization gap). Results show that our method generally outperforms the baseline. }\label{fig:scatter:our_method-baseline}
\end{figure}

Across our experiments, our results show that our early-stopping method based on Neural Coherence outperforms validation-based early-stopping. On average, our method closes 47.8\% of the generalization gap between the validation baseline and optimal performance (oracle). Notably, when the baseline generalization gap is larger (baseline performance is $<$ 0.95 $\times$ oracle accuracy), we close the generalization gap by 69.2\%. Moreover, when this gap is already small (baseline performance $\geq$ 0.95 $\times$  oracle accuracy), we still manage to slightly outperform the validation baseline, closing the gap by 14.5\% on average. More detailed experimental discussions are provided below. 
In Fig.~\ref{fig:scatter:our_method-baseline} we show the gain in target accuracy from our early-stopping method (Neural Coherence), compared to validation-based early-stopping (baseline). We also show the oracle performance, \textit{i.e.} the performance obtained if early-stopping had been optimal in each experiment. For each setting (a tuple of algorithm, source dataset, and target dataset), we show the accuracy difference against the generalization gap exhibited by the validation-based early-stopping. Results show that our method outperforms the baseline. When the generalization gap of the baseline is small and there is not much gain to have, our method is generally on par with validation-based early-stopping, and only in a very few instances do we suffer from a slight generalization drop. As the baseline generalization gap increases, our method tends to fill this gap. This results in better performance on the target task distribution as observed across all tested Meta-Learning algorithms. In Tab.~\ref{tab:performance} we show the average generalization performance of Neural Coherence based early-stopping, compared against validation-based early-stopping, and the oracle performance. Results indicate that Neural Coherence outperforms validation-based early-stopping, closing the generalization gap towards oracle performance. We break the results into two categories : performance per source dataset, averaged over all Meta-Learning algorithms and target datasets (Tab.~\ref{tab:performance:source_datasets}); performance per Meta-Learning algorithm (Tab.~\ref{tab:performance:algorithms}).

\renewcommand{\tabcolsep}{3pt}
\begin{table}
    \centering
    \begin{minipage}[c]{0.63\textwidth}
    \subfloat[Performance per source dataset]{
    \resizebox{\textwidth}{!}{%
        \begin{tabular}{|c|cccccc|}
\hline
\multirow{2}{*}{Acc. (\%)} & \multicolumn{6}{c|}{Source dataset} \\ \cline{2-7} 
 & \multicolumn{1}{c|}{cu\_birds} & \multicolumn{1}{c|}{aircraft} & \multicolumn{1}{c|}{omniglot} & \multicolumn{1}{c|}{vgg\_flower} & \multicolumn{1}{c|}{mini-imagenet} & quickdraw \\ \hline
oracle & \multicolumn{1}{c|}{43.74} & \multicolumn{1}{c|}{40.34} & \multicolumn{1}{c|}{34.27} & \multicolumn{1}{c|}{40.55} & \multicolumn{1}{c|}{47.21} & 38.74 \\ \hline \hline
Baseline & \multicolumn{1}{c|}{41.75} & \multicolumn{1}{c|}{38.71} & \multicolumn{1}{c|}{33.26} & \multicolumn{1}{c|}{39.59} & \multicolumn{1}{c|}{45.71} & 36.31 \\ \hline
\textbf{\begin{tabular}[c]{@{}c@{}}Neural Coherence (ours)\end{tabular}} & \multicolumn{1}{c|}{\textbf{43.11}} & \multicolumn{1}{c|}{\textbf{39.27}} & \multicolumn{1}{c|}{\textbf{33.77}} & \multicolumn{1}{c|}{\textbf{40.09}} & \multicolumn{1}{c|}{\textbf{46.20}} & \textbf{37.44} \\ \hline
\end{tabular}}%
        \label{tab:performance:source_datasets}
        }
    \end{minipage}
    \begin{minipage}[c]{0.36\textwidth}
    \subfloat[Perf. per Meta-Learning algorithm]{
    \resizebox{\textwidth}{!}{%
\begin{tabular}{|cccc|}
\hline
\multicolumn{1}{|c|}{\multirow{2}{*}{Acc. (\%)}} & \multicolumn{3}{c|}{Meta-learning algorithm} \\ \cline{2-4} 
\multicolumn{1}{|c|}{} & \multicolumn{1}{c|}{MAML} & \multicolumn{1}{c|}{ProtoNet} & Matching Net \\ \hline
\multicolumn{1}{|c|}{oracle} & \multicolumn{1}{c|}{41.11} & \multicolumn{1}{c|}{38.13} & 43.19 \\ \hline \hline
\multicolumn{1}{|c|}{Baseline} & \multicolumn{1}{c|}{39.01} & \multicolumn{1}{c|}{36.87} & 41.78 \\ \hline
\multicolumn{1}{|c|}{\textbf{\begin{tabular}[c]{@{}c@{}}Neural Coherence (ours)\end{tabular}}} & \multicolumn{1}{c|}{\textbf{40.51}} & \multicolumn{1}{c|}{\textbf{37.35}} & \textbf{42.09} \\ \hline
\end{tabular}%
}\label{tab:performance:algorithms}
}
    \end{minipage}
    \caption{Generalization performance of early-stopping based on Neural Coherence, compared to validation-based early-stopping (Baseline). For each setting, ``oracle" is the maximum achievable generalization performance, if early-stopping had been optimal (with an oracle). We show results per Meta-Learning algorithm (averaged over all source and target datasets), and per source dataset (averaged over all Meta-Learning algorithms and target datasets). The first row shows the maximum achievable target accuracy. Results show that for each source dataset and Meta-Learning algorithm, our method consistently outperforms the validation baseline. To measure the variability in performance of our method, for each experiment (algo, source dataset, target dataset), we compute the standard deviation among the generalization performances obtained with the 50 different target tasks (independent trials). We then average those standard deviations across all experiments. We present them on a per-algorithm basis. Those standard deviations are (in percentages of accuracy): 0.35\% for MAML,  0.13\% for Prototypical Network, and 0.38\% for Matching Network.}
    \label{tab:performance}
\end{table}

\begin{figure}[ht]
    \centering
    \includegraphics[width=0.99\linewidth]{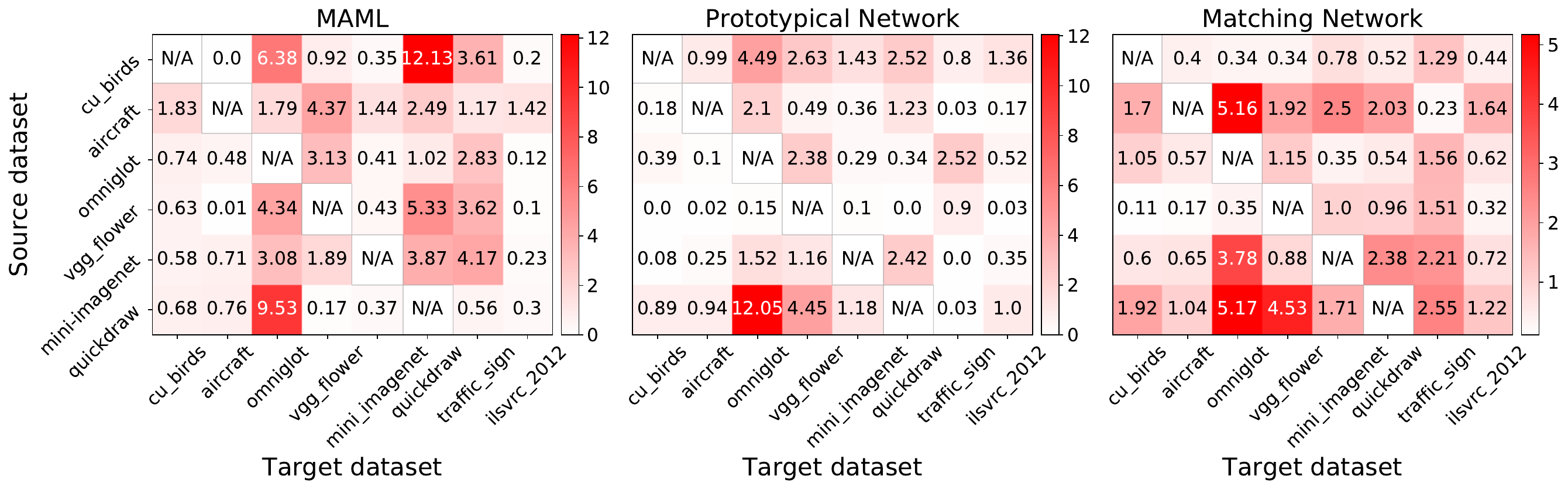}
    \\
    \includegraphics[width=0.99\linewidth]{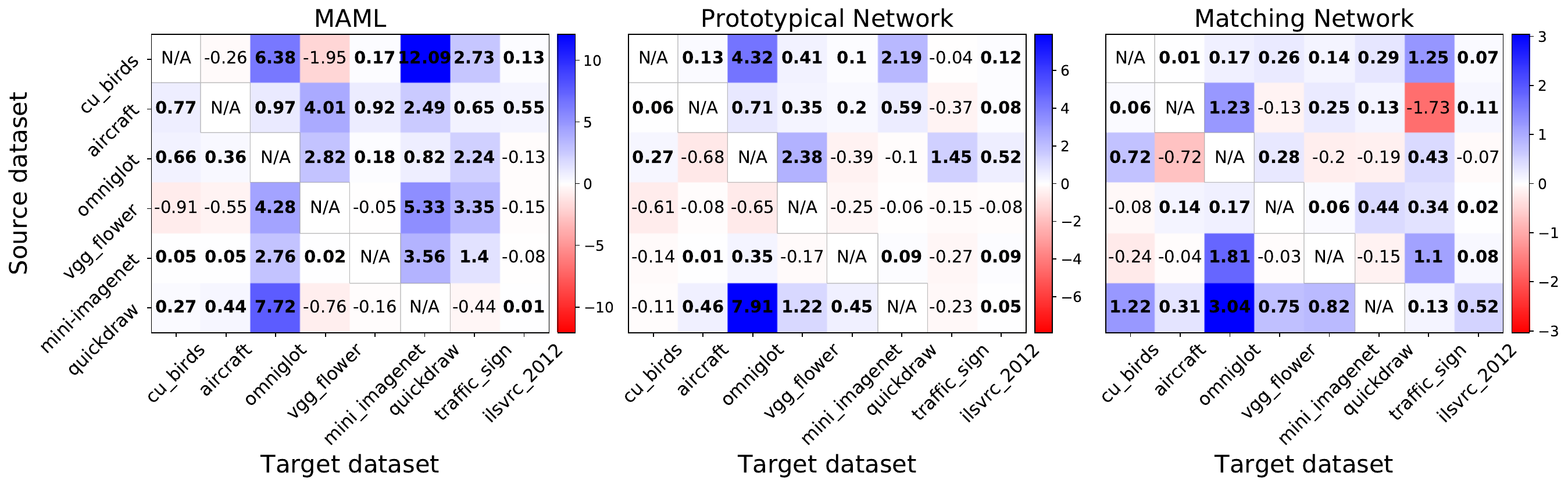}
    \caption{\textbf{(top) Degradation in target performance (red) using validation-based early-stopping, with respect to the maximum possible accuracy} -- we refer to this as the generalization gap. For each experiment, we compute the target accuracy obtained when early-stopping with the validation baseline, and the actual maximum target accuracy that was achievable during the experiment, if early-stopping had been optimal, and we compute the difference.
    \textbf{(bottom) Improvement in generalization performance (blue) of Neural Coherence (our method) compared to the validation baseline, versus degradation (red).} Difference in generalization performance between Neural Coherence and the baseline of validation early-stopping. For each experiment, we compute the target accuracy obtained when early-stopping with the validation baseline, and target accuracy obtained when early-stopping with the validation baseline, and we compute the difference.}
    \label{fig:baseline_gap_heatmap:horizontal}
\end{figure}

In the \textit{top row} of Fig.~\ref{fig:baseline_gap_heatmap:horizontal} we present the degradation in target performance from using validation-based early-stopping (baseline), across all our settings, namely for each pair of source and target datasets, and across all three Meta-Learning algorithms. This degradation, also referred generalization gap, is the difference between the oracle target accuracy (if early-stopping had been optimal) and the target accuracy obtained from validation-based early-stopping, \textit{i.e.} : $Acc_{oracle} - Acc_{baseline}$.
In the \textit{bottom row} of Fig.~\ref{fig:baseline_gap_heatmap:horizontal}, we show the generalization gains of using Neural Coherence, and how it is able to close this generalization gap, where blue indicates an improvement over the baseline, and red indicates a deterioration. We observe that, overall, our method improves over the baseline, with very few instances of notable deterioration of generalization. Second, by comparing our method (bottom row) with the baseline (top row), we see that our method offers gains (blue) where the baseline suffers more severely (red). Meanwhile, when the baseline performance is close to the optimum (white and light color) and there isn't much to gain, we perform on par with the baseline (white and light color), as desired. A comparison of the numerical values reveals that we often fill a large portion of the baseline generalization gap, sometimes even completely.

To expand our empirical validation beyond simple convolutional networks, we then demonstrate Neural Coherence with another architecture class, namely Residual Network. We use a ResNet-18, with MiniImagenet and Omniglot respectively as our source and target datasets, and across the three meta-learning algorithms. We use 5-way 1-shot learning tasks. In each experiment trial, NC has access to a fixed target task support set (5 examples), to perform checkpoint selection, with performance averaged over 50 independent trials per experiment. Results show that for three meta-learning algorithm, our method outperforms the Source-Validation baseline.

\begin{table}
\centering
\begin{tabular}{|cccc|}
\hline
\multicolumn{4}{|c|}{ResNet-18} \\ \hline
\multicolumn{1}{|c|}{\textbf{Method}} & \multicolumn{1}{c|}{MAML} & \multicolumn{1}{c|}{ProtoNet} & Matching Network \\ \hline
\multicolumn{1}{|c|}{\textit{Oracle}} & \multicolumn{1}{c|}{55.9} & \multicolumn{1}{c|}{66.9} & 64.9 \\ \hline \hline
\multicolumn{1}{|c|}{\begin{tabular}[c]{@{}c@{}}Source-Val \end{tabular}} & \multicolumn{1}{c|}{52.4} & \multicolumn{1}{c|}{61.4} & 57.3 \\ \hline
\multicolumn{1}{|c|}{\textbf{NC}} & \multicolumn{1}{c|}{\textbf{55.4}} & \multicolumn{1}{c|}{\textbf{62.5}} & \textbf{62.9} \\ \hline
\end{tabular}
\caption{Generalization performance of Neural Coherence, using a ResNet-18 as the neural architecture. For each setting, ``Oracle" is the maximum achievable generalization performance, if early-stopping had been optimal (with an oracle). We used Mini-Imagenet as the source dataset and Omniglot as the target dataset. The first row shows the maximum achievable target accuracy. Results show that for Meta-Learning algorithm, our method outperforms the Source-Validation baseline.}
\label{tab:resnet}
\end{table}

\paragraph{Additional baseline : Support loss of a single task}\label{sec:appendix:exp_results:baseline_single-task_support_loss}

Since our method has access to the unlabelled data from a single target task, we also compare its performance to another baseline : tracking the support loss of the single target task, after fine-tuning on its  examples (since we use 1-shot, we can randomly label the examples). This captures how well can the model classifies the support examples of the task. We use the cross-entropy loss instead of the classification accuracy which can easily saturate even after a single step. We performed this analysis using MAML (as it is not applicable with the other two algorithms). We used all of the source-target dataset pairs considered so far. For each setting we use, as with our Neural Coherence based approach, a fixed set of 50 target tasks, and keep their sets of support examples. At the end of each training epoch, for each of the 50 tasks, we feed the support examples to the model to get predictions. Since we use 1-shot of examples, we randomly label them,  compute the support loss, and fine-tune the model (only the final classification layer, and freeze the feature extractor). After fine-tuning, we reevaluate the support loss using the adapted parameters of the classifier. For each task, we track this loss throughout the entire training experiment, we early-stop at the minimum value of the loss, and evaluate the target generalization at that point in time. We average this performance over the 50 target tasks.

\begin{figure}[ht]
    \centering
    \begin{minipage}{0.55\linewidth}
        \caption{Performance when using the support loss of a single target task for early-stopping (baseline method). 
        Difference in generalization performance between this baseline and the baseline of validation early-stopping. This baseline performs significantly worse than using Neural Coherence.}
        \label{fig:performance_heatmap:single-task_baseline:horizontal}
    \end{minipage}
    \hfill
    \begin{minipage}{0.4\linewidth}
        \includegraphics[width=\linewidth]{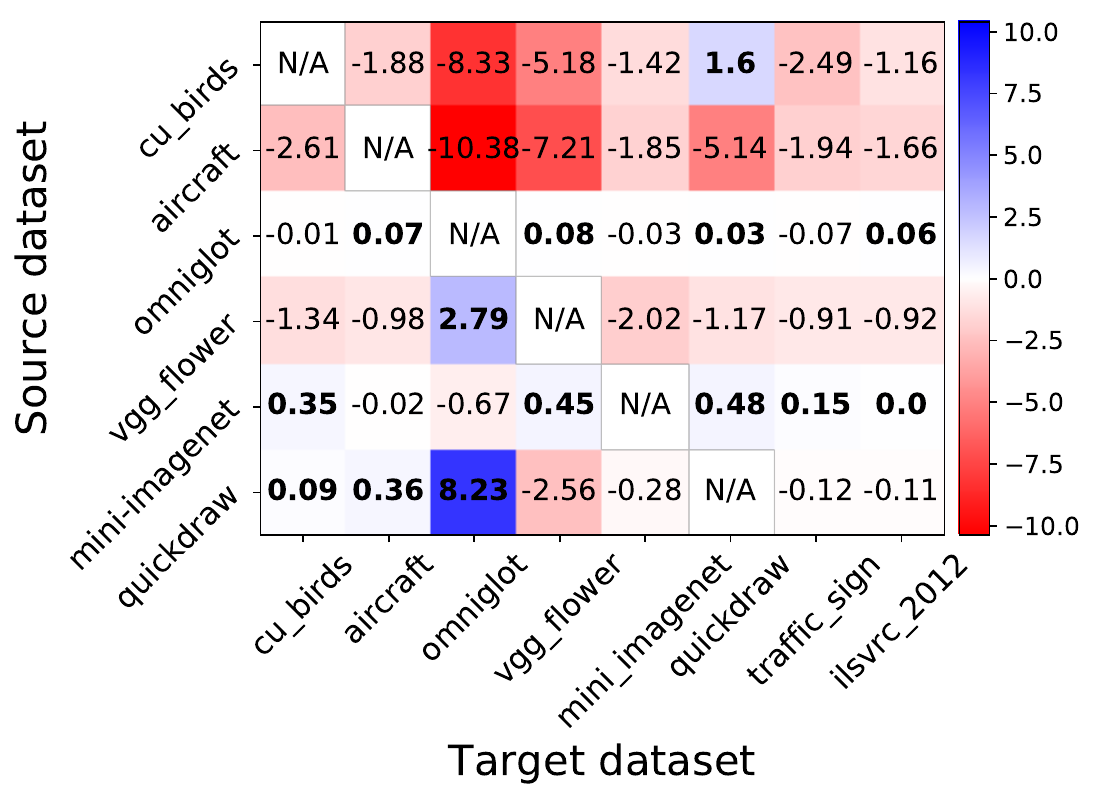}
    \end{minipage}%
\end{figure}

Results show that this baseline doesn't not perform as well as Neural Coherence, and in fact performs worse than the validation-based early-stopping, as it increases the generalization gap by 54.1\% on average, while with MAML, our method closes the generalization gap by 71.4\% on average. We also observed a much greater variance between the estimated stopping times for this baseline, when using different tasks, and thus a higher standard deviation in performance of 0.74\% (in accuracy, on average) compared to only 0.35\% for Neural Coherence used with MAML. See results in Fig. \ref{fig:performance_heatmap:single-task_baseline:horizontal} where there are mostly degradation of generalization (red color) and very few instances of improvement (blue color), as opposed Neural Coherence, in the bottom left subfigure of  Fig. \ref{fig:baseline_gap_heatmap:horizontal}.
This reinforces some of the insights of our work -- to infer target generalization, one might have to " look under the hood " and inspect the activations at the lower layers of the feature extractor. We also observed a much greater variance between the estimated stopping times for this baseline, when using different tasks, as compared to Neural Coherence.

\subsection{Large vision models}
\label{sec:exps:ckpt_selection:large_vision_models}

In these experiments we focus on the ConvNeXt-Large network \citep{convnext}, a modern state-of-the-art architecture, which has around 200 million parameters with more than two hundred hidden layers. The ConvNeXt family furthermore incorporates aspects of ResNet and Transformer architectures, making it a good indicator for the applicability of Neural Coherence (NC).
We train ConvNeXt-L using the setup of the original work of \cite{convnext}.
To demonstrate the performance of NC, we consider two different OOD paradigms : zero-shot generalization and transfer-learning under extreme sparsity of target data.
For each paradigm, we evaluate the performance of NC on four different target datasets. In each paradigm, target datasets fall into two scenarios:
In Scenario A, the epoch of maximum validation accuracy does not correspond to the time where maximum target accuracy is achieved. In this scenario, the Source-Val baseline suffers from a generalization gap. This offers the opportunity to assert if NC is able to fill some of that generalization gap to an extent, by stopping closer to the true optimum, improving downstream performance.
However, it would be methodologically incorrect to only use settings where Source-Val is not working well. Indeed, one cannot assume if Source-Val would stop at the right time. Thus, NC needs to perform well in such scenarios. Moreover, if we don't test for those settings, we don't know if NC is simply only stopping earlier than Source-Val. Such settings fall into the Scenario B, where the maximum performance is actually to be on par with the Source-Val baseline.
Each experiment, the performance of NC is evaluated for a given target dataset, over several independent trials. In each trial, NC has access to a fixed set of $n$ unlabelled target examples to compute the target trajectory $\psi(\mathbf{z}_{T}; t)$, and uses the whole validation split (from the source domain) to compute the source trajectory $\psi(\mathbf{z}_{S}; t)$. In each trial, NC provides a stopping epoch $\hat{t}$. The performance for the trial is obtained by selecting the pre-trained checkpoint of epoch $\hat{t}$ and evaluating its accuracy on the entire test set of the target dataset. For each experiment, the performance is averaged over all trials. If not mentioned otherwise we use $n=5$ samples from the target dataset. Note that 5 samples is drastically small compared to the size of the target datasets involved, which typically comprise tens of thousands of examples.

As baselines methods for comparison, we use:  1) Source-Validation, where we simply select the checkpoint with the highest accuracy on the entire validation set; 2)  (Target-Val), where this baseline uses $n$ fixed target labelled examples to approximate the target generalization such that $n$ matches the fixed number of unlabelled examples from the target problem that Neural Coherence (NC) uses; 3) Activation-Based Earlystopping (ABE) \citep{Guiroy2020}, an early stopping technique working on the same principle as Neural Coherence, also operating on $n$ unlabeled examples; 4) Negative Cross-Entropy (NCE) \citep{nce}; and 5) PARC \citep{parc} as additional model selection baselines that fall into the heuristic category. In each experiment, to give context on the obtained performance, we show the maximum performance that was achievable if an \textit{Oracle} could stop at the time of maximum target accuracy. We obtain this performance by evaluating all trained checkpoints, computing their accuracy on the entire test set of the target dataset. The Oracle performance is the best checkpoint's average accuracy on the full target test set. But the target test set is not accessible in practice, and the Oracle should not be confused as a practical baseline method, only a reference as the maximum achievable performance. 

\begin{figure}
    \centering
    \subfloat[Without weighting layers]{\includegraphics[width=0.25\textwidth]{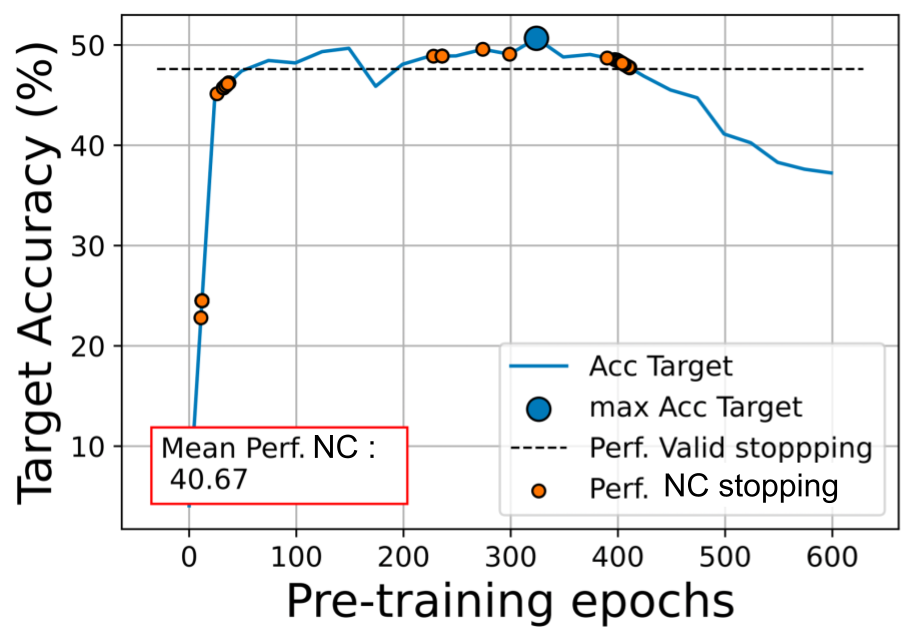}
    \label{fig:empirical_demos:weighting_neural_coherence:unweighted}}
    \subfloat[With layer weighting]{\includegraphics[width=0.25\textwidth]{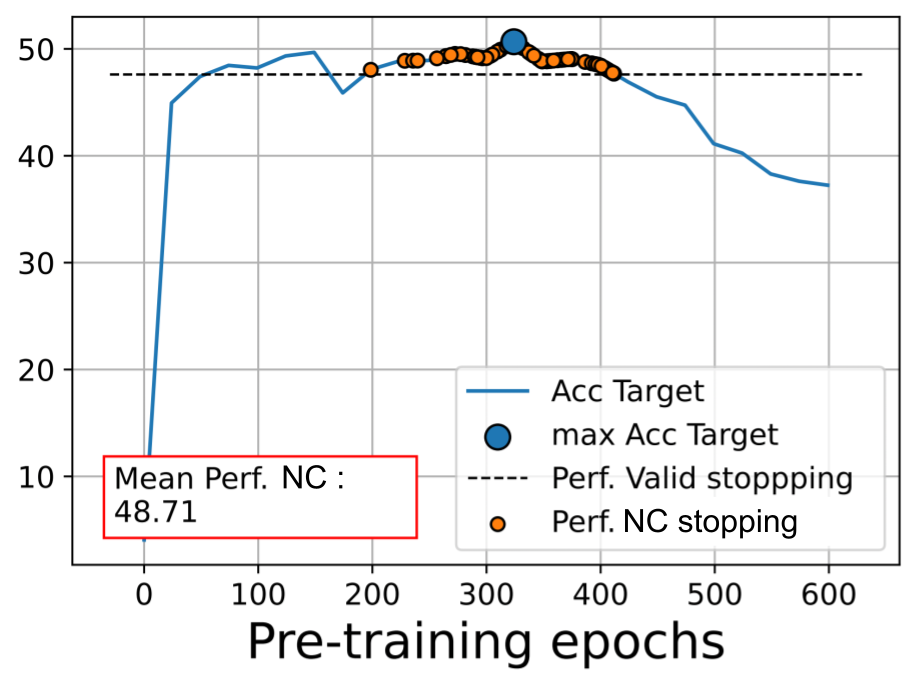}
    \label{fig:empirical_demos:weighting_neural_coherence:weighted}}
    \caption{Effect of weighting the Neural Coherence across layers. ConvNeXt-L trained on ImageNet1k with Food101 as the target. Multiple trials of checkpoint selection (orange dots) using a batch of 20 unlabeled examples. Without weighting Neural Coherence layer-wise (a), the high variance in estimated stopping time (orange dots) leads to a lower average target accuracy, while when weighting the layers contributions (b), the selected checkpoints are clustered around the global optimum (max Acc Target), outperforming the source validation earlystopping baseline (dotted line).}
    \label{fig:empirical_demos:weighting_neural_coherence}
\end{figure}

Before presenting the performance experiments, we start with a simple analysis and demonstration of the effect of weighting the neural coherence across the layers of the network as introduced in Sec. \ref{sec:practical_implementation:checkpoint_selection} for deeper architectures, as opposed to basing the stopping criterion on a single layer $f_l$. The results in Fig. \ref{fig:empirical_demos:weighting_neural_coherence} show that weighting the Neural Coherence across layers reduces the variance of selected checkpoints, which will cluster around the global optimum, thus improving average target performance, and outperforming the source-validation baseline. This unweighted version is the ABE algorithm (Activation-Based Earlystopping) from \cite{DBLP:conf/collas/GuiroyPMC22}, which is a simple implementation of the Neural Coherence principle not designed for large models. In the subsequent experiments of the current section (Sec. \ref{sec:exps:ckpt_selection:large_vision_models}), when reporting the performance of Neural Coherence (NC) early-stopping with large models, and comparing it with other baseline methods, we will add the performance of ABE. This will further demonstrate the necessity of weighting the neural coherence across layers in large networks.

\paragraph{Zero-shot generalization}
\label{sec:exp:zeroshot}

\begin{figure}[ht]
    \centering
    \subfloat[Target Accuracy]{\includegraphics[width=0.3\textwidth]{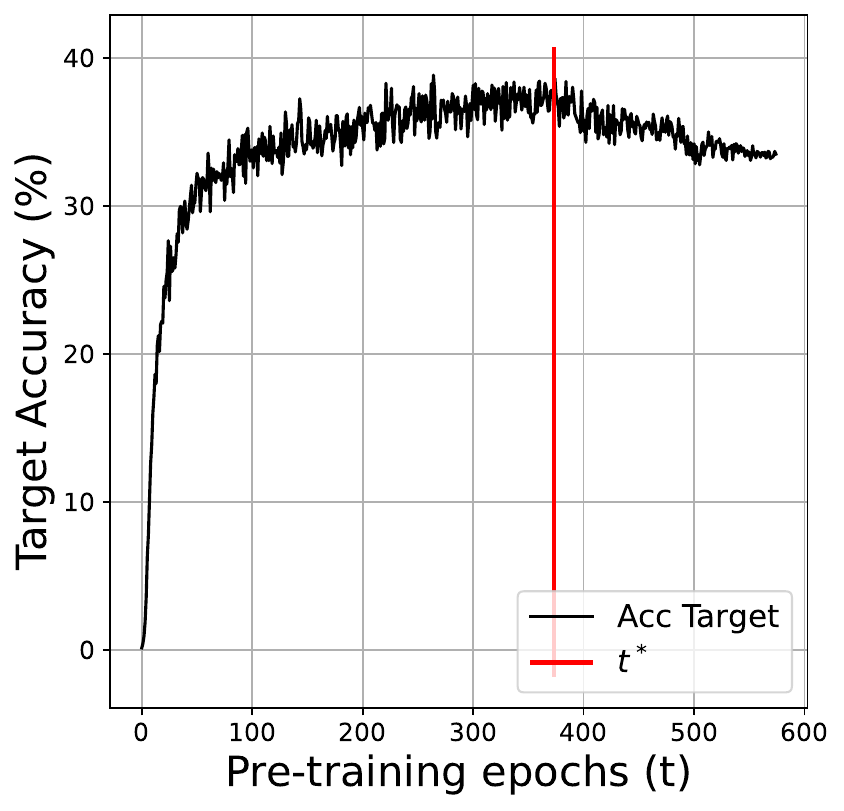}}
    \subfloat[Neural Activation Trajectories]{\includegraphics[width=0.3\textwidth]{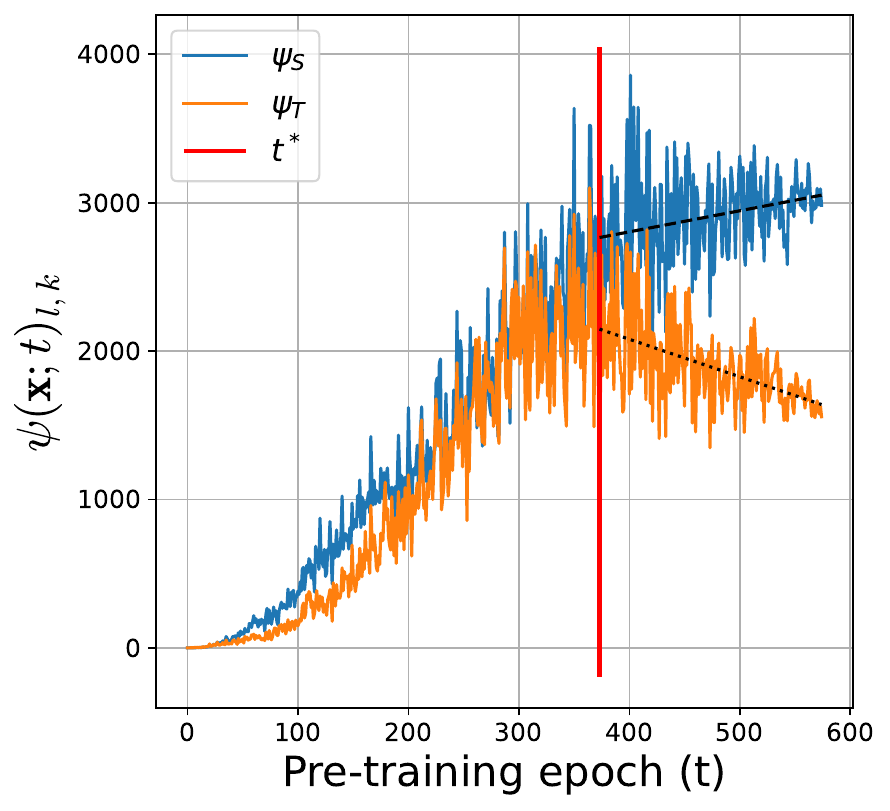}}
    \caption{Qualitative demonstration of Neural Coherence for checkpoint selection, with large vision models for zero-shot learning. A ConvNeXt-L is trained on ImageNet1k as the source task, and is deployed on ImageNet-Sketch \citep{imagenetsketch} as the target task. Neural Coherence yields the early-stopping time $t=t^*$ (expressed as training epoch), when the trajectories (of the activation distributions for the source and target input data) are no longer coherent with each other. This time $t^*$ occurs approximately when target accuracy reaches its maximum, while source validation accuracy, not visible here, keeps increasing. The target trajectory $\psi_T$ (orange) is computed using only 5 target input samples and the source trajectory $\psi_S$ (blue) uses the whole source validation set.}
\label{fig:empirical_demos:coherence_function}
\end{figure}

We investigate the capabilities of Neural Coherence in a simple zero-shot generalization setting.
We consider Imagenet-A \citep{imageneta}, Imagenet-R \citep{imagenetr}, ObjectNet \citep{objectnet} and Imagenet-Sketch \citep{imagenetsketch} as target datasets for our ImageNet1k-pretrained model. This allows for a simple zero-shot-generalization setup, since  all datasets have the same classes as ImageNet1k (except ObjectNet, for which we use the set of 113 overlapping classes with ImageNet1k). In this paradigm, the model performs no fine-tuning, and its performance is directly evaluated as its average accuracy on the test split of the target dataset (zero shot generalization). For NC, ABE and Target-Val, we run 250 independent trials, where each trial uses a fixed set of $n=5$ target examples. The Source-Val baseline has access to the entire validation split of Imagenet. For each experiment, the performance is computed on the full test split of the target dataset, and averaged over all the trials. The Oracle correspond to the maximum possible target accuracy, i.e. if an oracle knew exactly when to stop.

From Tab. \ref{tab:zeroshot}, we observe that NC is significantly outperforming ABE and Target-Val, for each experiment. The experiments are grouped into two scenarios A and B, which serve as positive and negative test cases for our early-stopping method. In the Scenario A (positive test case), the maximum target accuracy happens \textit{before} the maximum validation accuracy, hence and there is a generalization gap between Source-Val and the Oracle. Our results indicate that NC can stop at a more appropriate time to improve the downstream performance. Relative to the Source-Val performance, NC fills 30\% and 20\% on Imagenet-Sketch and Imagenet-R respectively. In the experiments of the scenario B (negtive test case), there is no better stopping time before the time of maximum validation accuracy. Despite only using 5 examples, NC performs comparatively to Source-Val, while this baseline uses the entire validation set of Imagenet. The results demonstrate that NC is capable of adequately addressing the domain shift. In particular, the significant performance improvements compared to Target-Val and ABE highlight that we successfully addressed the challenges of analyzing neural coherence in deep architectures. In both scenarios, PARC and NCE, while being established model selection methods, perform significantly worse.

\begin{table}[ht]
\centering
\begin{tabular}[t]{lcc|cc}
\toprule
&\multicolumn{2}{c}{\begin{tabular}{c}
     \textbf{Scenario A:} $\max Acc_{target}$\\
     happens before $\max Acc_{valid}$
\end{tabular}} & \multicolumn{2}{c}{\begin{tabular}{c}
     \textbf{Scenario B:} $\max Acc_{target}$ does\\
    \textit{not} happen before $\max Acc_{valid}$
\end{tabular}}
\\
\toprule
&ImageNet-S&ImageNet-R&Imagenet-A &ObjectNet\\
\midrule
Oracle&37.72&35.32&14.58&36.55\\ \hline \hline
Source-Val&35.75&34.08&14.58&36.55\\
Target-Val&30.96&28.58&9.55&30.44\\
ABE  &24.66 ± 0.83&32.71 ± 0.30&8.34 ± 0.68&30.15 ± 0.97\\
NCE&0.92 ± 0.00&15.31 ± 3.02&3.91 ± 1.12 &16.90 ± 3.34\\
PARC&23.76 ± 3.12&11.85 ± 2.84&8.34 ± 0.85&2.17 ± 1.40\\
 \textbf{NC@5 (ours)}&\textbf{36.34} ± 0.04&\textbf{34.32} ± 0.06&14.45 ± 0.01&36.51 ± 0.05\\
\bottomrule
\end{tabular}
\caption{Checkpoint Selection : Zero-Shot Generalization. ConvNeXt-Large pre-trained on Imagenet. Target datasets have the same classes as Imagenet, but different input domains. The Oracle provides the maximum achievable performance, but is not a method. Baselines are Source-Validation, Target-Validation, ABE \citep{DBLP:conf/collas/GuiroyPMC22}, NCE \citep{nce}and PARC \citep{parc}. Overall, Neural Coherence outperforms the other baselines. In Scenario A, the optimum for target accuracy happens before the source optimum ($Acc_{valid}$), Neural Coherence result in successful early-stopping, improving target generalization, compared to Source-Val. In Scenario B, the target optimum does not happen before the source optimum, and Neural Coherence successfully selects the optimum checkpoint.}
    \label{tab:zeroshot}
\end{table}%

\paragraph{Transfer Learning}
Pretraining on a large generic dataset and transferring the learned features to a smaller and application specific dataset, due to the rise of foundation models, has become the most significant application for checkpoint selection. For this reason, we use Imagenet1K as the source dataset. We use FOOD-101 \citep{food101}, iNaturalist \citep{horn2017inaturalist}, PlantNet300K \citep{Garcin2021PlntNet300KAP} and EuroSat \citep{eurosat} as target datasets.
To adapt a pre-trained model checkpoint, we train a new softmax classifier until convergence while keeping the rest of the network frozen.
Otherwise, we are using the same fine-tuning setup as in \cite{convnext}.
While this adaption does not produce state-of-the-art results, it allows us to more precisely evaluate the quality of the checkpoint selection. 
Since the softmax-classier is linear, the quality of the adapted model primarily depends on the frozen feature extractor and by extension the quality of the selected checkpoint. Note that because the classes are different between the target and source domains, this does not allow to directly compare NC with the Target-Val baseline. 

\begin{figure}[ht]
    \centering
    \subfloat[MAE Valid Loss]{\includegraphics[width=0.25\textwidth]{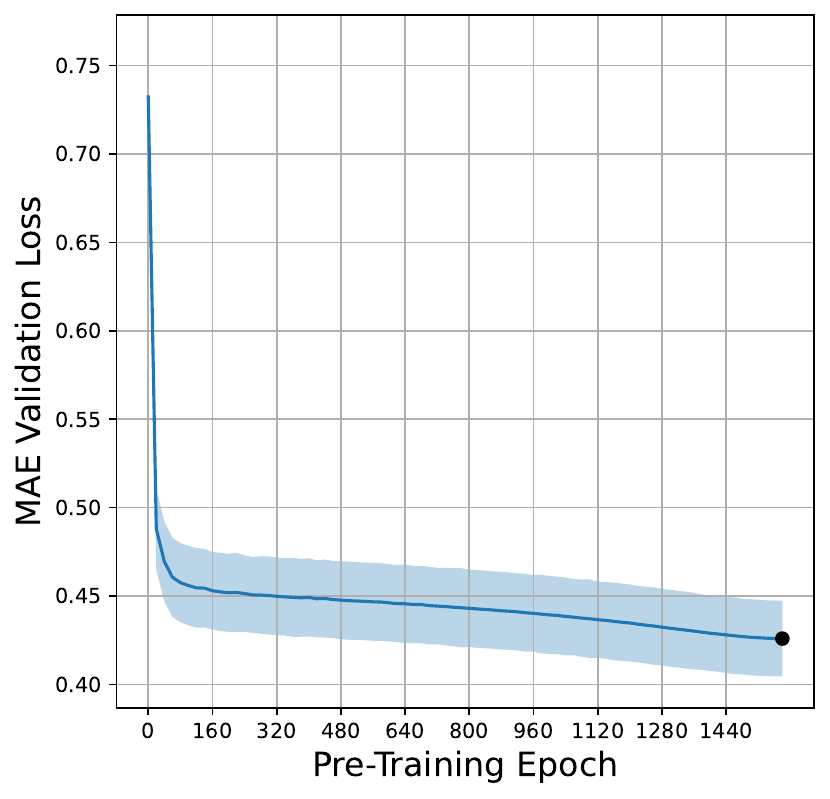}
    \label{fig:mae:valid_loss}}
    \subfloat[Target Acc on EuroSAT]{\includegraphics[width=0.25\textwidth]{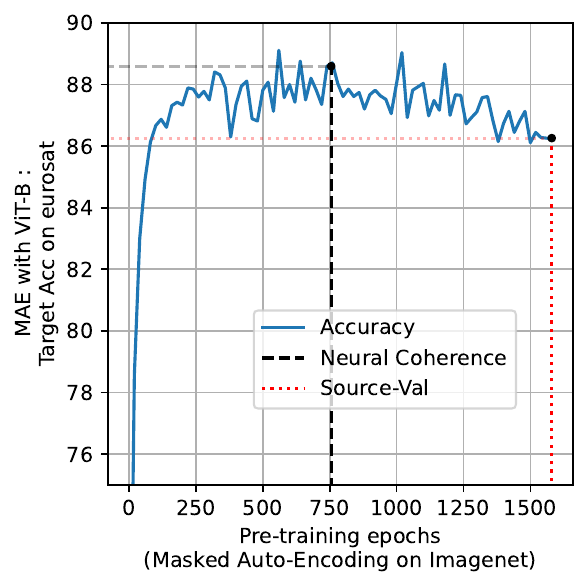}
    \label{fig:mae:target_acc}}
    \caption{Neural Coherence on a Vision Transfrormer (Vit-B), trained in Self-Supervised Learning with a Masked Autoencoder (MAE). The model is trained on Imagenet and the target problem is EuroSAT. While the validation loss kept decreasing, the target performance on EuroSAT peaks around epoch 750, and the Source-Val baseline yields sub-optimal checkpoint selection (dotted line). On the other hand, Neural Coherence yields an early-stopping checkpoint (dashed line) near the optimum, outperforming the Source-Val baseline.}
    \label{fig:mae}
\end{figure}

In Tab. \ref{tab:transfer}, similarly to the zero-shot experiments, NC outperforms ABE and Source-Val in Scenario A, even though Source-Val has access to the full validation set while NC is only using 5 unlabeled examples. For FOOD-101 and iNaturalist, NC respectively close 76\% and 82\% of the generalization gap that Source-Val suffers.
In Scenario B,  NC selects the optimal checkpoint reliably for PlantNet300k, while on EuroSAT it outperforms ABE and is almost on par with Source-Val.
In summary, the results demonstrate that NC selects high-quality pre-training checkpoints reliably in all tested scenarios. Furthermore, in Fig. \ref{fig:mae} we add an analysis suggesting that Neural Coherence also indicate target performance in transformer architectures, with a Vision Transformer (ViT) trained with Masked Autoencoding (MAE) trained in self-supervised learning \citep{vit,mae}.

\begin{table}[ht]
\centering
\begin{tabular}[t]{lcc|cc}
\toprule
&\multicolumn{2}{c}{\begin{tabular}{c}
     \textbf{Scenario A:} $\max Acc_{target}$\\
     happens before $\max Acc_{valid}$
\end{tabular}}&
\multicolumn{2}{c}{\begin{tabular}{c}
     \textbf{Scenario B:} $\max Acc_{target}$ does\\
    \textit{not} happen before $\max Acc_{valid}$
\end{tabular}}
\\
\toprule
&FOOD-101&iNaturalist&PlantNet-300K&EuroSat\\
\midrule
Oracle&44.45&21.40&35.08&74.87\\ \hline \hline
Source-Val&41.72&19.46&35.08&74.87\\
ABE&40.17 ± 0.62&17.23 ± 0.17&27.78 ± 0.13&63.47 ± 0.11\\
PARC&35.16 ± 0.36&0.07 ± 0.00&30.04 ± 0.15&34.43 ± 0.53\\
\textbf{NC@5 (ours)}&\textbf{43.80} ± 0.06&\textbf{21.13} ± 0.04&35.08 ± 0.00&74.53 ± 0.02\\
\bottomrule
\end{tabular}
    \caption{Checkpoint Selection : Transfer Learning. Target datasets have different classes than Imagenet, and different input domains. Overall, Neural Coherence outperforms the other baselines. In Scenario A, the optimum for target accuracy happens before the source optimum ($Acc_{valid}$), Neural Coherence result in succesful early-stopping, improving target generalization, compared to Source-Val. In Scenario B, the target optimum does not happen before the source optimum, and Neural Coherence successfully selects the optimum checkpoint.}
    \label{tab:transfer}
\end{table}%

\subsection{Statistical Efficiency of Neural Coherence}
\label{sec:exps:ckpt_selection:statistical_efficiency}

\begin{figure}
    \centering
    \subfloat[Imagenet-Sketch]{\includegraphics[width=0.26\textwidth]{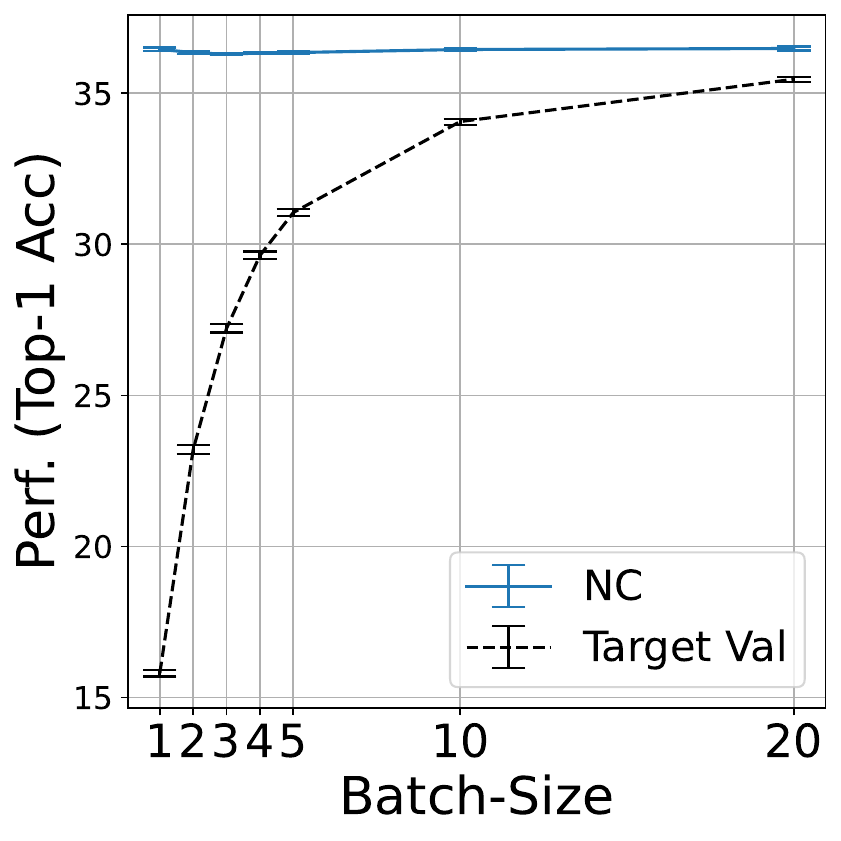}}
    \subfloat[Imagenet-R]{\includegraphics[width=0.24\textwidth]{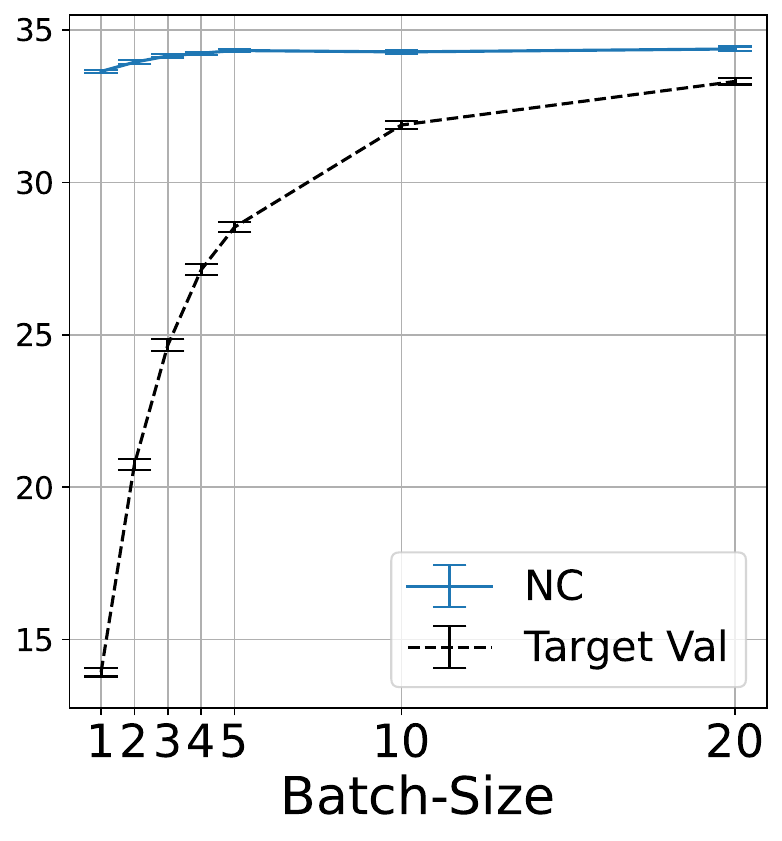}} 
    \subfloat[Imagenet-A]{\includegraphics[width=0.24\textwidth]{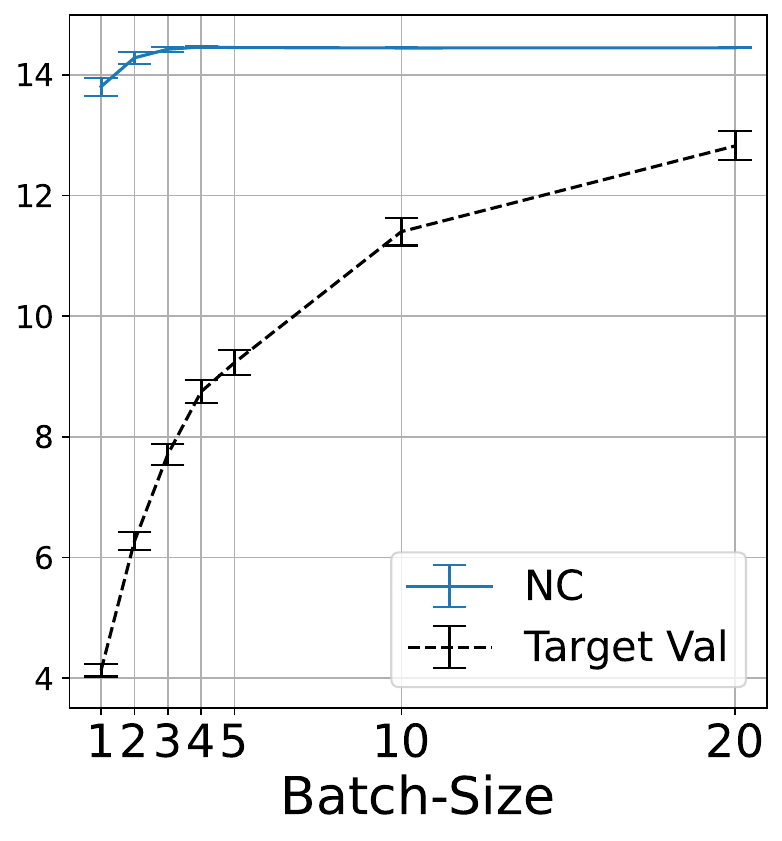}} 
    \subfloat[ObjectNet]{\includegraphics[width=0.24\textwidth]{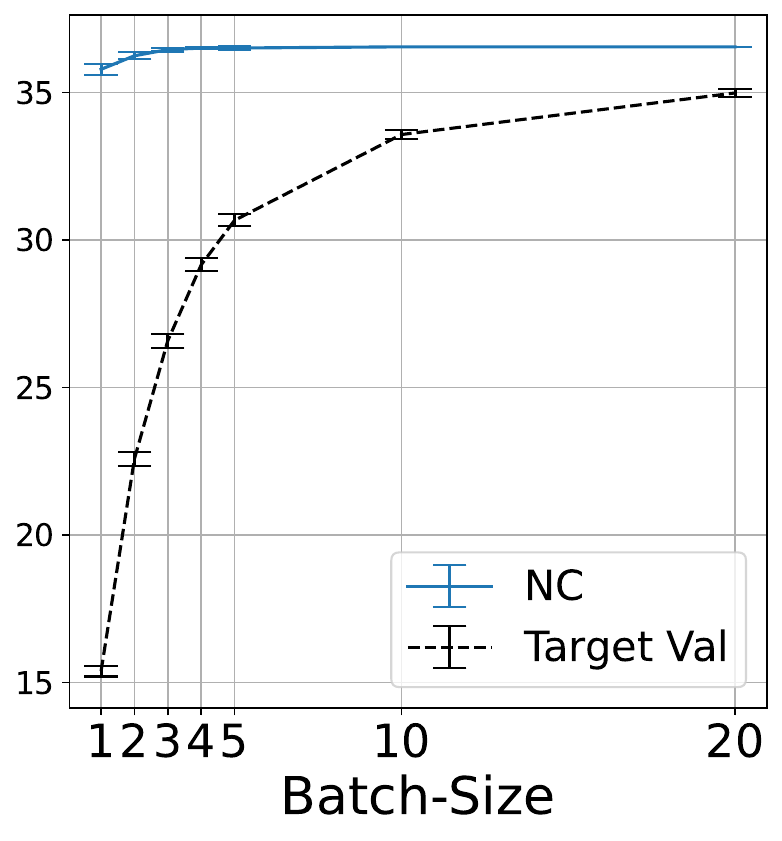}}    
\caption{Statistical efficiency of Neural Coherence across different low data regimes : We see that, for a same number $n$ of available examples, and across $n=\{1, 2, 3, 4, 5, 20\}$, Neural Coherence consistently outperforms the Target-Val baseline, in estimating the appropriate checkpoint. We used a ConvNext-Large network, trained on full Imagenet as the source and four different target datasets.}
    \label{fig:ablations:stat_efficiency}
\end{figure}

In this section we conduct a more systematic analysis of the statistical efficiency of Neural Coherence. Across a wider for the number of available examples used for model selection, we report experiments that demonstrate Neural Coherence consistently requires fewer examples than the Target-Val baseline, to achieve a similar level of downstream performance.
We repeat the experiment described in section \ref{sec:exp:zeroshot} comparing only NC@$n$ and Target-Val@$n$ (both methods using a fixed set of $n$ randomly picked target samples, to perform model selection). We evaluate the performance (with 95\% C.I. error bars) of both checkpoint selection methods for $n=\{1, 2, 3, 4, 5, 20\}$ to assess the ablation of predictive performance of the selected checkpoint as a function of the number used samples $n$. On all four target datasets used for this experiment we find that NC@$n$ is outperforming Target-Val@$n$ (See Fig. \ref{fig:ablations:stat_efficiency}).
In fact, in all those four scenarios, NC@$1$ using only a single unlabeled sample, outperforms Target-Val@$20$ (used 20 labeled samples), highlighting the statistical efficiency of Neural Coherence here, and perhaps, the inefficiency and inadequacy in using a low data direct evaluation of downstream performance.

\section{Using Neural Coherence for selecting good pre-training data} \label{sec:exps:data_selection}
\begin{figure}
    \centering
    \subfloat[Mini-Imagenet]{\includegraphics[width=0.2\textwidth]{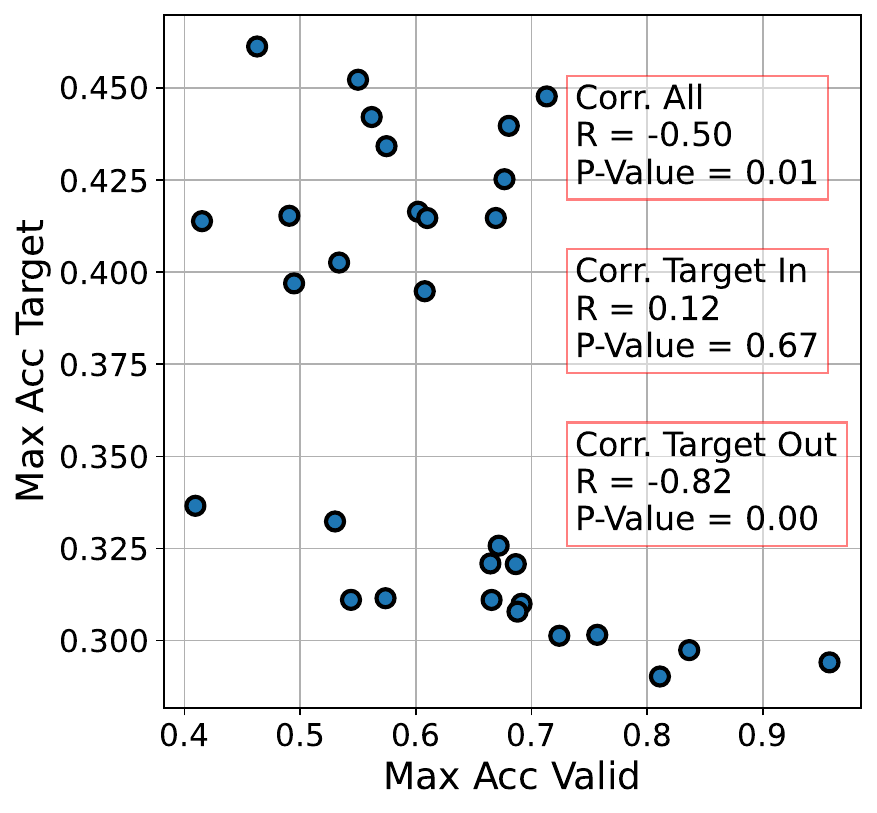}}
    \subfloat[Omniglot]{\includegraphics[width=0.2\textwidth]{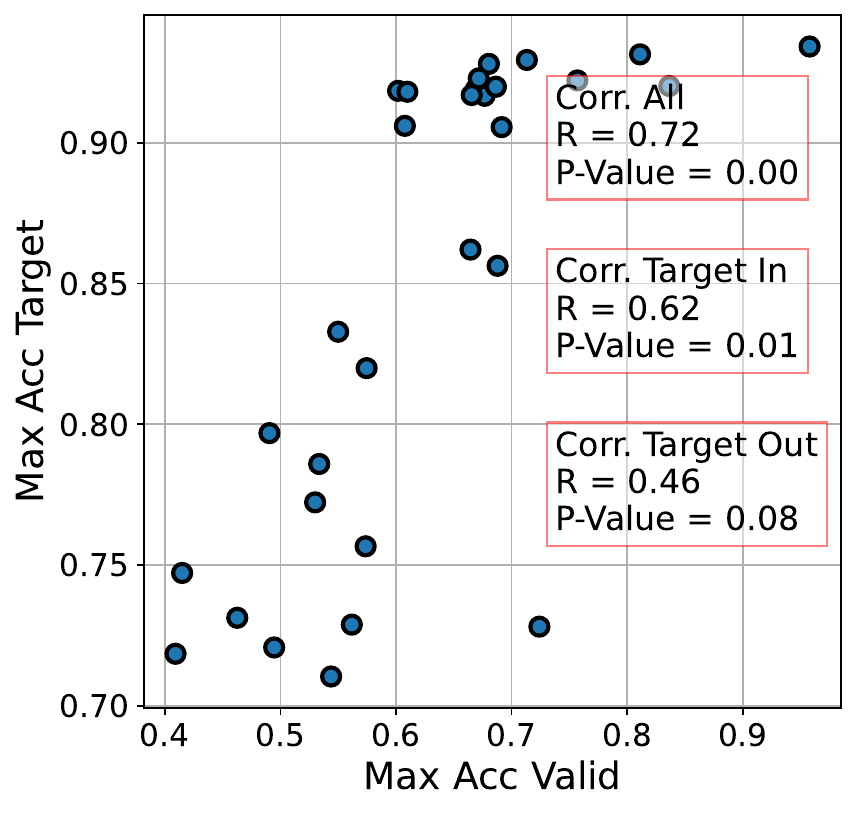}}
    \subfloat[Quickdraw]{\includegraphics[width=0.2\textwidth]{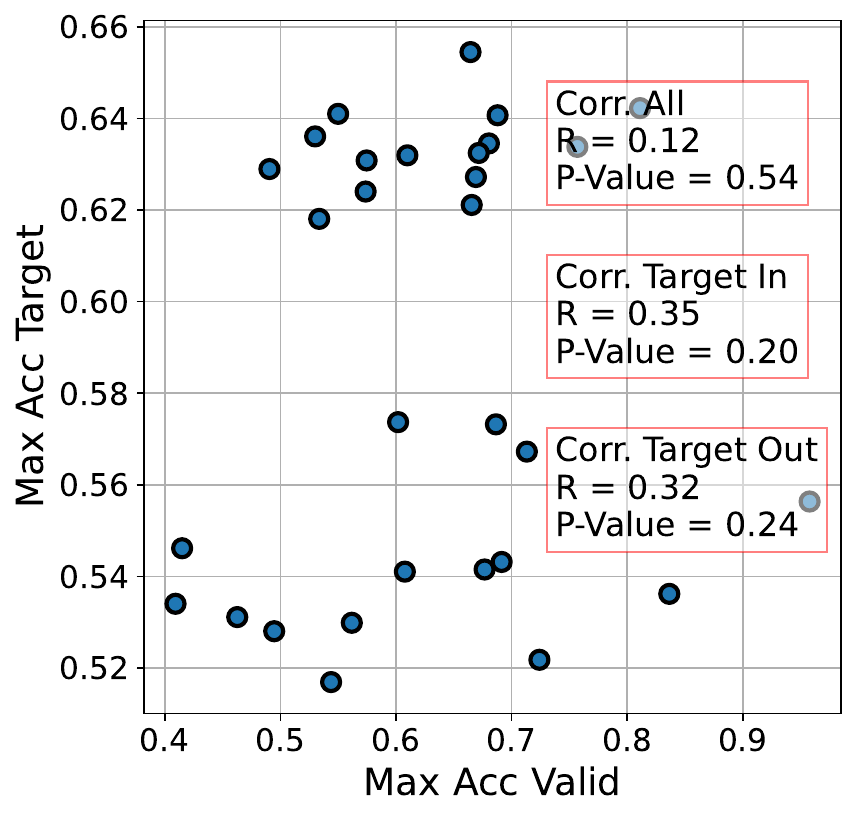}}
    \subfloat[Traffic Sign]{\includegraphics[width=0.2\textwidth]{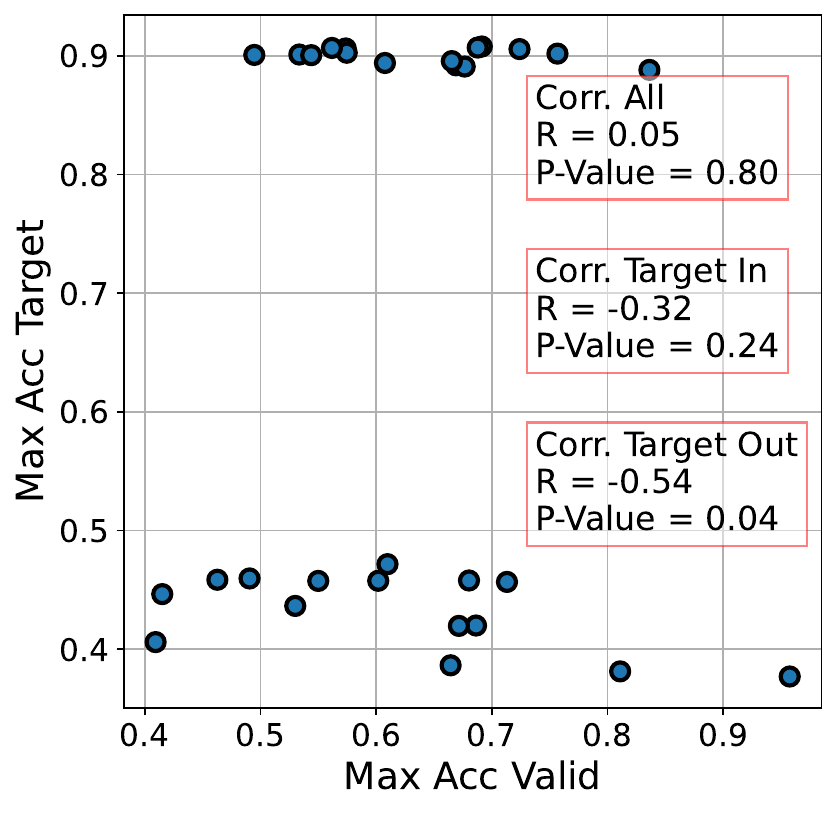}}
    \subfloat[DTD]{\includegraphics[width=0.2\textwidth]{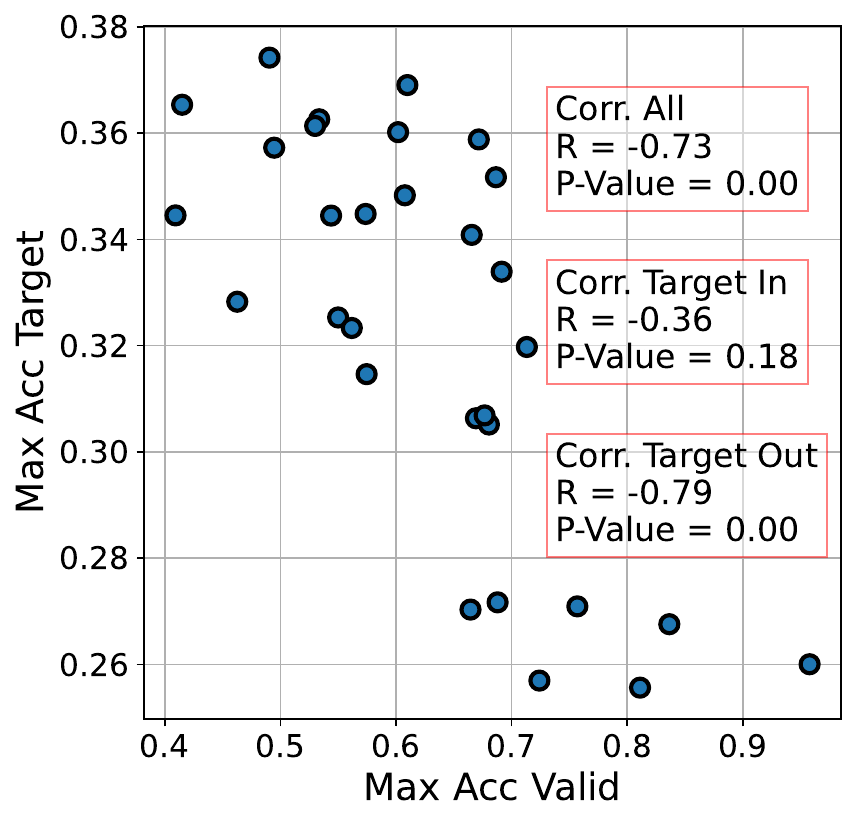}}
    \caption{For the problem of training data selection, In-Distribution Validation performance (Source-Val, x-axes) is not a good indication for target performance (y-axes).}
    \label{fig:corr_valid-acc_vs_target-acc}
\end{figure}

Neural Coherence is conceptually not bound to early stopping (selecting $t$), but is theoretically a versatile tool for different checkpoint selection scenarios, providing interesting avenues for future work. To achieve their best performance on a task of interest, modern large scale models are often pre-trained on a variety of different large datasets. For this reason, we will briefly demonstrate the versatility of Neural Coherence in a set of experiments on training dataset selection.
The dataset used for pretraining dramatically impacts the features the model will learn. Hence, the choice of pretraining data is a relevant task in scenarios where we have to rely on pretrained features.
Source performance on the pre-training set isn't guaranteed to correlate with the target generalization (See discussion below on Fig. \ref{fig:corr_valid-acc_vs_target-acc}). Thus, a more elaborate model selection scheme is necessary.
The data that one selects for pre-training is often selected heuristically. Here we will demonstrate that Neural Coherence can be used as a principled approach for selecting pre-training data. We achieve this by applying our method on models pre-trained on various datasets and combinations of datasets. 

\paragraph{Practical Implementation}

\begin{figure}
    \centering
    \includegraphics[width=0.7\linewidth]{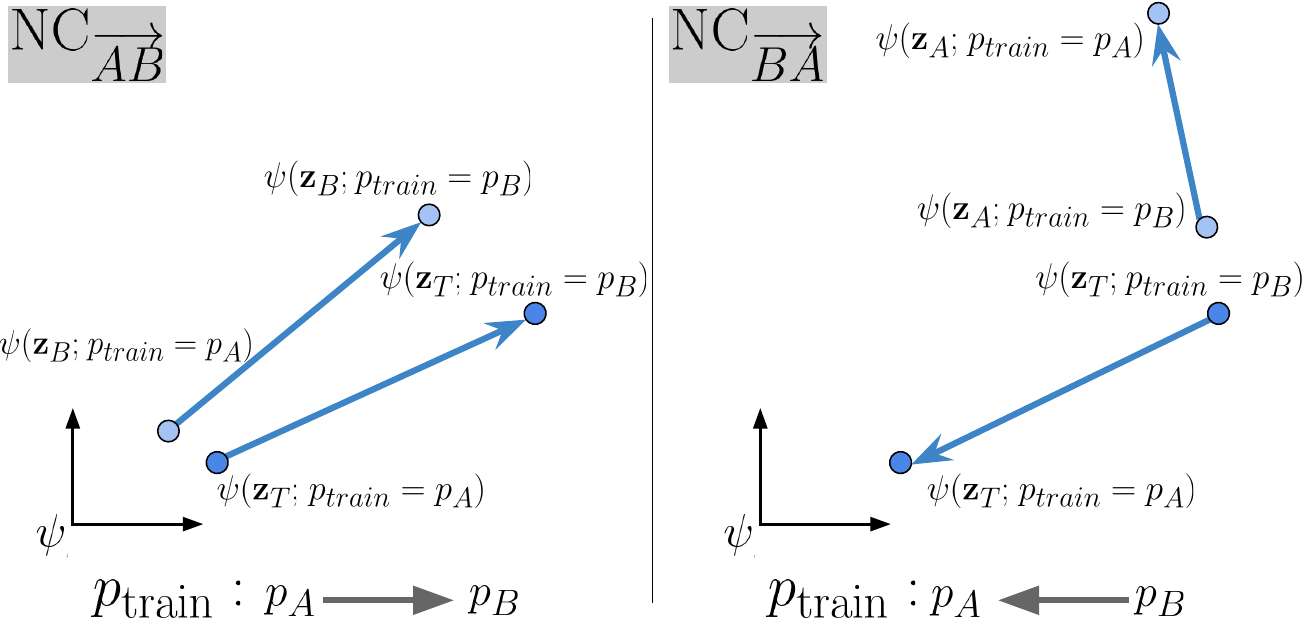}
    \caption{Neural Coherence with respect to the training data : Choosing between two candidate distributions $p_A$ and $p_B$. (\textit{left}) As $p_{\mathrm{train}}$ goes towards $p_B$, and generalization to $p_B$ increases, $\mathrm{NC}_{\overrightarrow{AB}}$ is the neural coherence measured between $\psi(\mathbf{z}_T; p_{\mathrm{train}})$ and $\psi(\mathbf{z}_B; p_{\mathrm{train}})$. In (\textit{right})  $\mathrm{NC}_{\overrightarrow{BA}}$ is measured between $\psi(\mathbf{z}_T; p_{\mathrm{train}})$ and $\psi(\mathbf{z}_A; p_{\mathrm{train}})$ as $p_{\mathrm{train}}$ goes towards $p_A$. If $\mathrm{NC}_{\overrightarrow{AB}} > \mathrm{NC}_{\overrightarrow{BA}}$, we select $p^*_{\mathrm{train}} = p_B$ as the training distribution, otherwise, we pick $p^*_{\mathrm{train}} = p_A$.}
    \label{fig:illustration:data-wise_nc}
\end{figure}

\begin{figure}
    \centering
    \subfloat[Target Perf vs. Train Data]{\includegraphics[width=0.3\textwidth]{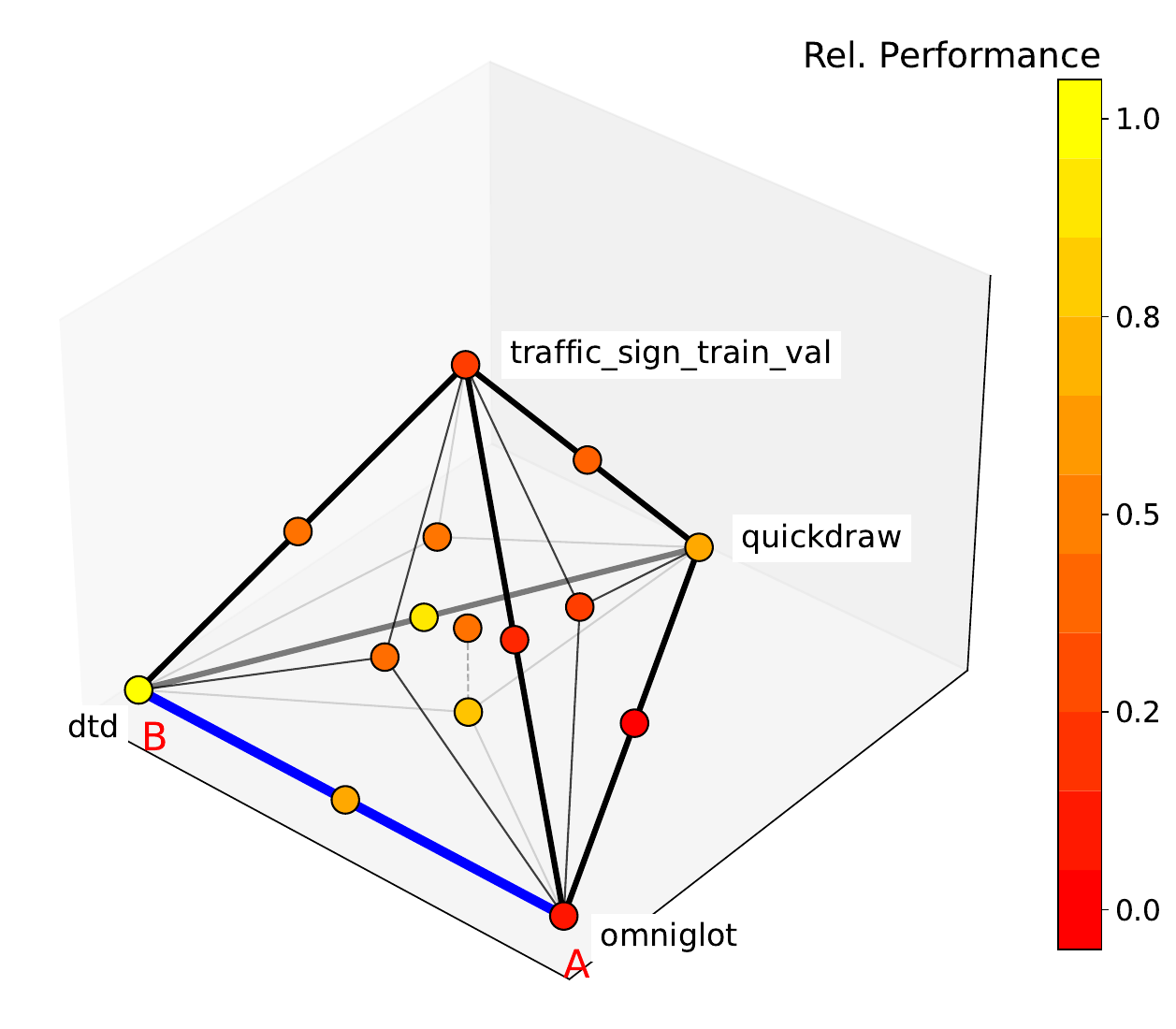}}
    \subfloat[Higher neural coherence towards B reflects that B is a better training dataset than A, given the target dataset.]{\includegraphics[width=0.59\textwidth]{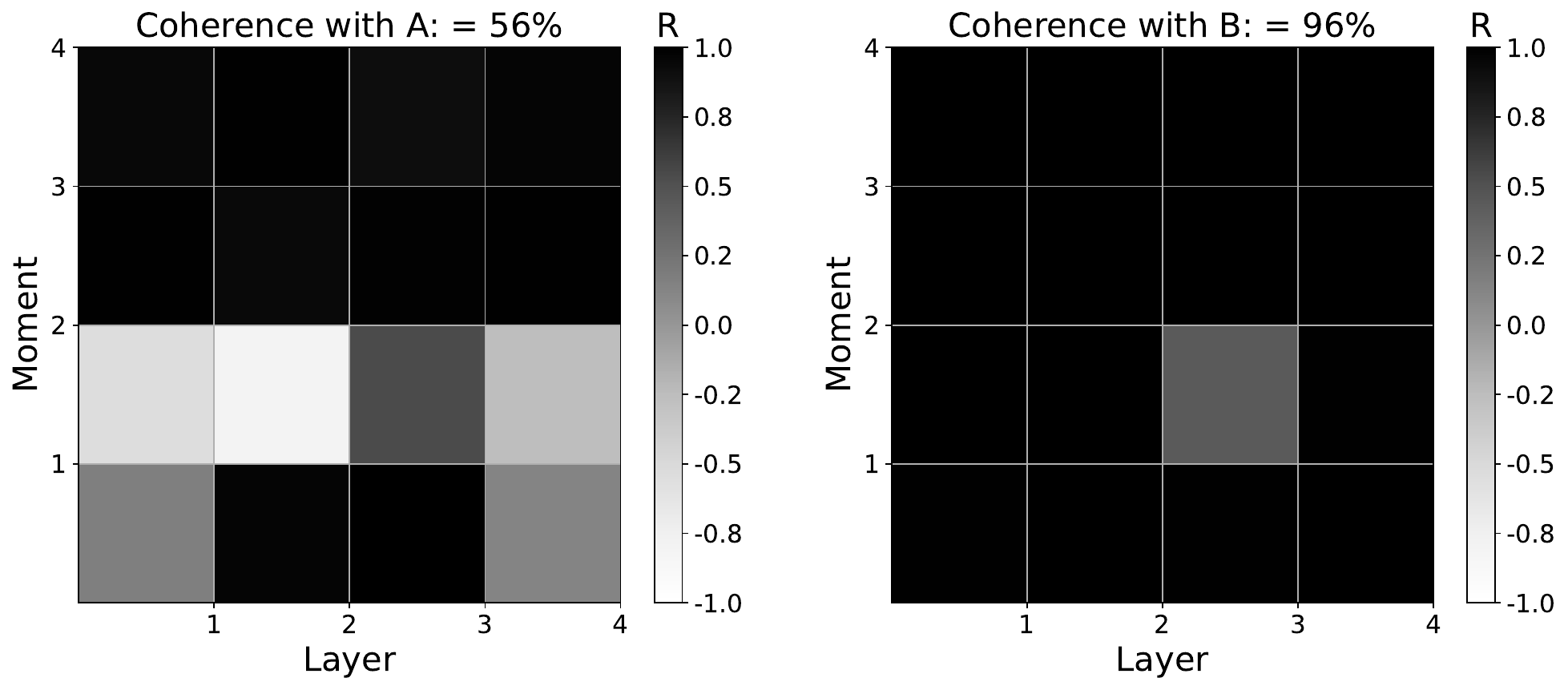}}
    \caption{Neural Coherence during training data selection. Dataset A: Omniglot, Dataset B: DTD, Target: Mini-Imagenet. (a) Target accuracy for different training mixtures—DTD (yellow) yields higher performance than Omniglot (red). (b) Neural coherence between source and target, across layers $f_l$ and moments $\hat{m}_k$, i.e. the elements of the matrix $\psi(\mathbf{z}; p_{\mathrm{train}})$, is stronger in when $p_{\mathrm{train}}$ goes towards dataset B (96\%), than towards A (56\%), indicating DTD aligns better with the target.}
    \label{fig:exp:data_selection:qualitative}
\end{figure}

Here we examine a simple formulation for modifying a training set distribution $p_{\mathrm{train}}(\mathbf{x})$ according to a single hyper-parameter. Consider two candidate distributions $p_{\mathrm{A}}$ and $p_{\mathrm{B}}$, we can define the balance between the use of data from $p_{\mathrm{A}}$ and $p_{\mathrm{B}}$ as in Eq.\ref{eq:data-selection:p_train}:
\begin{equation}
    p_{\mathrm{train}}(\mathbf{x}; \Omega) = \Omega \; p_{\mathrm{A}}(\mathbf{x}) + (1-\Omega) \; p_{\mathrm{B}}(\mathbf{x})
\label{eq:data-selection:p_train}
\end{equation}
where $\Omega \in [0, 1]$. Here the Neural Coherence is thus measured along this line, and used to estimate the optimal value for $\Omega$. In this work, given two candidate distributions $p_{\mathrm{A}}$ and $p_{\mathrm{B}}$, we focus our experiments in the simpler case of finding the best candidate among those two. However, given a set of candidate distributions, of size larger than two, we apply this approach sequentially to perform pair-wise comparisons and estimating the best candidate distribution among the set, which we apply in our experiments.
However, dealing with training data selection poses some extra difficulty. As opposed to an hyperparameter like training time, here the order over the set of models is not given implicitly. In other words, there is no trivial way to know if going from $p_{\mathrm{A}}$ to $p_{\mathrm{B}}$ necessarily improves target generalization up to a point, or if it should be the other way around. This problem is exacerbated when the optimum lies at either extremity of $\Omega$.
To address the difficulty of interpreting whether the trajectories are coherent or divergent, along a given direction, we propose to compare their coherence between the forward and backward direction. In other words, along the forward direction where $p_{\mathrm{train}}$ goes towards $p_{\mathrm{B}}$, i.e. from $p_{\mathrm{A}}$ to $p_{\mathrm{B}}$, the performance and generalization to $p_{\mathrm{B}}$ will improve, as more of its data is seen during training. We thus measure the neural coherence between the activation trajectories $\psi(\mathbf{z}_T; p_{\mathrm{train}})$ and $\psi(\mathbf{z}_B; p_{\mathrm{train}})$, as defined in Eq.\ref{eq:data-selection:N_AB}, along this direction. And inversely, along the backward direction, where $p_{\mathrm{train}}$ goes towards $p_{\mathrm{A}}$, and where generalization to $p_{\mathrm{A}}$ will improve, we measure the coherence between $\psi(\mathbf{z}_T; p_{\mathrm{train}})$ and $\psi(\mathbf{z}_A; p_{\mathrm{train}})$, as defined in Eq.\ref{eq:data-selection:N_BA}.
\begin{minipage}{0.48\linewidth}
\begin{equation}
    \mathrm{NC}_{\overrightarrow{AB}} \doteq 
    \mathrm{NC}(\mathbf{z}_B, \mathbf{z}_T; \Omega_0=1, \Omega_{\tau}=0)
\label{eq:data-selection:N_AB}
\end{equation}
\end{minipage}\hfill
\begin{minipage}{0.48\linewidth}
\begin{equation}
    \mathrm{NC}_{\overrightarrow{BA}} \doteq 
    \mathrm{NC}(\mathbf{z}_A, \mathbf{z}_T; \Omega_0=0, \Omega_{\tau}=1)
\label{eq:data-selection:N_BA}
\end{equation}
\end{minipage}
where $\mathbf{x}_A \sim p_{\mathrm{A}}(\mathbf{x}), \quad \mathbf{x}_B \sim p_{\mathrm{B}}(\mathbf{x}), \quad \mathbf{x}_T \sim p_{\mathrm{target}}(\mathbf{x})$. 
Hence, if the neural coherence $\mathrm{NC}_{\overrightarrow{AB}}$ measured between $\psi(\mathbf{z}_T; p_{\mathrm{train}})$ and $\psi(\mathbf{z}_B; p_{\mathrm{train}})$ as the model is being optimized for $p_B$, is stronger than the neural coherence $\mathrm{NC}_{\overrightarrow{BA}}$ measured between $\psi(\mathbf{z}_T; p_{\mathrm{train}})$ and $\psi(\mathbf{z}_A; p_{\mathrm{train}})$ as the model is being optimized for $p_A$, i.e. $\mathrm{NC}_{\overrightarrow{AB}} > \mathrm{NC}_{\overrightarrow{BA}}$, then we conclude that $p^*_{\mathrm{train}} = p_B$ is a better training distribution than $p_A$, given our target problem. Conversely, if $\mathrm{NC}_{\overrightarrow{BA}} > \mathrm{NC}_{\overrightarrow{AB}}$, then we conclude that $p^*_{\mathrm{train}} = p_A$ is the better training distribution.
The procedure for selecting the training distribution from Neural Coherence is be given in Eq.\ref{eq:neural-coherence_data-selection}. We schematized this approach in Fig. \ref{fig:illustration:data-wise_nc}, and provide an experimental demonstration \ref{fig:exp:data_selection:qualitative}).
\begin{equation}
    p^*_{\mathrm{train}}(\mathbf{x}) =
    \begin{cases}
        p_{\mathrm{B}} & \text{if} \;\; \mathrm{NC}_{\overrightarrow{AB}} > \mathrm{NC}_{\overrightarrow{BA}}, \\
        p_{\mathrm{A}} & \text{otherwise}
    \end{cases}
\label{eq:neural-coherence_data-selection}
\end{equation}

Before presenting the performance experiments, first provide an empirical analysis for the problem of training data selection when faced with an OOD target task (few-shot image classification). In Fig. \ref{fig:corr_valid-acc_vs_target-acc}, for each of the five target datasets, we report correlation between the source validation accuracy of models trained on different binary mixtures of those five datasets (a dataset is either completely included or excluded from the training data) on the horizontal axis, compared to the target performance that they yield on the vertical axis. We report correlations for mixtures that include the target domain (Corr. Target In), those that exclude it (Corr. Target Out) and for all mixtures (Corr. All). We observe that overall, rather unsurprisingly, there is no consistent correlation between the source performance (Max Acc Valid) and the target performance (Max Acc Target). The overall observation is that one cannot know in advance whether validation accuracy will be correlated with target accuracy. In our experiments, for some datasets there is a positive correlation, for others there is negative correlation, and while for others there is no correlation at all.

\textbf{Performance Experiments:}
For our experiments, we use few-shot image classification datasets, with 5-way 1-shot tasks. We use the MAML algorithm for training the models, an established meta-learning algorithm for dealing with few-shot learning scenarios. We use the standard 4-layer ConvNet architecture proposed by \cite{DBLP:journals/corr/VinyalsBLKW16}. We rely on a set of five diverse image datasets : Mini-Imagenet, Omniglot, Quickdraw, Traffic Sign, Describable Textures. For each dataset used as the target, the model selection task consists in selecting which of the remaining four datasets to pre-train on, so as to achieve the best performance given the downstream dataset. 
We demonstrate in Tab. \ref{tab:exp:data_selection} the performance of the Neural Coherence approach in predicting the best model depending on the pre-training dataset. Neural Coherence outperforms the Source Validation baseline in all scenarios. We report the target performances obtained from the Oracle, the baseline, and Neural Coherence. We also report the accuracy of Neural Coherence in selecting the optimal training dataset, with random chance prediction being 25\%. For most target problems (Mini-Imagenet, Omniglot, Textures), the validation baseline is significantly subpar, Neural Coherence outperforms it, filling a significant portion of the generalization gap, and completely fills the gap (Textures). For Quickdraw, where the baseline performs as well as the Oracle hence one cannot perform better, and Neural Coherence performs just as well. Finally, for Traffic Sign, the baseline struggles massively, and Neural Coherence performs significantly better, but accurate data selection remains difficult.

\begin{table}[H]
\centering
\begin{tabular}[t]{l|ccc|cc}
\toprule
\textbf{Target Dataset}
&
\multicolumn{3}{c}{\begin{tabular}{c}
     \textbf{Target performance}
\end{tabular}}
& \textbf{Selection Acc. of NC}
\\
&\textit{Oracle} &Source-Val&\textbf{NC}& (random=25\%)\\
\midrule
Mini-Imagenet&46.1&29.4&\textbf{35.5 ± 0.1}&88.0 ± 5.2\\
Omniglot&86.2&72.8&\textbf{82.7 ± 0.9}&74.0 ± 7.0\\
Quickdraw&55.6&\textbf{55.6}&\textbf{55.6 ± 0.1}&98.7 ± 1.8\\
Traffic Sign&90.6&37.7&\textbf{41.4 ± 0.6}&42.0 ± 7.9\\
Describable Textures&32.8&26.0&\textbf{32.7 ± 0.1}&98.7 ± 1.8\\
\bottomrule
\end{tabular}
\caption{Performance of Neural Coherence for training data selection : Neural Coherence outperforms the Source Validation baseline. For each target dataset, the task consists of predicting which of the remaining four constitutes the best training dataset. Each experiment is repeated over 50 trials, and NC uses $n=5$ samples. The \textit{target performances} are reported in the first three numerical columns. In the fourth column we report the \textit{selection accuracy} of Neural Coherence in predicting the best training dataset. Compared to the Oracle, when Source-Val is poor, NC fills most or all of the generalization gap. When Source-Val is on par with the Oracle, so is NC. When Source-Val struggles significantly, NC performs better, but accurate data selection remains difficult.
} \label{tab:exp:data_selection}
\end{table}%

\section{Empirical work leading to Neural Coherence} \label{sec:exps:foundation}

In this section we present an overview of our empirical work and experimental observations on the nature of out-of-distribution generalization, and its relation to the neural activation dynamics, that lead to our formulation of the Neural Coherence principle.

\begin{figure}[H]
\centering
    \subfloat[5-way 1-shot 5-step]{%
        \includegraphics[width=0.33\linewidth]{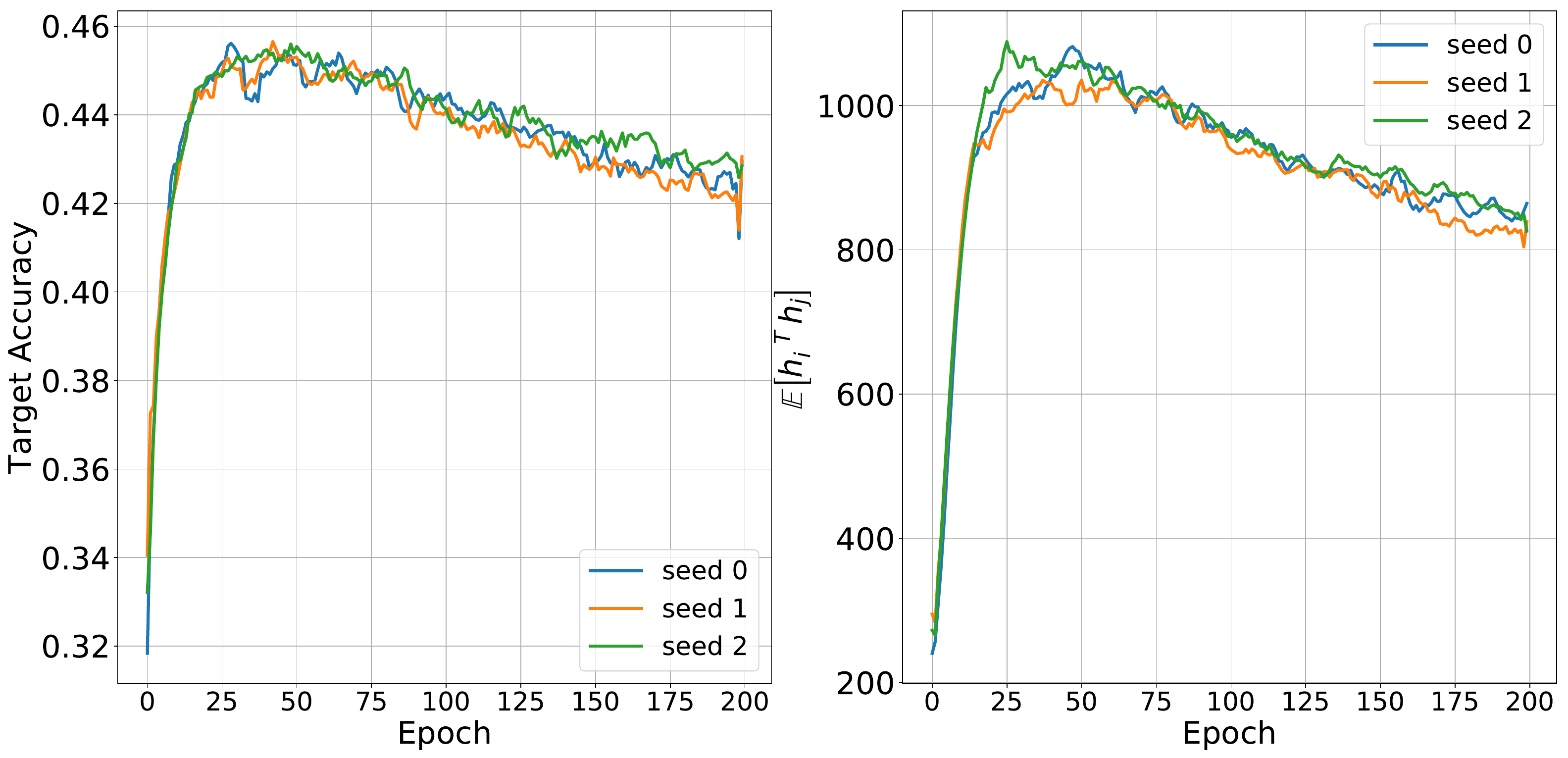}
        \label{fig:model_selection:mini_5w1s5s_fo}}
    \subfloat[5-way, 1-shot, 1-step]{%
        \includegraphics[width=0.33\linewidth]{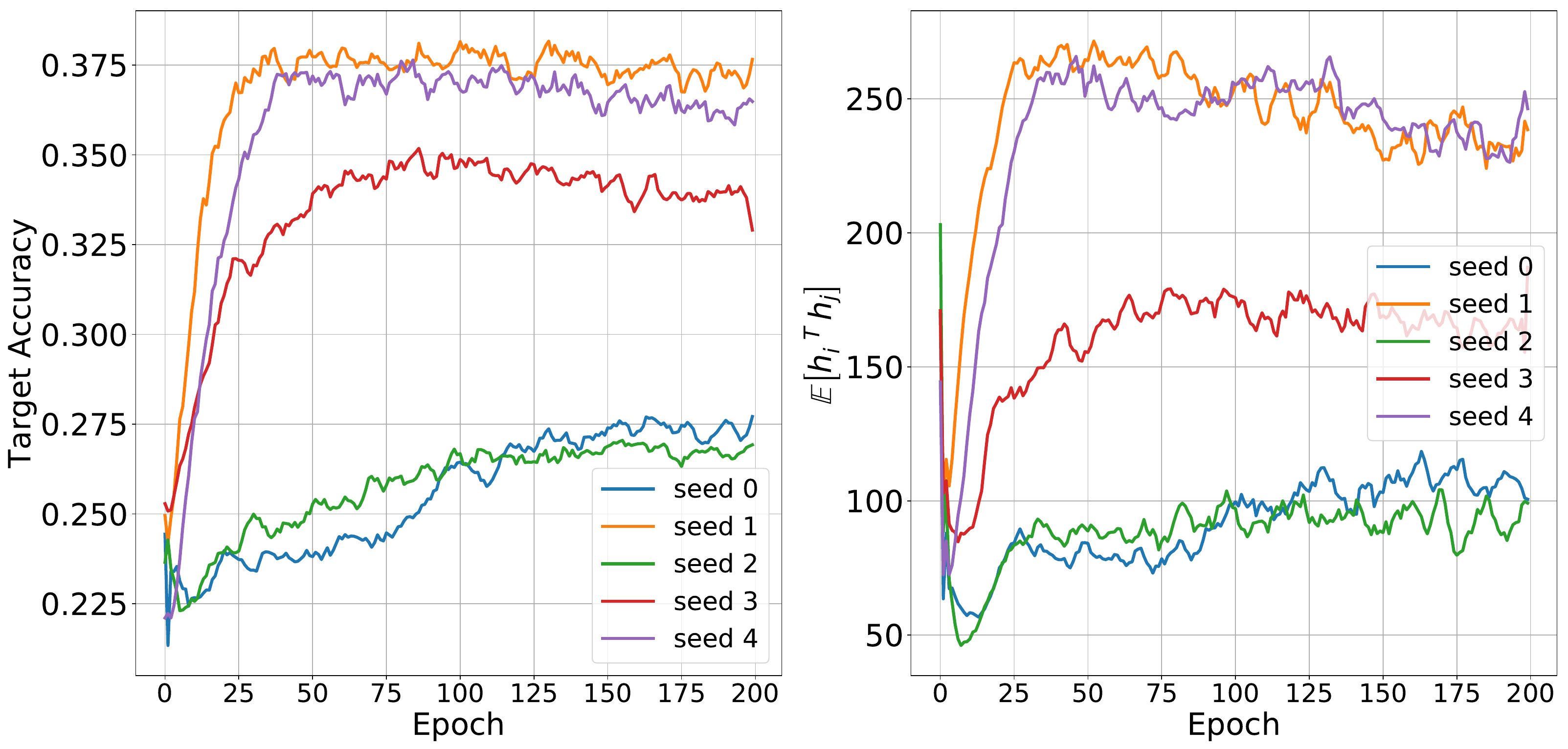}
        \label{fig:model_selection:mini_5w1s1s_fo}}
    \subfloat[5-way, 5-shot, 1-step]{%
        \includegraphics[width=0.33\linewidth]{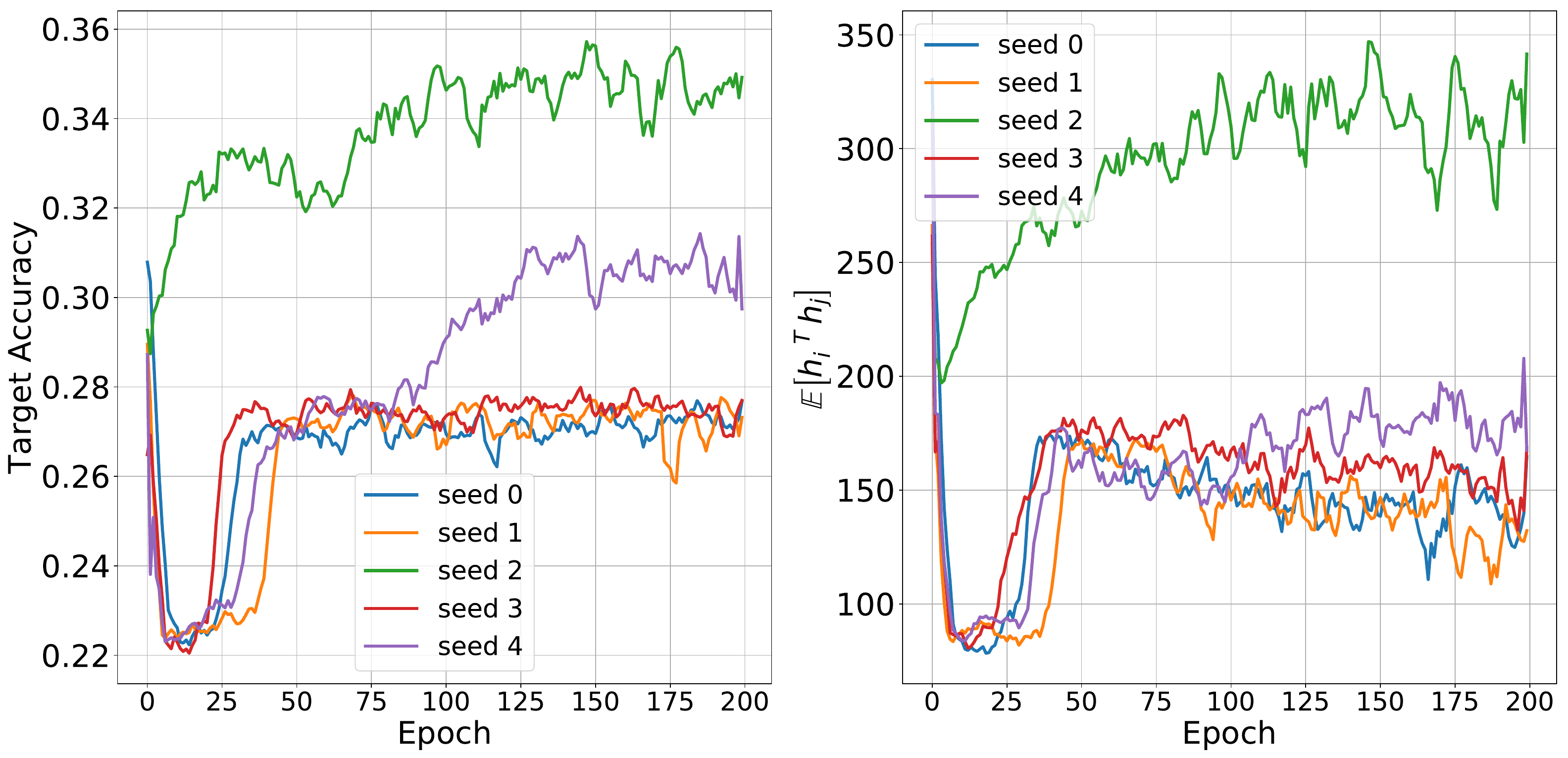}
        \label{fig:model_selection:mini_5w5s1s_fo}}
        \caption{Comparison between average inner product between representation vectors at the final layer $L$ of the feature extractor, and average target accuracy, in few-shot learning. The expected inner product is a linear combination of the four aggregated moments defined in Eq.\ref{eq:aggregated_moments}. We use different regimes of MAML and First-Order MAML on MiniImagenet. Here the expression $\mathbb{E}[\mathbf{h}_i^T\mathbf{h}_j]$ is the expected inner product between representations $\mathbf{h} \doteq f_L(\mathbf{x})$.
        }
        \label{fig:exp:inner_product:few-shot_learning} 
\end{figure}

\begin{figure}[H]
\centering
    \subfloat[Quickdraw to Quickdraw]{%
        \includegraphics[width=0.25\linewidth]{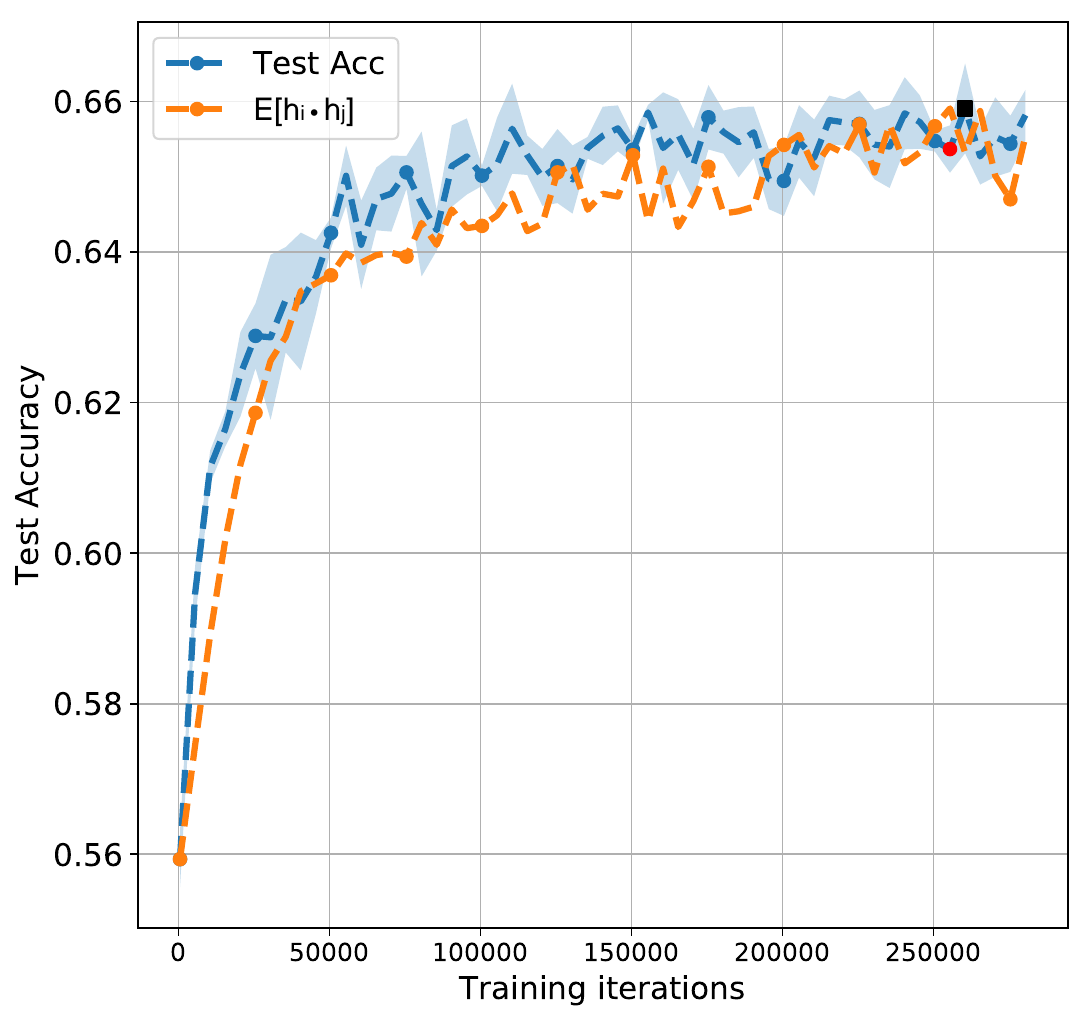}
        \label{fig:erip:quickdraw}}
    \subfloat[Quickdraw to Omniglot]{%
        \includegraphics[width=0.25\linewidth]{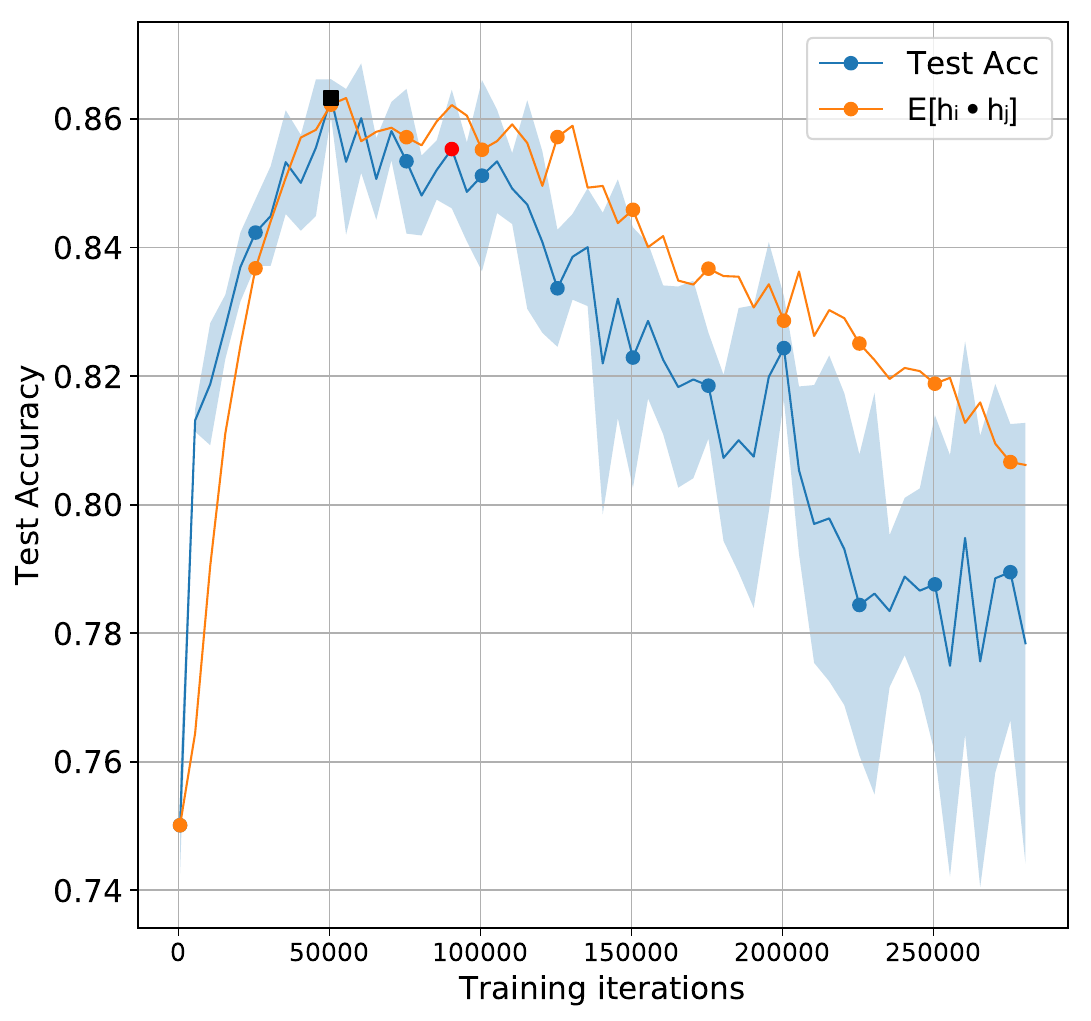}
        \label{fig:erip:miniimagenet}}
    \subfloat[Quickdraw to Imagenet]{%
        \includegraphics[width=0.25\linewidth]{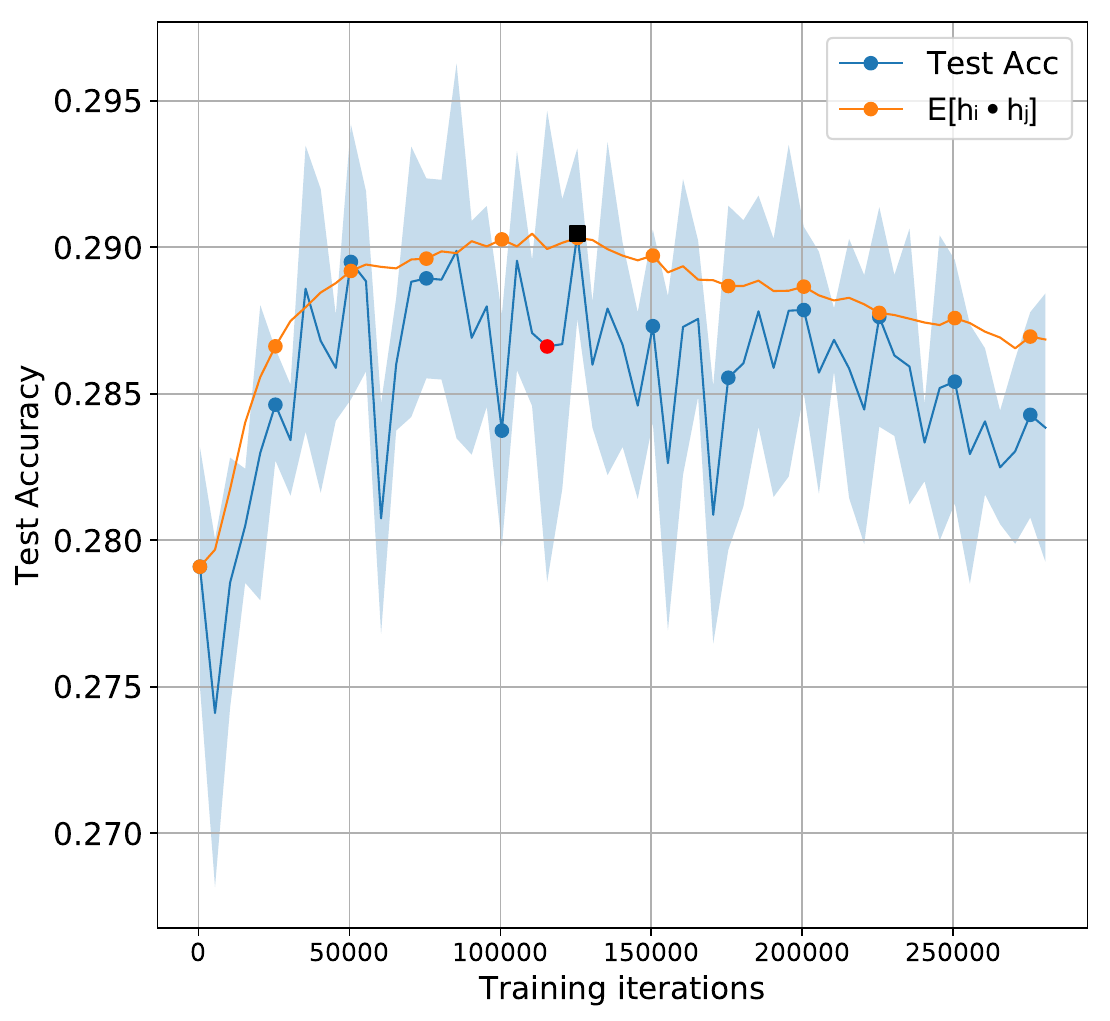}
        \label{fig:erip:aircraft}}
\caption{Measuring the average inner product between representation vectors, in Few-Shot Transfer Learning. MAML, 5-way 1-shot, training dataset : Quickdraw. The estimated early-stopping time of the metric $\hat{t}$ coincides well with the true optimal early-stopping time $t^*$. 
Considering the gap between 1) the average performance of validation early-stopping across the three settings (58.69\%); and 2) the maximum generalization across the three settings (61\%), the average performance of the metric is at 59.7\%, closing nearly half of the gap (43.74\% of the gap).
}
\label{fig:expected_inner:few-shot_transfer_learning}
\end{figure}

\textbf{Representation space and target generalization:}
We started off from previous work (Guiroy et al 2019) showing the relation between the out-of-distribution generalization of meta-learning, and the inner product between gradient vectors of downstream tasks. Upon expanding the scope of experimental setting, this parameter specific metric did not always reflect generalization. It was later observed (Guiroy et al, 2021) that the distribution of representations was more indicative of out-of-distribution generalization. More concretely, the expected inner product $\psi(\mathbf{z}) = \mathbb{E}_{\mathbf{z}_i, \mathbf{z}_j \sim p(\mathbf{z})}[  \mathbf{z}_i^T \mathbf{z}_j ]$, measured between the representation vectors $\mathbf{z} = f_L(\mathbf{x})$, correlated with downstream generalization in a broader range of settings.
See Fig. \ref{fig:exp:inner_product:few-shot_learning}, Fig. \ref{fig:expected_inner:few-shot_transfer_learning}.

\begin{figure}[H]
    \centering
    \subfloat[Matching Network, ResNet-18, 5-way 5-shot]{%
        \includegraphics[width=0.49\linewidth]{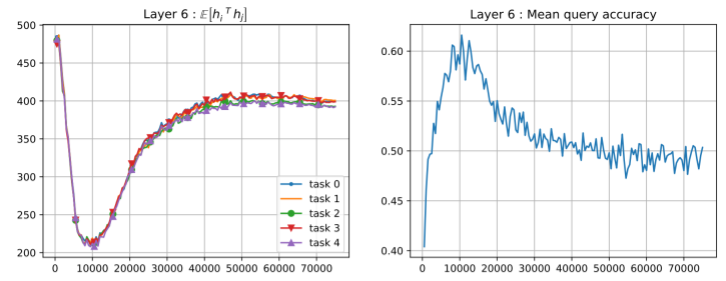}}
    \subfloat[Prototypical Network, ResNet-18, 5-way 1-shot]{%
        \includegraphics[width=0.49\linewidth]{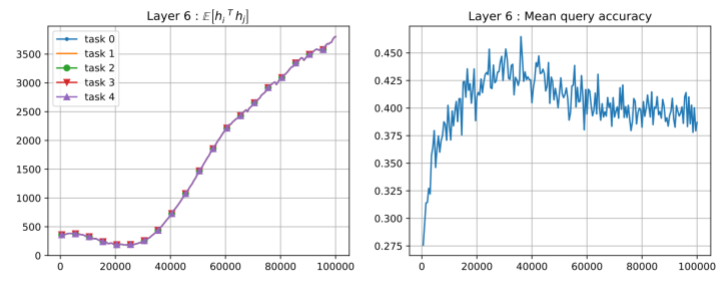}
        \label{fig:model_selection:mini_5w1s5s_fo_pn}}
\caption{ResNet-18 on MiniImagenet: The correlation is strong, but now negative, between the expected inner product of representations, and the target accuracy, where the \textit{minimum} of the metric coincides with the maximum of generalization, suggesting a more complex characterization of representation is needed. 
}
\label{fig:inverted_correlation}
\end{figure}

\begin{figure}[H]
    \centering
    \subfloat[Exclusive correlation of a specific metric with $Acc_{target}$]{%
        \includegraphics[width=0.25\linewidth]{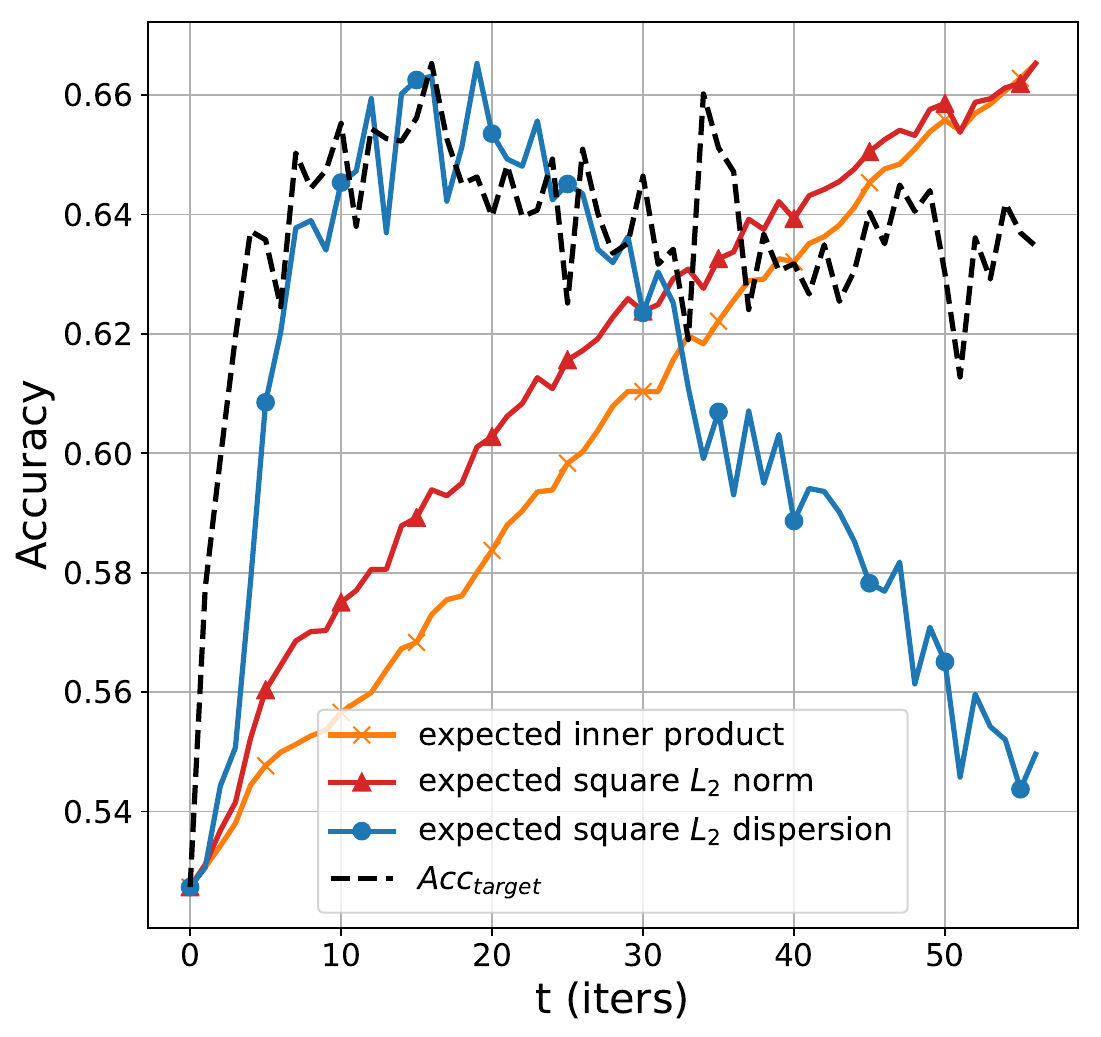}
        \label{fig:various_metrics:a}}
    \subfloat[Expected $l_2$ Norm]{%
        \includegraphics[width=0.25\linewidth]{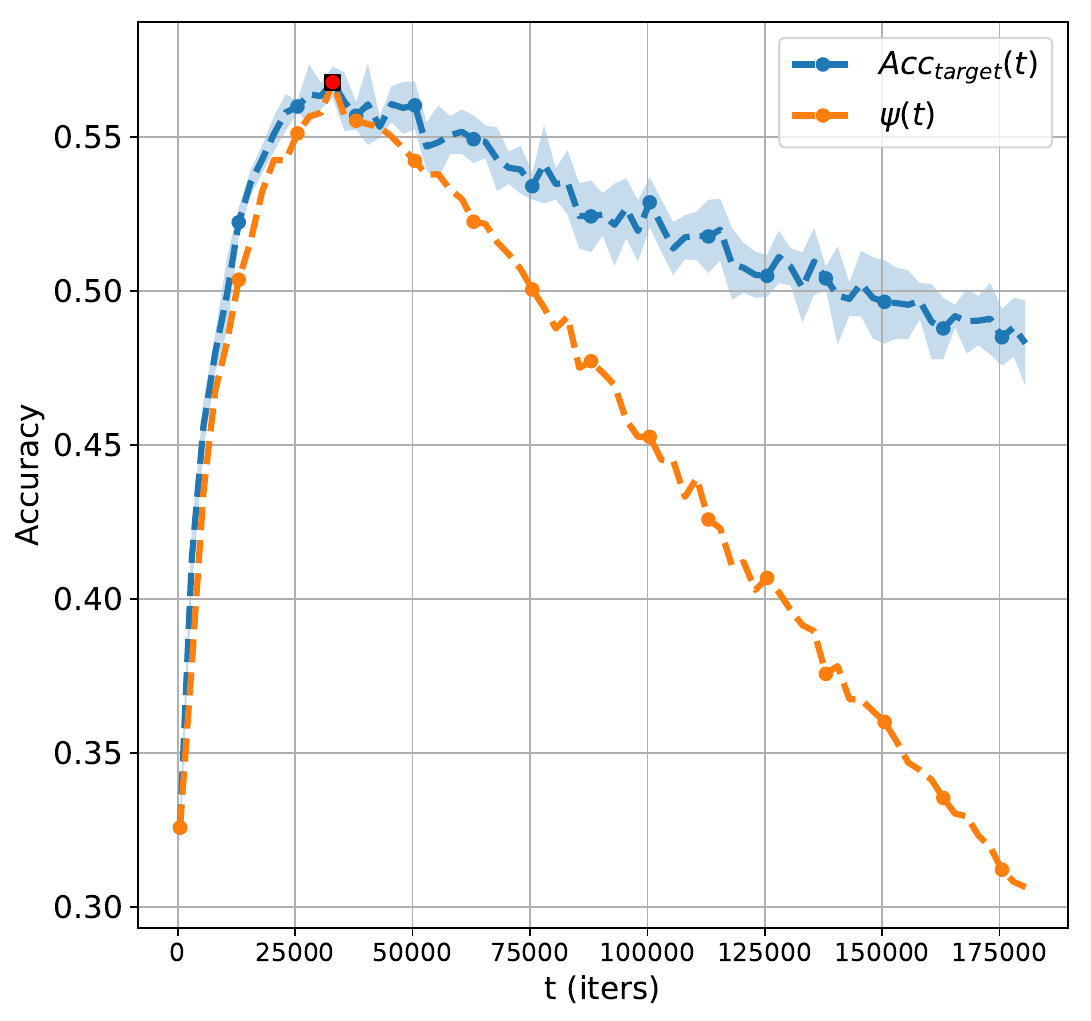}}
    \subfloat[Expected $l_2$ Dispersion]{%
        \includegraphics[width=0.25\linewidth]{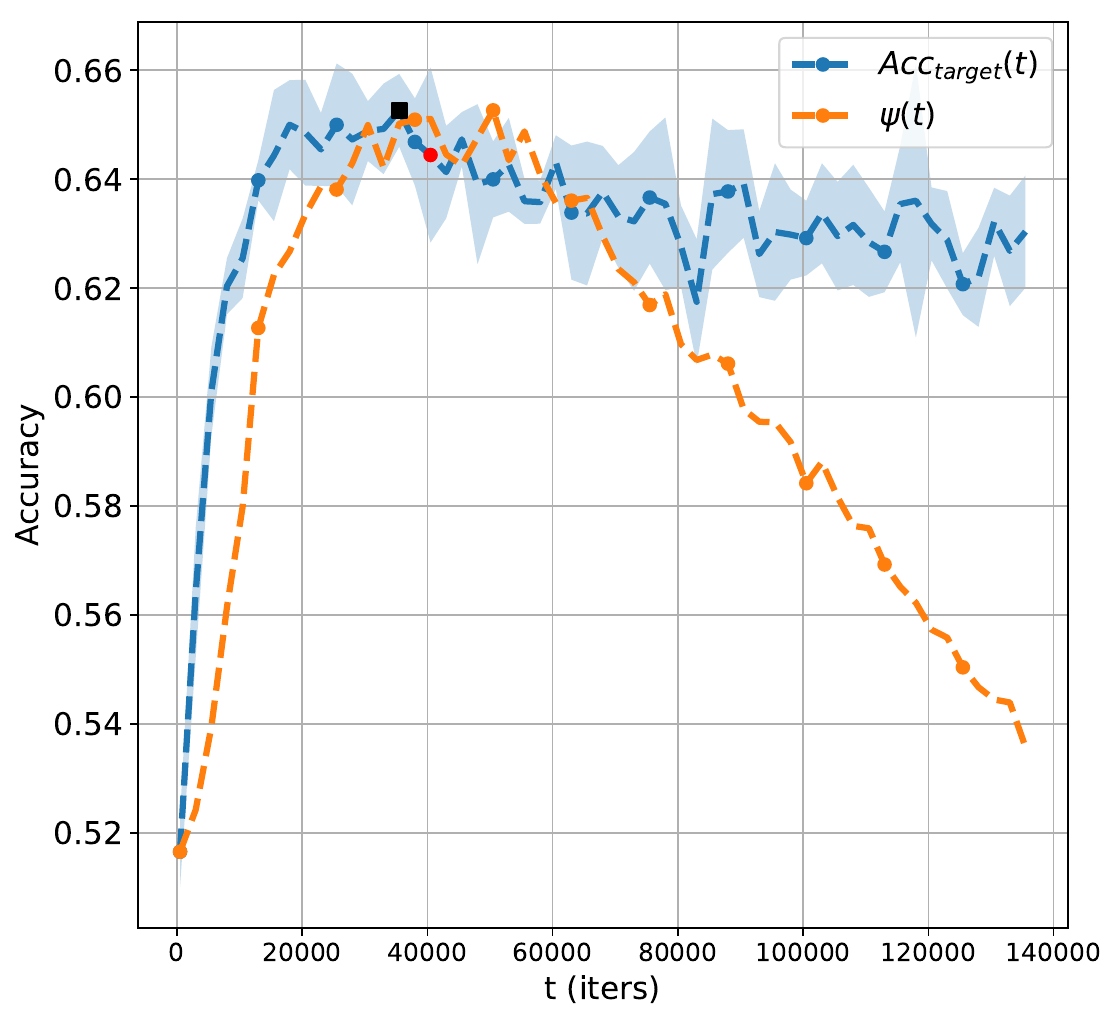}}
    \subfloat[Expected Feature-Wise Variance]{%
        \includegraphics[width=0.25\linewidth]{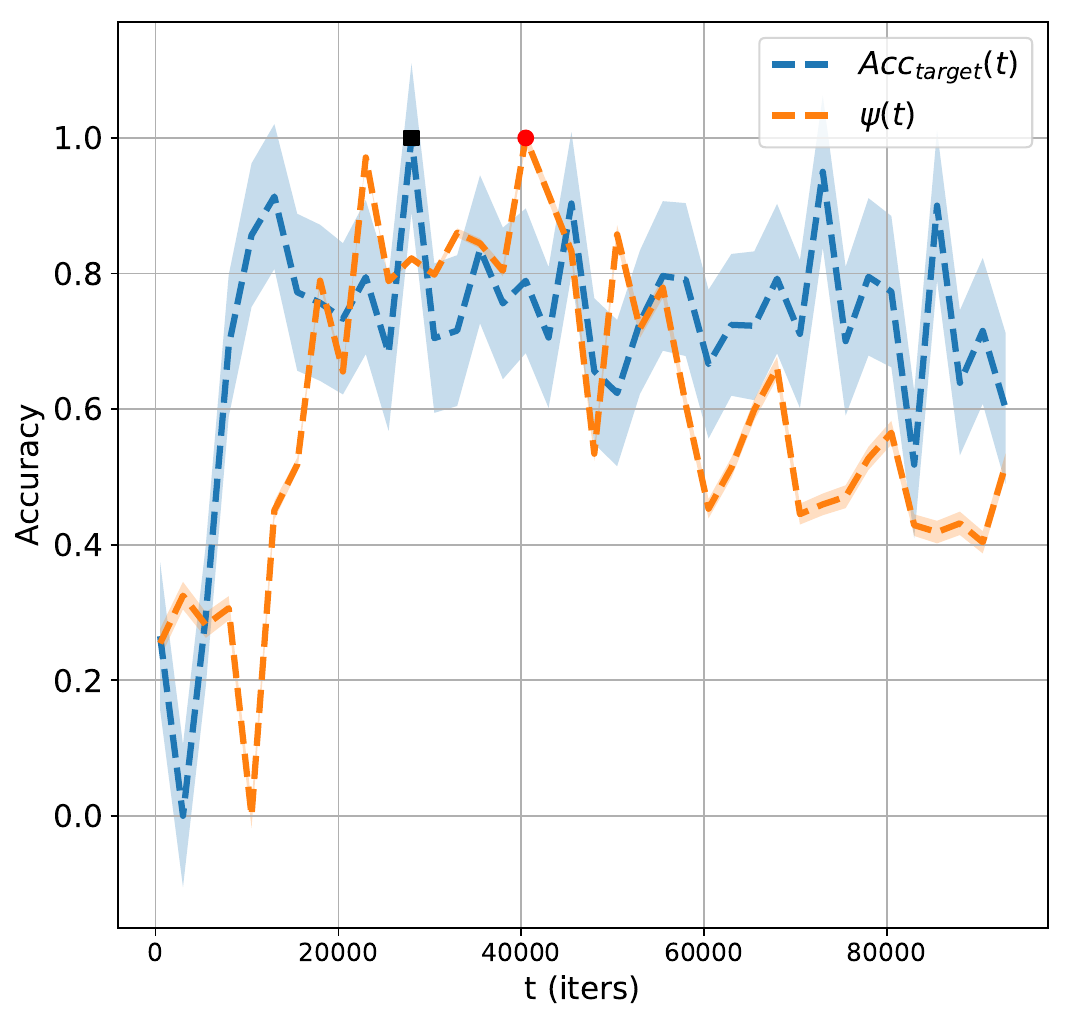}}    
    \caption{Different metrics of the representation space may have a strong correlation with generalization, other than the expected inner product. a) Prototypical, VGG Flower, 5-way 1-shot : out of three metrics which in other cases may be related with generalization (as in b), c), d)), here only the expected $l_2$ dispersion has a strong relation with generalization. b) Expected $l_2$ norm; c) Expected square $l_2$ dispersion, Prototypical Network, VGG Flower; d) Expected feature-wise variance, Prototypical Network, Omniglot to Quickdraw. 
    }
    \label{fig:various_metrics}
\end{figure}

\textbf{Beyond a Single Metric: Moment-Based Characterization of Representations}
Upon further expanding the scope of applications, i.e. by changing not only the dataset but also the learning algorithm, we again observed that this metric, even when measured at different layers, would not always reflect downstream generalization. However, by trying different metrics, we would retrieve the correlation with generalization. These metrics turned out to be linear combinations of the moment-wise moments computed over the batch-wise moments, defined earlier. This suggested that rather than focusing on a single metric, the distribution of the neural activations should be characterized by a more general function $\psi$. See Fig. \ref{fig:inverted_correlation}, Fig. \ref{fig:various_metrics}.

\begin{figure}[H]
\centering
    \includegraphics[width=0.24\linewidth]{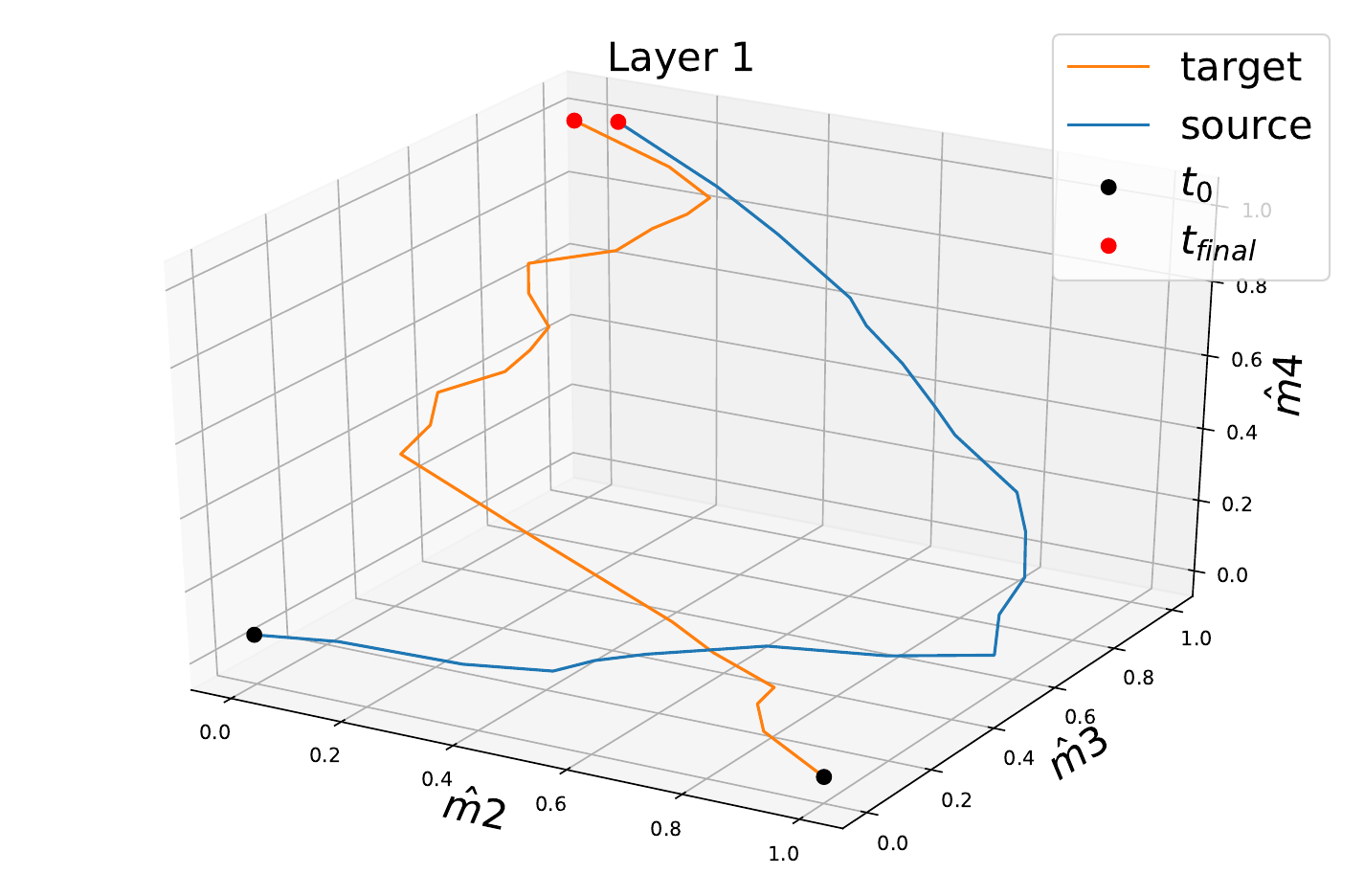}
    \includegraphics[width=0.24\linewidth]{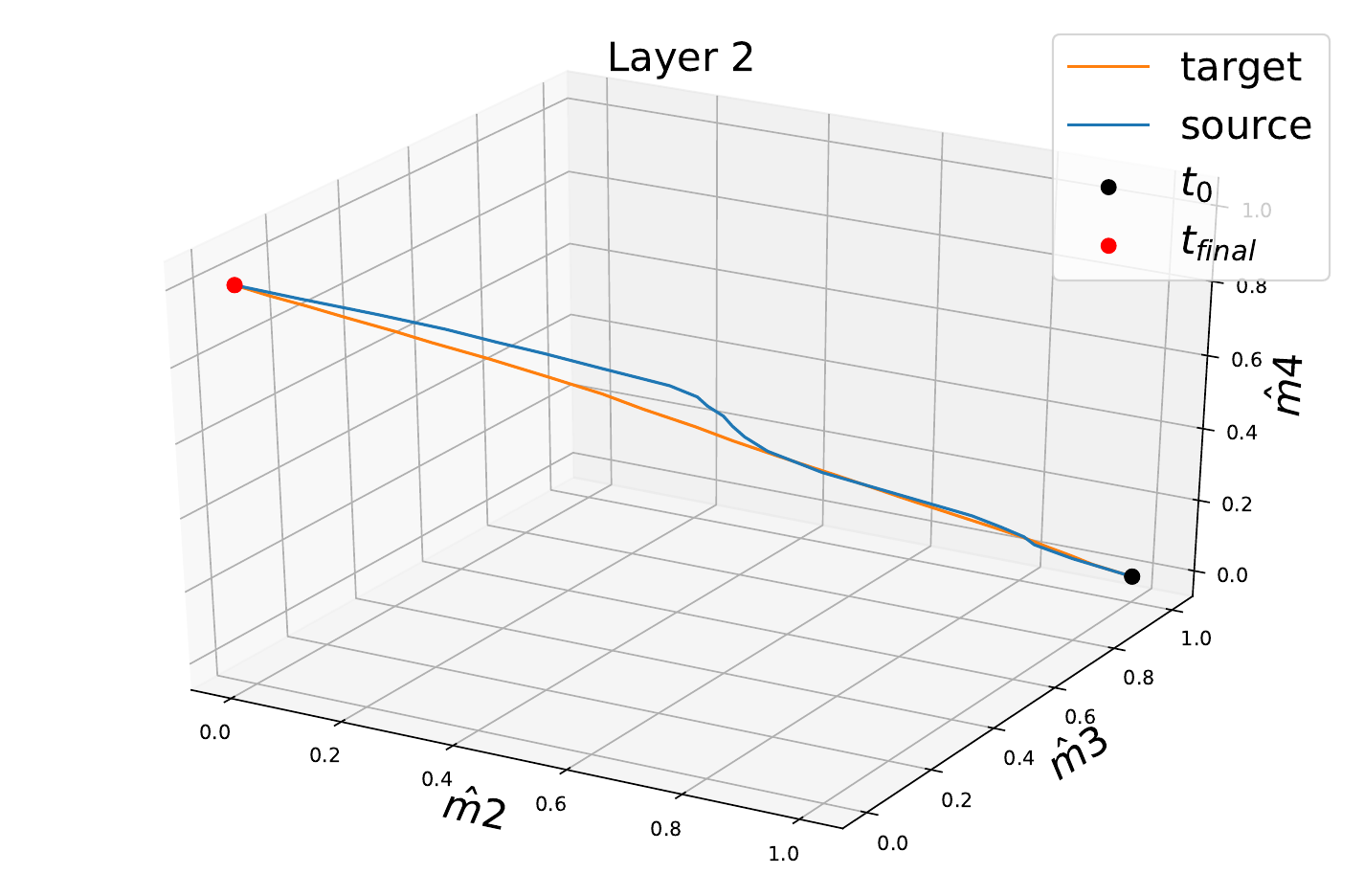}
    \includegraphics[width=0.24\linewidth]{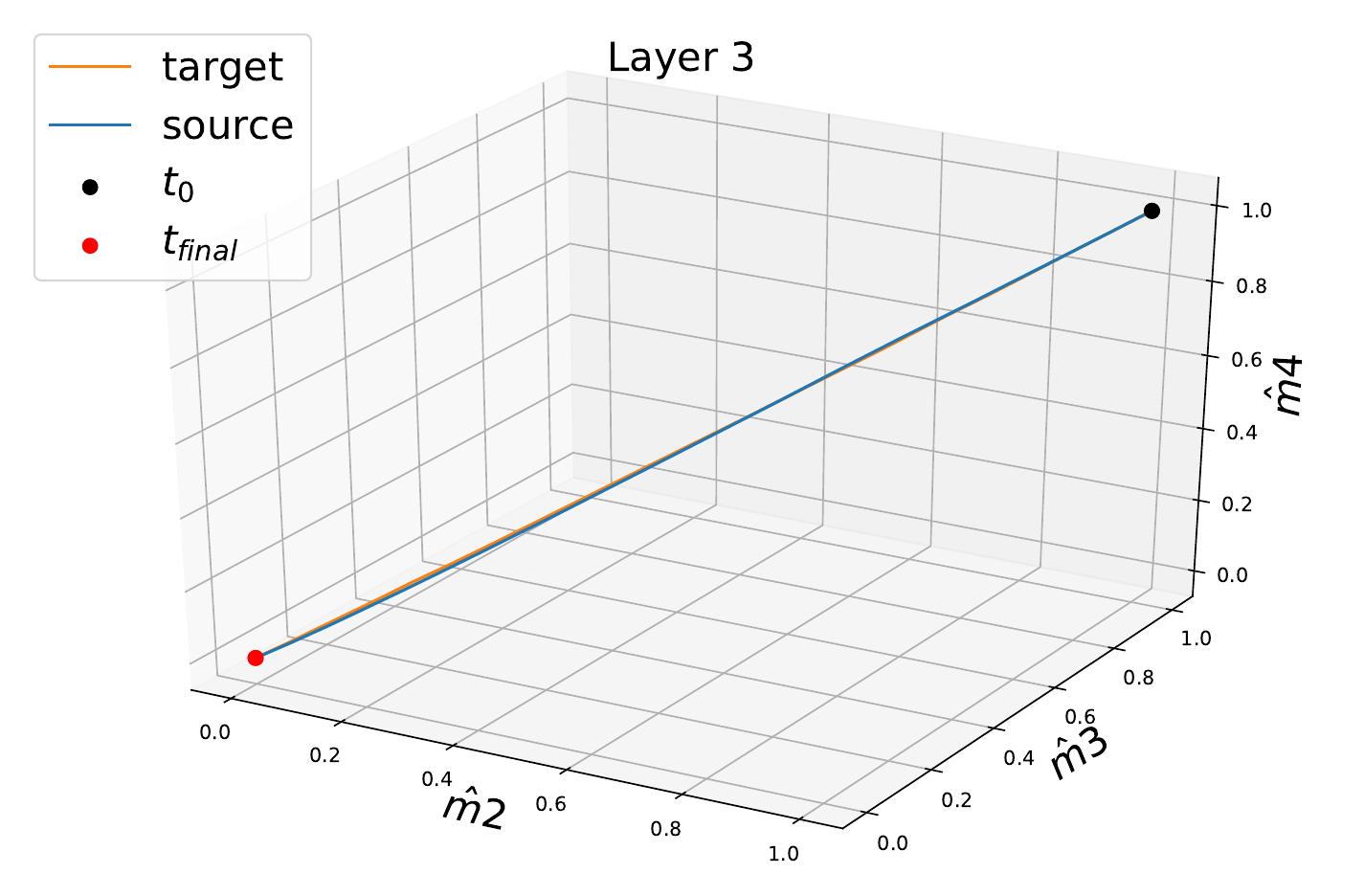}
    \includegraphics[width=0.24\linewidth]{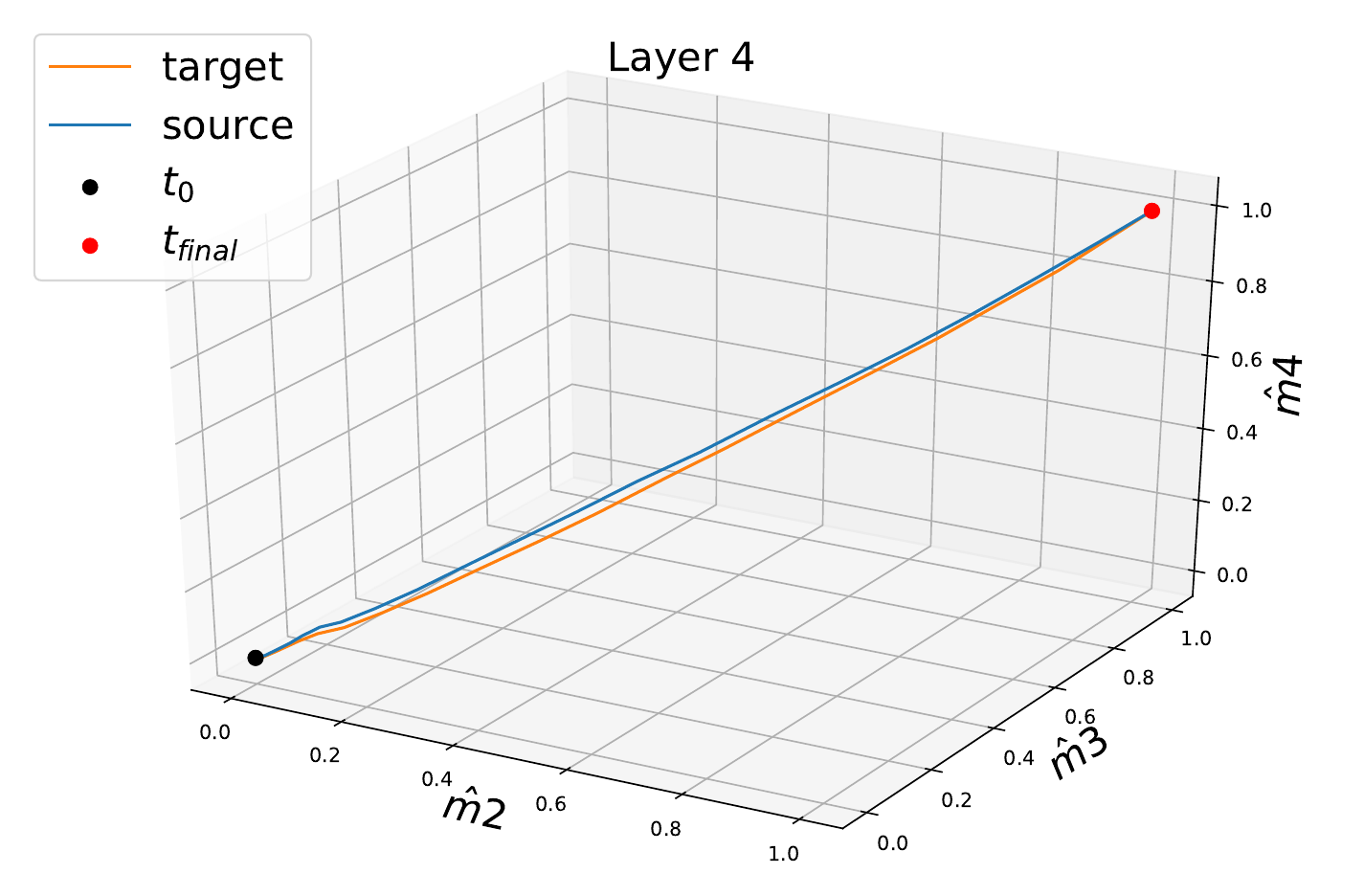}
    \newline
    \includegraphics[width=0.24\linewidth]{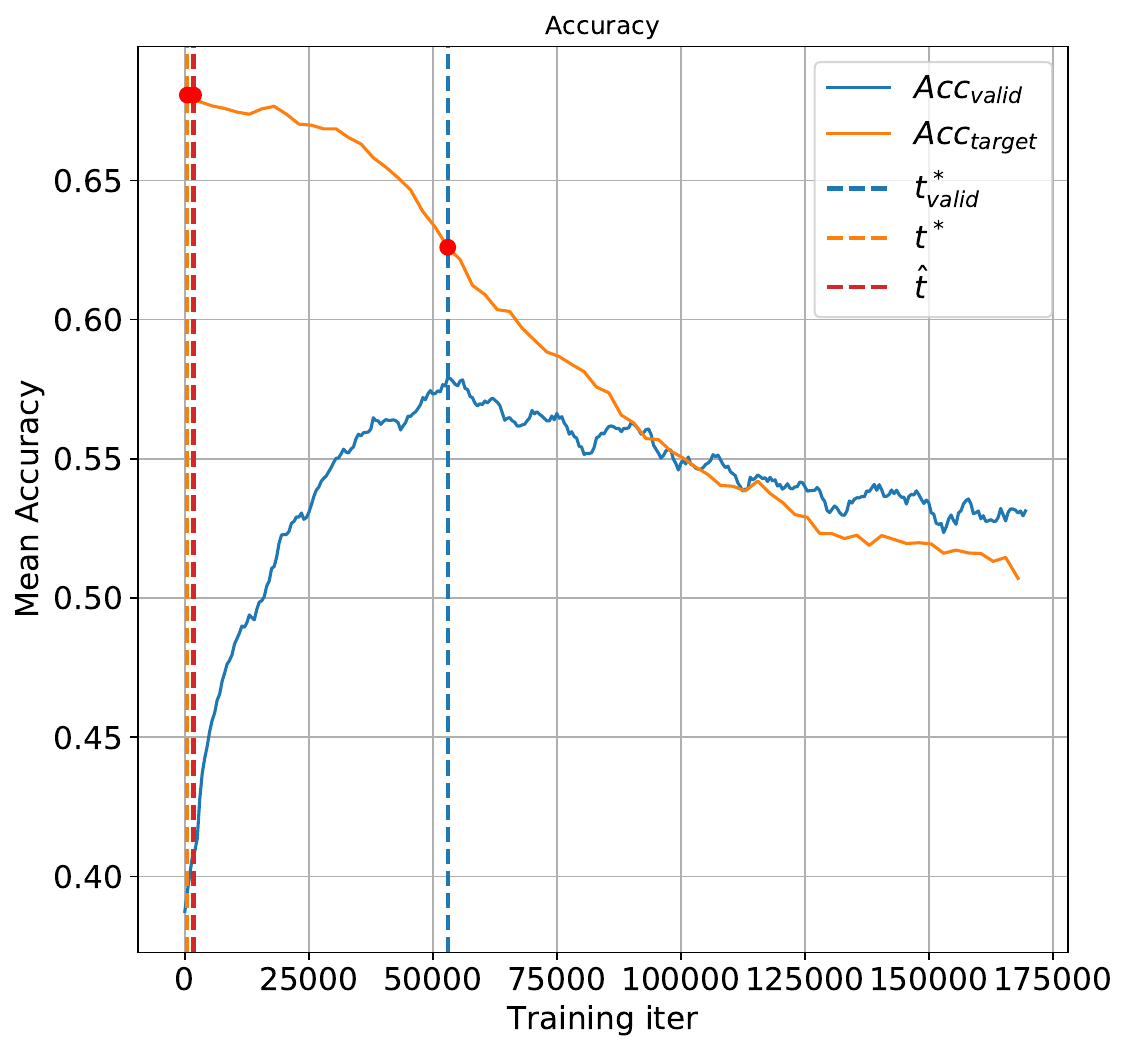}
\caption{MAML : CU Birds to Omniglot. The divergence between the target and source activation trajectories happens at layer 1, mostly along the aggregted moment $\hat{m}_2$, and is detected from the beginning of training. This accurately corresponds to when overfitting occurs. The resulting early-stopping time $\hat{t}$ is close to the optimum $t^*$, which drastically improves generalization, compared to validation-based early-stopping. For better visualization, we only show the three most significant aggregated moments and trajectory curves are normalized. 
}
\label{fig:critical-depth:maml-birds-to-omniglot}
\end{figure}

\begin{figure}[H]
\centering
    \includegraphics[width=0.24\linewidth]{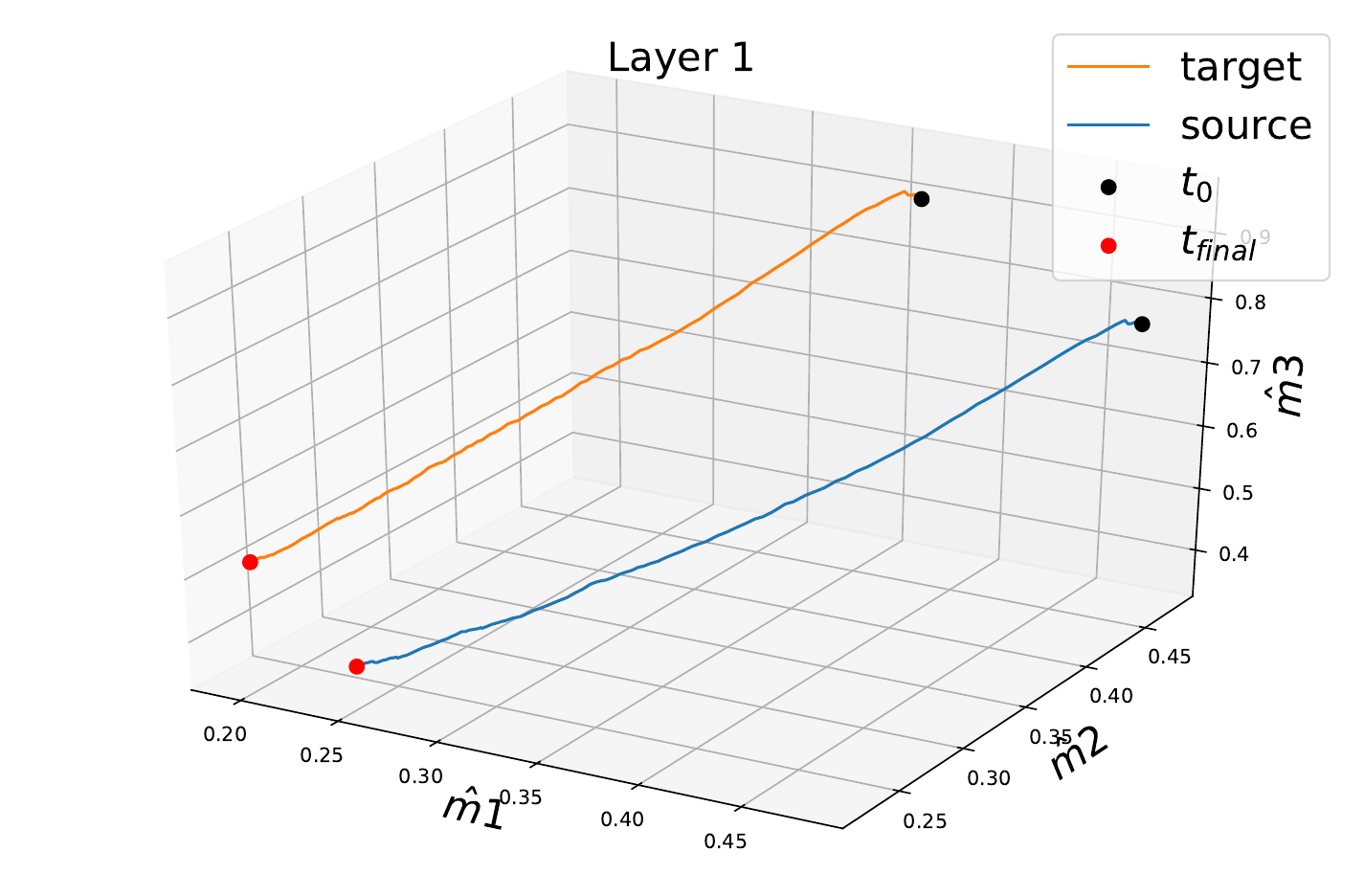}
    \includegraphics[width=0.24\linewidth]{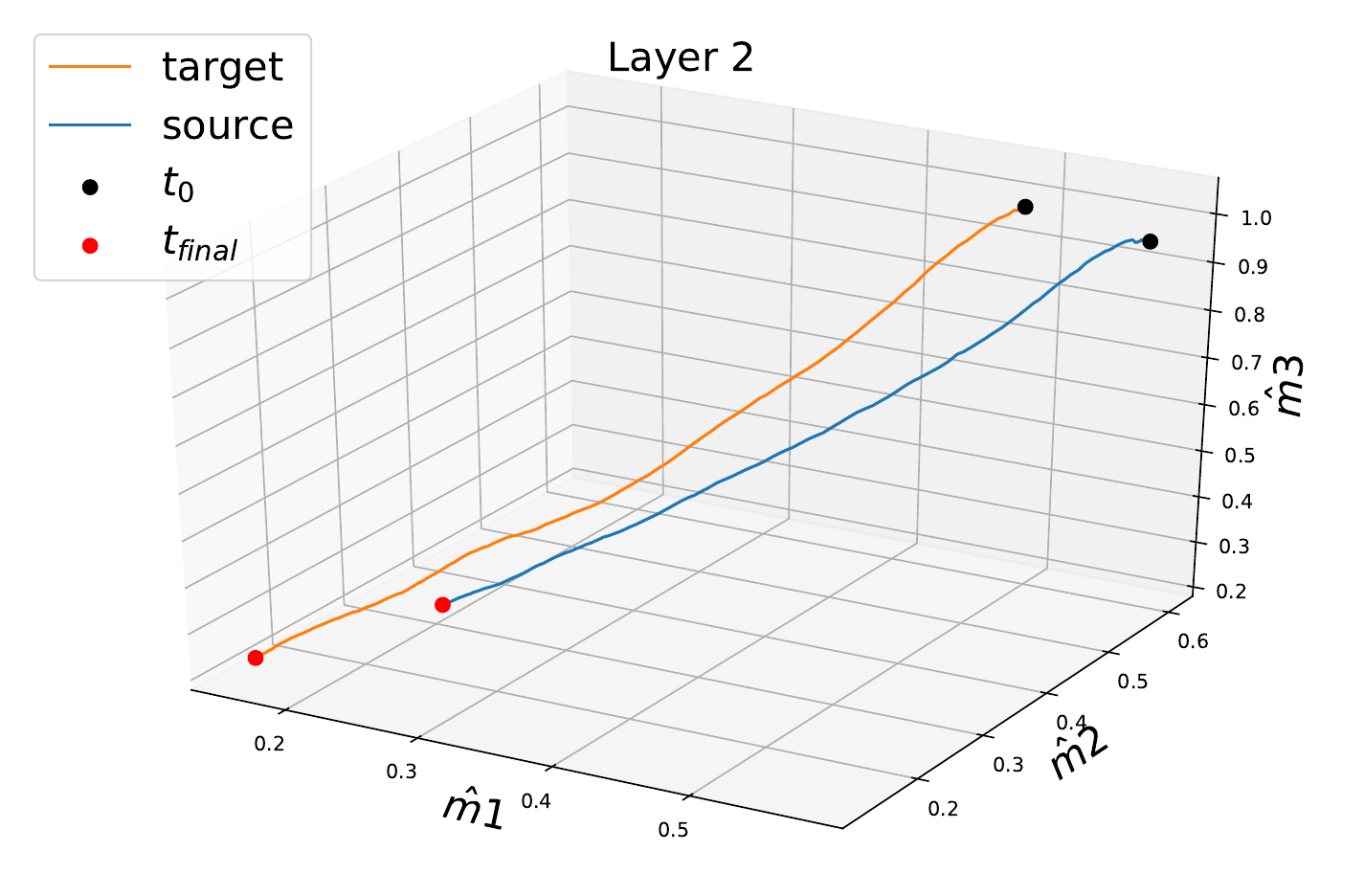}
    \includegraphics[width=0.24\linewidth]{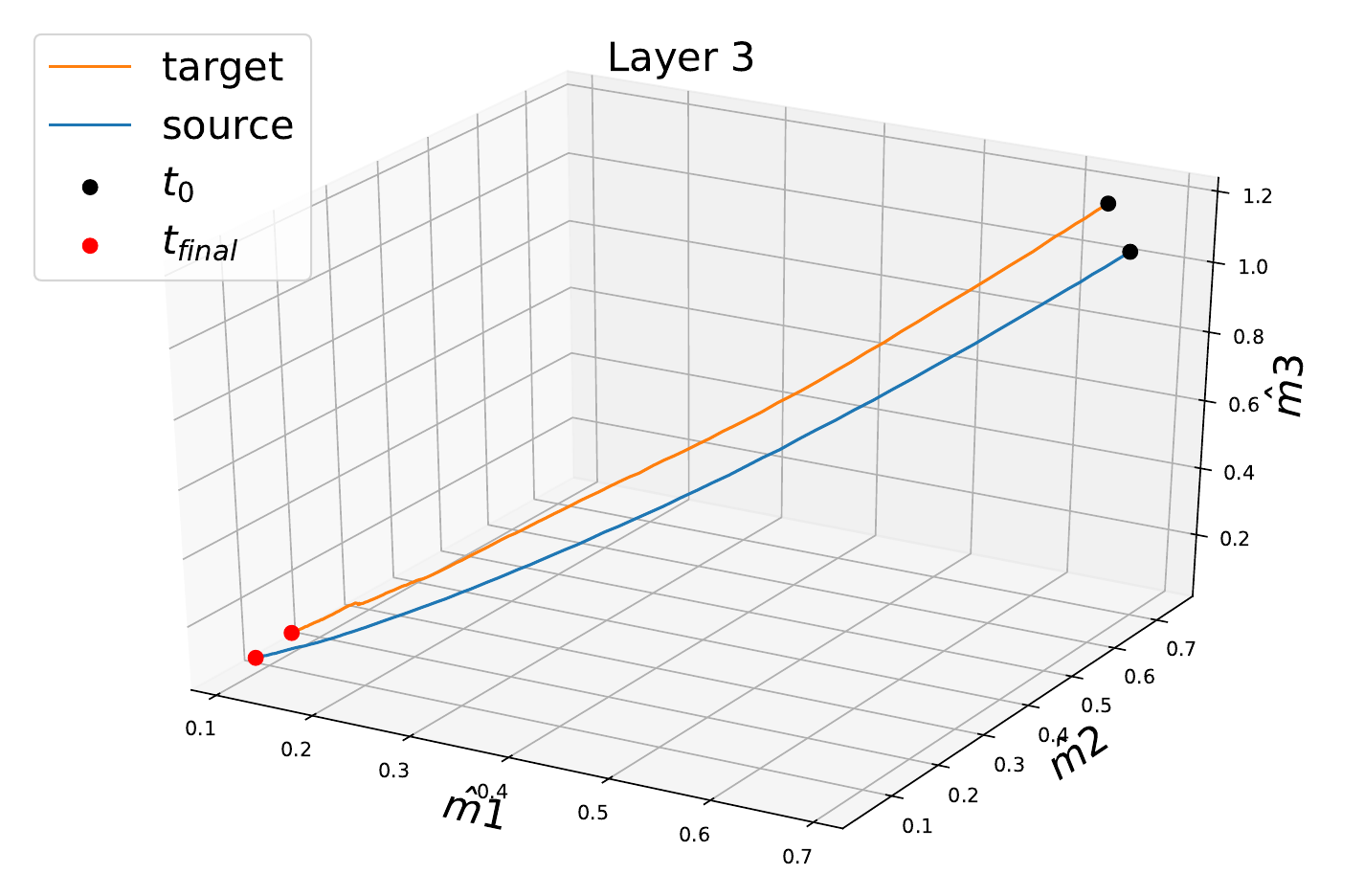}
    \includegraphics[width=0.24\linewidth]{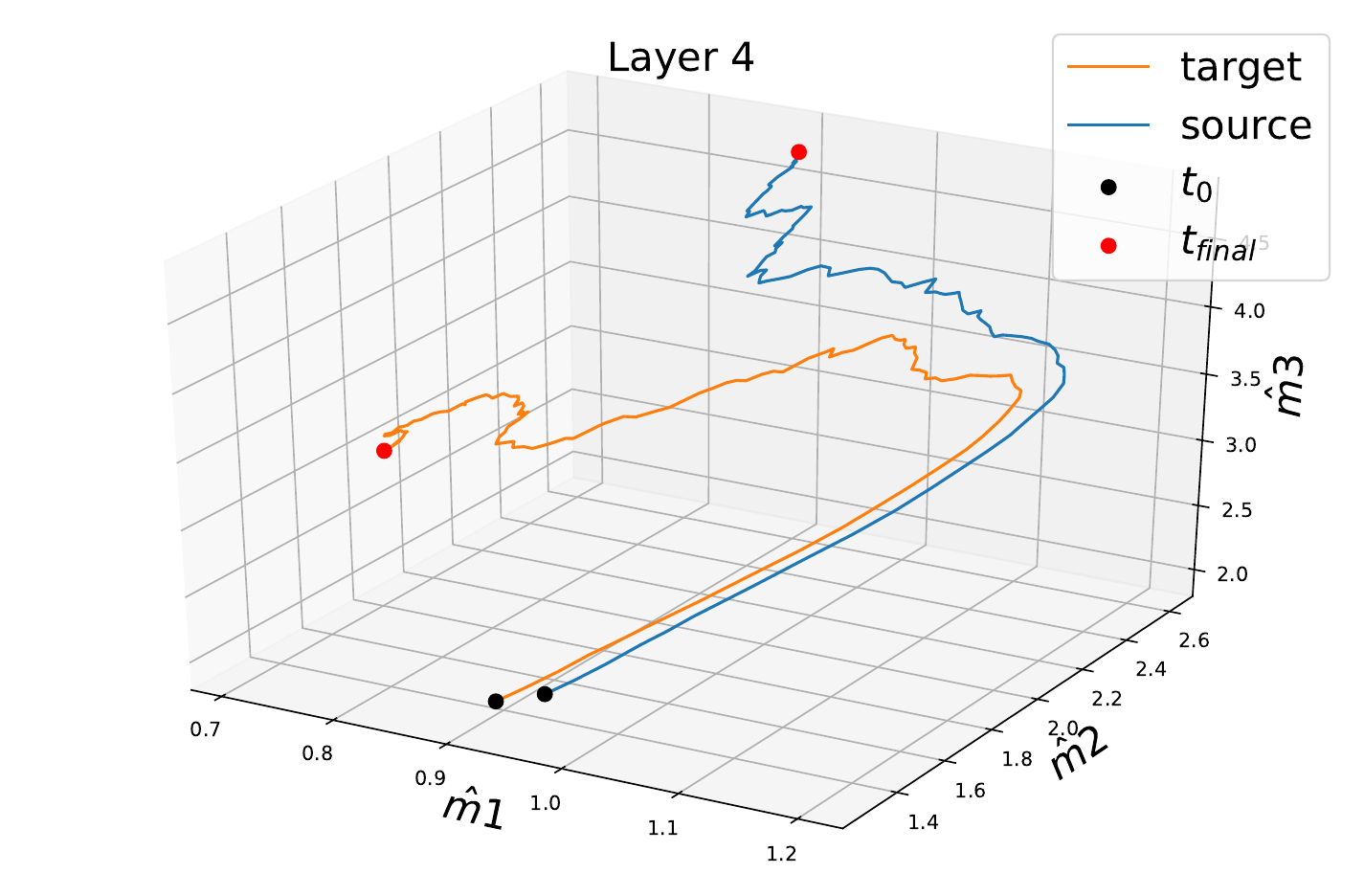}
    \newline
    \includegraphics[width=0.24\linewidth]{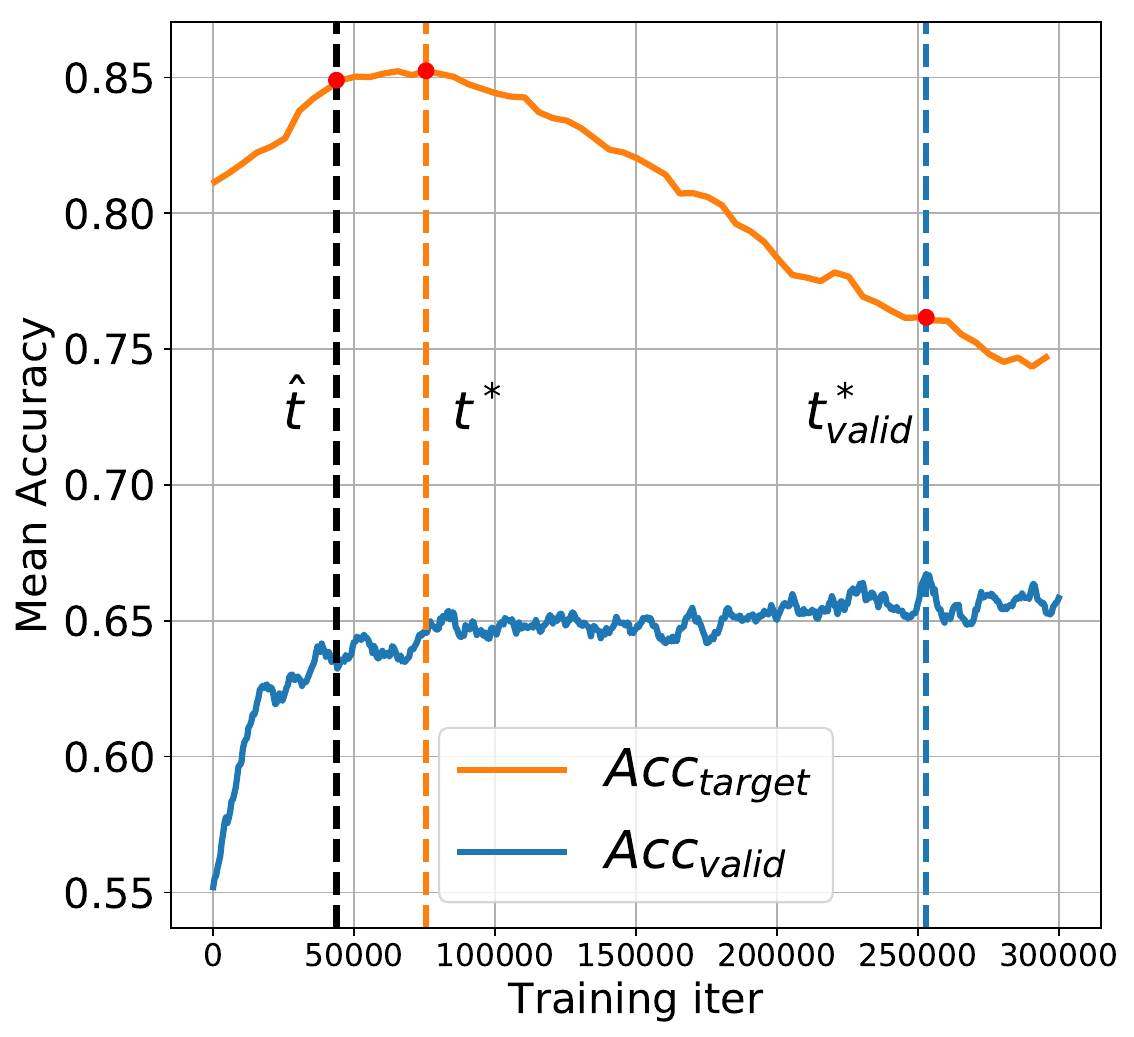}
\caption{MAML : Quickdraw to Omniglot. The divergence between the target and source activation trajectories happens at layer 4, mostly along $\hat{m}_2$. This accurately corresponds to when overfitting occurs. The resulting early-stopping time $\hat{t}$ is close to the optimum $t^*$, which drastically improves generalization, compared to validation-based early-stopping. 
}
\label{fig:critical-depth:maml-quickdraw-to-omniglot}
\end{figure}

\textbf{Layer-wise Trajectories and Directional Coherence}
Later on, we expanded the scope of few-shot learning datasets under consideration. We observed that the expected inner product of representation vectors $\mathbb{E}[ \mathbf{z}_i^T \mathbf{z}_j ]$ did not always reflect generalization. 
However, when allowing to measure this statistic on the neural activation vectors on other layers, this correlation with generalization would reappear.
See Fig. \ref{fig:critical-depth:maml-birds-to-omniglot}, Fig. \ref{fig:critical-depth:maml-quickdraw-to-omniglot}.
The point-wise distance between the trajectories did not reflect well downstream performance. Indeed in Fig. \ref{fig:critical-depth:maml-birds-to-omniglot}, right after $t_0$ the source and target trajectories get closer, yet target generalization already starts to decrease. 

However, when looking at the \textit{direction} of the trajectories, we see that, from $t_0$ they start to go in opposite directions. Similarly, when looking at target and source accuracy, their generalization gap also decreases from the beginning, but we don't care about that, what want to find is the instant $t^*$ where $Acc_{target}$ is optimum, in other words, when the target and source accuracies start to go in different directions. This insight laid to defining the principle of Neural Coherence as a measure of directional similarity of activation trajectories, rather than a point-wise distance.

\begin{figure}[H]
    \centering
    \subfloat[Critical layers]{%
        \includegraphics[width=0.25\linewidth]{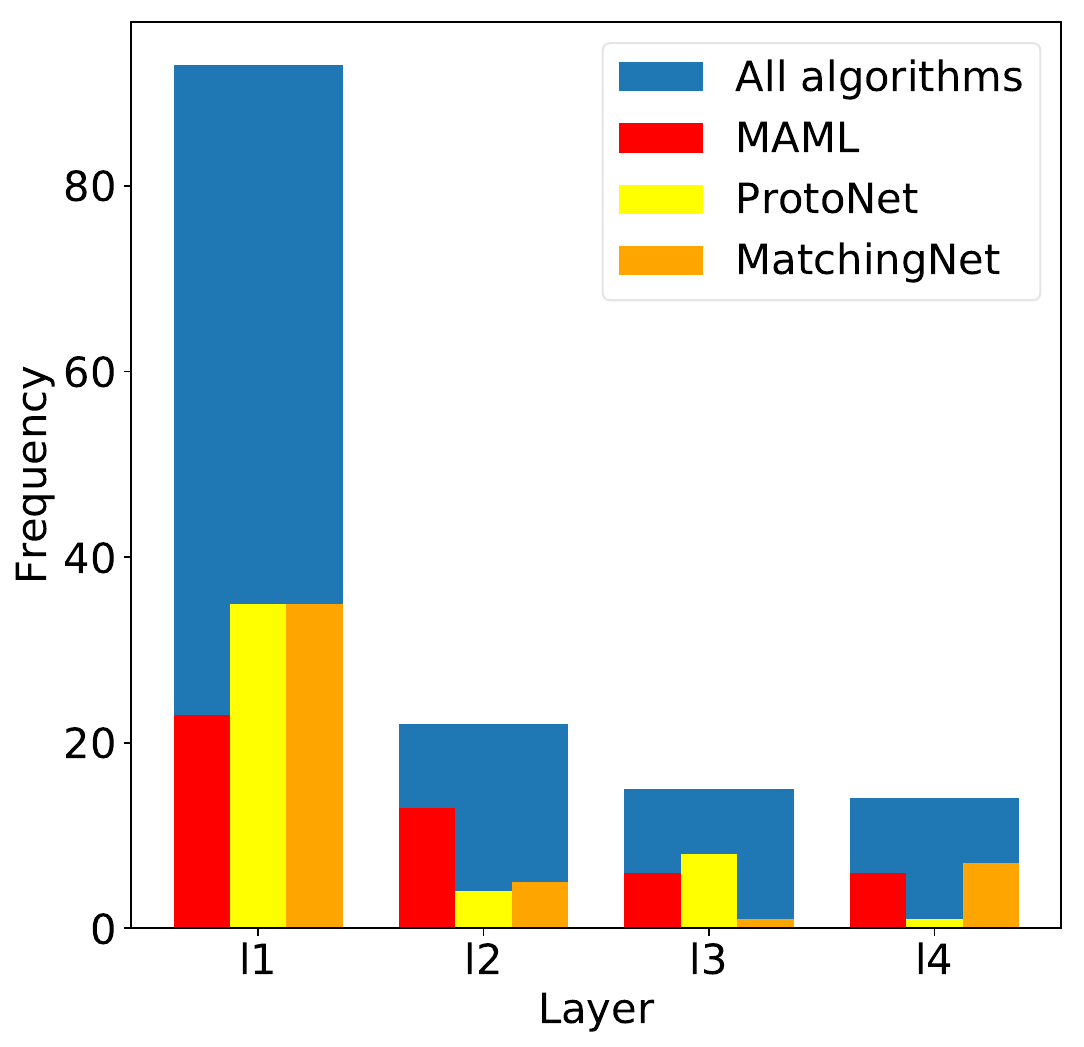}
        \label{fig:hist:critical_layers}}
    \hspace{0.03 in}
    \subfloat[Critical moments]{%
        \includegraphics[width=0.25\linewidth]{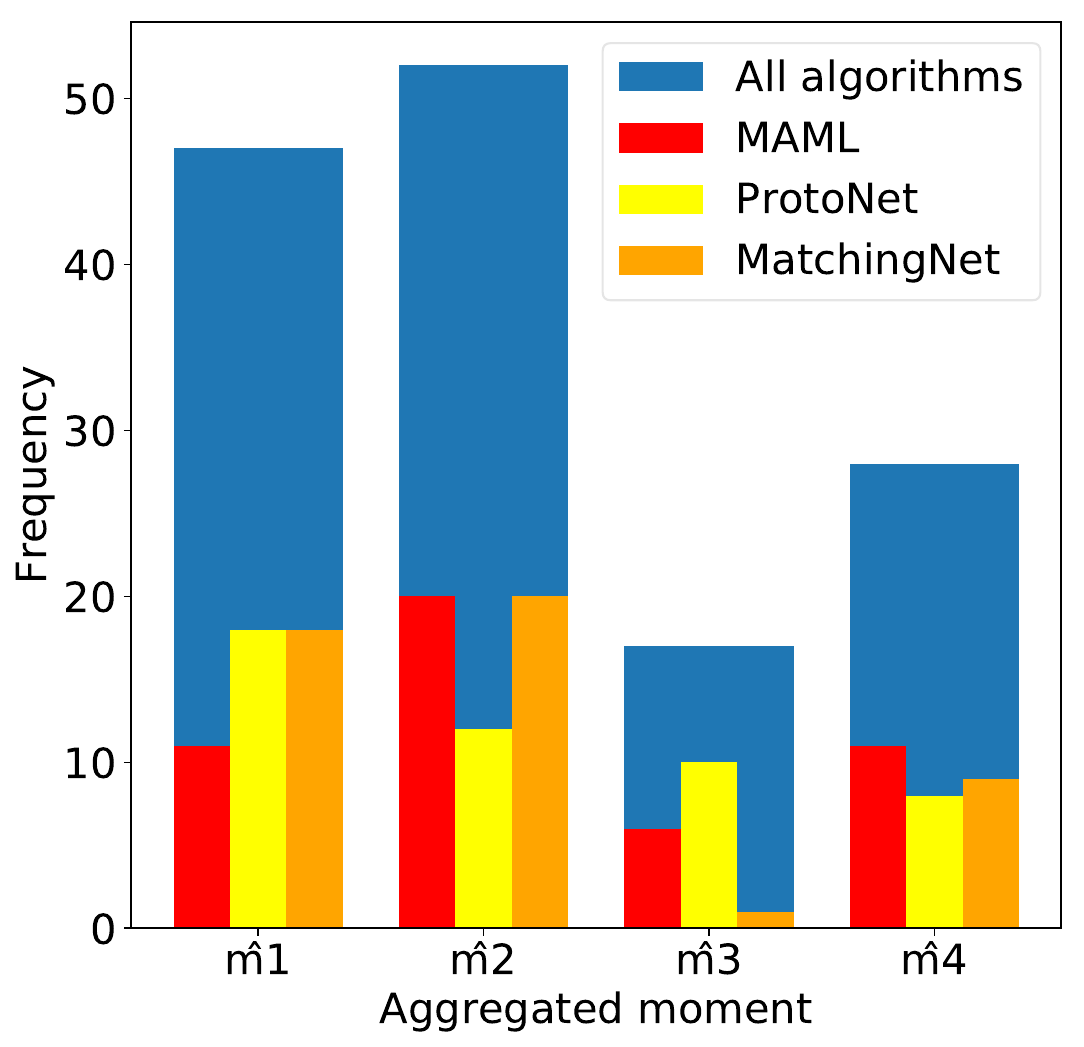}
        \label{fig:hist:critical_moments}}
    \caption{Critical layers and critical moments histograms, across all experiments. For a given experiment, the tuple of critical layer and critical moment represents the layer $l$ and moment $\hat{m}$ where target and source activation distribution trajectories diverge the most. Results show that any layer of the feature extractor may be critical, but they happen more often at shallow depths (a). Likewise, results also indicate that all moments may be critical .(b).}\label{fig:hist:critical_layers_and_moments}
\end{figure}

\textbf{Critical moments and critical layers}
We then analyze the frequency of which layer and which aggregated moments are critical, \textit{i.e.} where we observed the strongest divergence between the trajectories of the target and source activations. In Fig.~\ref{fig:hist:critical_layers}, we observe that the strongest divergence may happen at all layers, but tends to happen more frequently in earlier layers. This further motivates the idea of ``looking under the hood'' to infer how target generalization is evolving, rather than simply looking at the representations of the last layer. Moreover, Fig.~\ref{fig:hist:critical_moments} suggests that all of the aggregated moments that we introduced, from $\hat{m}_1$ to $\hat{m}_4$, might be critical in driving the divergence between the activation trajectories.

\section{Conclusion} \label{sec:conclusion}
In this work we introduced \textit{Neural Coherence}, a general principle for model and data selection in deep learning under distribution shift. Neural Coherence views a trained network through the lens of its activation statistics across layers and over training, and uses the alignment between source and target activation trajectories to infer when improvements on the source task begin to harm performance on an out-of-distribution target. This perspective yields a family of practical criteria that can operate with only a handful of unlabeled target examples, and that are agnostic to architecture and training objective. 
We instantiated this principle for checkpoint selection and evaluated it across a broad set of OOD scenarios, including meta-learning for few-shot classification, zero-shot generalization, and transfer learning with modern vision foundation models. Our experiments show that Neural Coherence consistently outperforms common practices such as source-domain validation-based early stopping and simple heuristics, often recovering a large fraction of the gap to an oracle that has full access to target labels. In particular, we demonstrate strong gains even when restricted to as few as 5 unlabeled target samples, highlighting the statistical efficiency of activation-trajectory–based criteria. Beyond checkpoint selection, we further showed that the same principle can guide the choice of pre-training data, again using only a small number of target examples.
More broadly, Neural Coherence suggests that activation dynamics—rather than final-layer performance alone—provide a rich signal for reasoning about generalization in deep networks, especially in the presence of dataset shift and limited supervision. We believe this offers a useful design pattern for future work on hyperparameter, architecture, and data selection: define a trajectory over training or configuration space, characterize how source and target activation distributions co-evolve along that trajectory, and select operating points where their coherence breaks down. Extending this framework to other hyperparameters (e.g., learning rate schedules, regularization strength, batch size), to other modalities such as language and multimodal models, and to more principled theoretical analyses of when and why Neural Coherence succeeds are promising directions for further research.

\paragraph{Acknowledgements and Disclosure of Funding} We would like to acknowledge funding support from NSERC, CIFAR, and compute support from Digital Research Alliance of Canada (Compute Canada) and Calcul Qu\'ebec.

\vskip 0.2in
\bibliography{references}

\appendix
\section{Additional experimental details}\label{sec:app:exp_detail}
\textbf{Standard ConvNet: }
We use the architecture proposed by \cite{DBLP:journals/corr/VinyalsBLKW16} and used by \cite{DBLP:journals/corr/FinnAL17}, consisting of 4 modules stacked on each other, each being composed of 64 filters of of 3 $\times$ 3 convolution, followed by a batch normalization layer, a ReLU activation layer, and a 2 $\times$ 2 max-pooling layer. With Omniglot, strided convolution is used instead of max-pooling, and images are downsampled to 28 $\times$ 28. With MiniImagenet, we used fewer filters to reduce overfitting but used 48 while MAML used 32.
\noindent
\textbf{ResNet: }
We use the same implementation of the Residual Network as in \cite{Triantafillou2020Meta-Dataset}. For most of the hyperparameters, we follow the setup of \cite{Triantafillou2020Meta-Dataset}, but we set the main few-shot learning hyperparameters so as to follow the original MAML setting more closely, and in each setting, we consider a single target dataset at a time, with a fixed number of shots and classification ways. We use 5 steps of gradient descent for the task adaptations, 15 shots of query examples to evaluate the test accuracy of tasks. We don't use any learning rate decay during meta-training, and step-size of 0.01 when fine-tuning the models to new tasks.
\noindent
\textbf{ConvNeXt}
Training the ConvNext-L architecture was performed on 32 Tesla V100 GPU in parallel. The model was trained for 600 epochs using the otherwise same hyperparameter setup as in the original publication \citep{convnext}, necessitating nearly two weeks of training time. The code use relies primarily on the PyTorch library. The source code and model checkpoints will subsequently be made publicly available.
\noindent
\textbf{Vision Transformer}
We used the standard hyperparams of the Vision Transformer \citep{mae}.

\end{document}